# Utilization and Experimental Evaluation of Occlusion Aware Kernel Correlation Filter Tracker using RGB-D

by

**Srishti Yadav**

B.Tech., Dr. A.P.J. Abdul Kalam Technical University, 2016

Thesis Submitted in Partial Fulfillment of the
Requirements for the Degree of
Master of Applied Science

in the
School of Engineering Science
Faculty of Applied Science

© Srishti Yadav 2021
SIMON FRASER UNIVERSITY
Spring 2021

# Declaration of Committee

| | |
|---|---|
| **Name:** | **Srishti Yadav** |
| **Degree:** | **Master of Applied Science** |
| **Thesis title:** | **Utilization and Experimental Evaluation of Occlusion Aware Kernel Correlation Filter Tracker using RGB-D** |
| **Committee:** | **Chair:** Bonnie Gray<br>Professor, Engineering Science |
| | **Shahram Payandeh**<br>Supervisor<br>Professor, Engineering Science |
| | **Jie Liang**<br>Committee Member<br>Professor, Engineering Science |
| | **Zhenman Fang**<br>Examiner<br>Assistant Professor, Engineering Science |

# Abstract

Unlike deep-learning which requires large training datasets, correlation filter-based trackers like Kernelized Correlation Filter (KCF) uses implicit properties of tracked images (circulant matrices) for training in real-time. Despite their practical application in tracking, a need for a better understanding of the fundamentals associated with KCF in terms of theoretically, mathematically, and experimentally exists. This thesis first details the workings prototype of the tracker and investigates its effectiveness in real-time applications and supporting visualizations. We further address some of the drawbacks of the tracker in cases of occlusions, scale changes, object rotation, out-of-view and model drift with our novel RGB-D Kernel Correlation tracker. We also study the use of particle filter to improve trackers' accuracy. Our results are experimentally evaluated using a) standard dataset and b) real-time using Microsoft Kinect V2 sensor. We believe this work will set the basis for better understanding the effectiveness of kernel-based correlation filter trackers and to further define some of its possible advantages in tracking.

# Acknowledgments

Working on the thesis has been a long yet fulfilling journey and it is the courtesy of the wonderful people who I have worked with for the completion of this thesis and people who have supported me during this journey. I am incredibly grateful for the wonderful support of the below-mentioned people.

I would like to show my greatest appreciation to my senior supervisor Dr. Shahram Payandeh. I can't thank him enough for your valuable guidance and tremendous support throughout my thesis work. Without his incredible patience, moral support, especially during the COVID era, it would not have been possible to reach my goals. This acknowledgment would be incomplete without sincere thanks to his prompt email responses (a bar set too high for me to achieve) which expedited my research at every stage. His consistent encouragement for publishing my work and giving me the freedom to explore my interests has made me a better researcher. I am also thankful for his review of all my work. I feel privileged to have worked with him.

I am also very grateful to my examining committee for reviewing my thesis. I would like to acknowledge my supervisor, Dr. Jie Liang, for guiding me in the duration of my thesis program, especially in my initial years. I would like to thank Dr. Zhenman Fang for being the internal examiner of my thesis and taking time out for my thesis. I would also like to express my appreciation to Dr. Bonnie Gray for chairing the presentation of my thesis.

Special thanks to my friends and lab partners Maryam Rasoulidanesh and Nasreen Mohsin for their help in my research work and life in general. It was fantastic to have the opportunity to work with them in Networked Robotics and Sensing Laboratory. The lab spirit has always been about doing good work. Each lab member has taught me something, and I will try to absorb the lessons with complete sincerity.

I would also like to thank Mary, graduate program assistant of the department, for her quick replies in resolving all my queries, especially during the last few months of my thesis program. I am also very grateful to Wen Dee, the graduate program coordinator of the department who was most patient and helped me a lot in my graduate program queries. These two people have played big roles, time and again, in making my time at school a smooth one.

I benefitted from the help of many study participants during my research. Thanks to the graduate students of ENSC that I have not mentioned by name, for their time and effort during my experiments.

I am grateful to my beloved parents, my younger brother and Supratim, who provided me with moral and emotional support in my life. I am the first girl in all generations of my family and extended family who has made her way to international studies and it would not have been possible without the faith my family put in me at all stages of my life. I sincerely owe it to my parents for trusting me as I explored charters unknown to them. I owe it to my brother for his endless calls in the past three years and for keeping me motivated for my career goals. Also, this journey wouldn't have been possible without Supratim, for his constant encouragement, love, and moral support.

Finally, I would like to express my heartfelt thanks to my friends and colleagues Anurag Sanyal, Mallika Oberoi, Lisette Lockhart, Naveen Vedula, Rakesh Gangwar, Reetu Yadav, Riya Dashoriya, Saad Mehmood, Sanjhal Jain, Yue Ling (Eric), and all others who have been a part of this journey but have not been explicitly named above.

# Table of Contents

# List of Tables

# List of Figures

# List of Acronyms

| | |
|---|---|
| AP | Average Precision |
| BC | Background Clutter |
| CF | Correlation Filter |
| CFT | Correlation Filter Tracker |
| DCF | Discriminative Correlation Filter |
| DEF | Deformation |
| DFT | Discrete Fourier Transform |
| FFT | Fast Fourier Transform. |
| FM | Fast Motion |
| FN | false negative |
| FP | false positive |
| FPS | Frames Per Second. |
| HOG | Histogram Of Gradient |
| IOU | Intersection Of Union |
| IPR | In-Plane Rotation |
| IV | Illumination Variation |
| KCF | Kernelized Correlation Filter |
| LR | And Low Resolution |
| LRR | Linear Ridge Regression |
| MB | Motion Blur |
| MOSSE | Minimum Output Sum of Squared Error |
| OCC | Occlusion |
| OPR | Out-Of-Plane Rotation |
| OV | Out-Of-View |
| PF | Particle Filter |
| PPI | Pixel Per Inch |
| QPF | Quantum Particle Filter |
| RGB | Red Green Blue |
| ROI | Region Of Interest |
| SOTA | State Of The Art |
| SR | Success Rate |
| SV | Scale Variation |

| TLD | Tracking-Learning-Detection |
|-----|------------------------------|
| TN | true negative |
| TP | true positive |
| VOT | Visual Object Tracking |

# Nomenclature

| | |
|---|---|
| $\boldsymbol{w}$ | the weight vector |
| $x_i$ | The input variable/samples |
| $\lambda$ | the regularizer |
| $P^u$ | the permutation matrix |
| $s$ | the number of |
| $C(\boldsymbol{x})$ | the circulant matrix |
| $\mathbb{R}$ | the set of real numbers |
| $\boldsymbol{A}, \boldsymbol{X}$ | the data matrix |
| $a_{ij}$ | the image features of a data matrix |
| $\boldsymbol{a_i}$ | the rows of data matrix $\boldsymbol{A}$ |
| $\boldsymbol{X}$ | the block circulant matrix |
| $G$ | the Gram matrix |
| $\boldsymbol{\alpha}$ | the vector of coefficients that defines the solution of linear ridge regression |
| $n$ | the number of samples in gram matrix |
| $m$ | the number of features in gram matrix |
| $\kappa$ | the kernel function |
| $K$ | the kernel matrix |
| $\hat{\boldsymbol{y}}$ | the regression labels in the Fourier domain |
| $F$ | the Fourier transformation function |
| $F^{-1}$ | the inverse Fourier transformation function |
| $\hat{\phantom{x}}$ | the symbol for Discrete Fourier Transform (DFT) for a vector |
| $\varphi$ | the function that maps/transforms any data points from input space to a higher dimensional feature space |
| $S$ | the object state of a bounding box |
| $\Delta_G$ | the set of ground truth annotation |
| $\Delta_t$ | the set of predicted annotation |
| $x_g$ | the top-left x co-ordinates of ground truth annotation |
| $y_g$ | the top-left y coordinates of ground truth annotation |
| $w_g$ | the width of ground truth annotation |
| $h_g$ | the height of ground truth annotation |
| $x_t$ | the top-left x co-ordinates of predicted annotation |
| $y_t$ | the top-left y coordinates of predicted annotation |
| $w_t$ | the width of the predicted annotation |
| $h_t$ | the height of predicted annotation |
| $t$ | the time |
| $R_t$ | the region of the object/target at any time $t$ |
| $C$ | the center of the object/target |
| $C_G$ | the center of the ground truth bounding box |
| $C_T$ | the center of the predicted bounding box |
| $\odot$ | the dot operator but element-wise multiplication |
| $\sigma$ | the standard deviation e.g. in Gaussian kernel |
| $\cap$ | the intersection operator |
| $\cup$ | the union operator |
| $ROI_{T_i}$ | the target bounding box in the $i$-th frame |
| $ROI_{G_i}$ | the $ROI_{G_i}$ is the ground truth bounding box in the $i$-th frame |
| $r_t$ | the minimum overlapping area |
| $\sum$ | the summation operator |
| $SR_{avg}$ | the average success rate |
| $P$ | the precision of tracking algorithm |
| $s$ | the state vector of particle filter |
| $w_k^i$ | the weight of the $i^{th}$ particle in particle filter |
| $x_{est_k}$ | the estimate of the particle filter algorithm at iteration $k$ |
| $\dot{x}$ | the velocity of the top-left coordinate of the target at $x$ axis |
| $\dot{y}$ | the velocity of the top-left coordinate of the target at the y-axis |
| $\alpha_t$ | the scaling factor in particle filter configuration set |
| $\sum_R$ | the covariance of the process noise in particle filter algorithm |
| $d$ | the similarity measure used in particle filter algorithm |
| $k(r)$ | the weighting function in particle filter |

# Chapter 1.

# Introduction

## 1.1. Motivation

Tracking can be considered as finding the minimum distance in feature space between the current position of the tracked object to the subspace represented by the previously stored data or previous tracking results. It's a process where, given the initial state of a target in the *first* frame, the goal of tracking is to predict states (e.g. position) of a target in a video. Visual tracking systems have, for a long time, employed hand-crafted similarity metrics for this purpose, such as the Euclidean distance [1], Matusita metric [2], Bhattacharyya coefficient [3], and Kullback–Leibler [4], and information-theoretic divergences [5].

This field has seen tremendous progress in recent years in robotics and surveillance applications. It is trying to address the issues caused by noise, clutter, occlusion, illumination changes, and viewpoints (e.g. in mobile or aerial robotics). There have been numerous attempts in designing and deploying a robust tracking method. However, achieving full tracking accuracy under realistic conditions presents various challenging scenarios. For example, it may detect and track the target correctly in isolation but may fail to continue to track in the crowd (crowded campus hallway or a road in an urban area during traffic time). The efforts to resolve these issues have borne fruits and led to work in various applications like automation [6], surgery [7], and surveillance and security [8] [9], [10]. However, there remain numerous challenges to achieve real-time performance with high speed coupled with a high degree of tracking accuracy. A typical scenario of visual tracking is to track an object such as a person initialized by a bounding box in subsequent image frames. Traditional approaches that rely on deterministic feature search have difficulty tracking a target when it is temporarily occluded or contaminated by the background, like in cases of a cluttered environment (see  Figure 1.1 ). Other challenges include appearance changes (of both target and the environment), changes in illumination, shape, pose, and color.

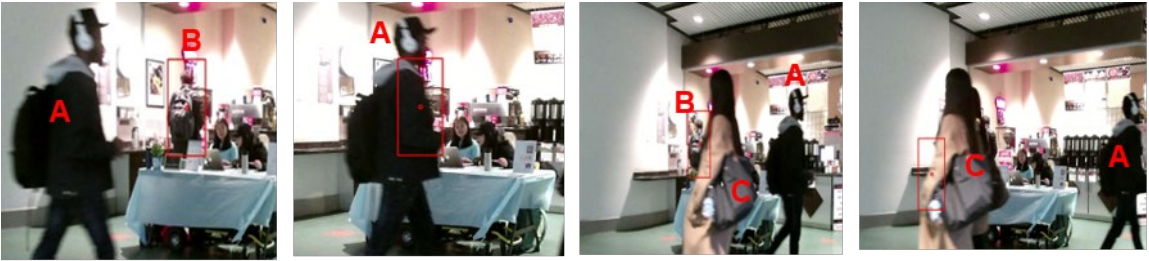

**Figure 1.1.**     **Visual tracking is erroneous when the target being tracked is occluded by the person. In the video frame, the person in the background (A) is the target. However, as soon as he is occluded by two people (people labeled B and C) passing by, the target (A) is no longer accurately tracked.**

The majority of the trackers discussed in the field of visual tracking can be categorized into a) generative tracker and b) discriminative tracker. The generative trackers [11]–[13] is an appearance-based tracker that focuses on the appearance of the object in interest. They look for similarity in appearance (like color, height, etc.) to identify the nearest possible match with the target object.  Discriminative trackers [14], on the other hand, do not consider any similarity of appearances or features, they, as the name suggests, differentiates the target from the background for tracking through classification.  The most significant contributions in the application of machine learning in visual tracking have been done using the discriminative trackers.

Despite existence of neural network based architectures, recent years have seen a significant shift in the attention towards trackers that learn "on-the-fly" i.e. approaches which model how an object varies visually over time, as and when new data becomes available. Many Discriminative Correlation Filter (CF) tracking methods have adopted this approach. Though neural network-based architectures (or deep learning methods) have shown good accuracy, they have significant disadvantages in terms of cost, training time and compute power: For example, XLNet model [15] costs $61,000 to train, uses 512 TPU v3 chips with batch size of 2048 (for comparison, an individual person or a small research lab normally uses batch size of 32 with normal compute) and  takes 2.5 days to train. Grover, a billion-parameter neural network costs $25k to train for two weeks on TPU v3.  Image classification models like EfficientNet [16] can take 20-23 hours to give results using google cloud compute. These training times including the high cost of GPUs and TPUs, do not necessarily lead to best results on every training run and hence can be a huge bottleneck of researchers who spend significant amount of time to fine tune the neural network. Reproducibility of deep neural networks Is another problem in the research community. With as many as a billion tunable parameters, it has become increasing difficult to replicate the performance claimed by existing models and research papers. Recently, the research community is encouraging researchers to follow reproducibility checklist [17] to ensure better use of neural network models, however, it is still an issue which has not been resolved. These issues have encouraged the visual tracking research community to look for faster and competitive alternative in correlation filter-based trackers. They offer solutions for real-time tracking with good real-time performance. Correlation filters have got significant attention because of their high frame per second (FPS), low computation power (they work significantly good with CPUs but can be made faster with few GPUs), and high efficiency.

One of the earliest works which proved its efficiency is Minimum Output Sum of Squared Error (MOSSE) [18] which could reach 700 frames per second (FPS). Since then, this area has seen development of various tracking methodologies like introduction of feature representation [19] [20], non-linear kernel [21], scale estimation [22], max-margin classifiers [23], spatial regularization [24], and continuous convolution [25]. Tracking evaluations like [26] [27] and benchmarks like [28] have shown that these trackers have greatly improved and significantly advanced state-of-art.

Despite their impressive performance, these discriminative learning tracking, for example, CFTs, have still few limitations. Firstly, they tend to drift while tracking over a long period of time. This drift can be more closely observed in a dynamic environment and while tracking multiple targets, thereby affecting its accuracy. Secondly, in such discriminative learning methods, negative samples of the image are as equally important as the positive samples. In fact, training with a higher number of negative samples (image patches from different locations and scales) has been found to be highly advantageous in better training of any tracking algorithm. However, any increase in the samples can lead to a higher computational burden which can adversely affect the time-sensitive nature of trackers. On similar lines, limiting the samples, however, can sacrifice performance, which is a trade-off.

Kernelized Correlation Filter (KCF) [21] talks about addressing this issue of handling thousands of sample data yet keeping the computation load low by exploring tools of kernel trick and properties of Circulant Matrices. Kernel trick is a way of computing the dot product of any two vectors **m** and **n** in some (possibly very high dimensional) feature space. In machine learning, a "kernel" is usually used to refer to the kernel trick, a method of using a linear classifier to solve a non-linear problem. Circulant matrices, on the other hand, allow us to enrich the toolset provided by classical signal processing like that of correlation filters, by working in the Fourier domain, making the process of training faster. Readers are advised to go to Chapter 2, Section 2.2 for more details.

Despite the increased efficiency brought by KCF, the current number and variety of datasets that can be used for real-time performance studies are inadequate. In the past, most of the experiments done on available state of art trackers (TDAM [29], MDP [30], DP-NMS [31], and NOMT [32] ), were done on sample videos. This observation is sometimes very significant since it is obvious that for real-life robotic monitoring and surveillance and due to various factors beyond such as environmental changes, the tracking accuracy depicted by online trackers may not reflect real-time scenarios. The limitations include using their validations only on few datasets (example videos) or shorter length of videos. To cover for the lack of needed datasets for given scenarios, we collected our own dataset in real time using the Microsoft Kinect sensor. This dataset reflects real time scenarios in diverse settings. Also, when testing for cases of occlusion, we collected datasets which are particularly suited for such scenarios to eb able to make better observations.

Despite practical applicability in tracking, a need for understanding KCF theoretically, mathematically, and experimentally using a real-time dataset exists [33], [34] . We refer readers to a comprehensive review of visual tracking using datasets in [35] and specifically using the correlation filter with kernel trick in [36]. Unlike other survey papers, this thesis explains the workings of the tracker, investigates its effectiveness, and provides supporting code and visualizations. We make observations on performance metrics and identify issues like short-term tracking ability etc. Finally, we compare KCF with the current SOTA on OTB-50, VOT-2015, and VOT-2019.

This thesis also proposes a novel architecture to make the KCF tracker more robust during occlusion by using additional depth information. This information along with RGB image information can not only help detect occlusions but also help in re-detection of the target, the two very challenging scenarios of real-time tracking.

We further explore another complementary tool to KCF for enhancing the robustness during the occlusion phase. It is based on locating the target in the dynamic region using particle filter. Hence, we experiment and showcase where particle filter can help the tracker and cases where it doesn't.

Finally, we conclude our thesis with discussions on the role of depth and particle filter in kernel correlation filter and their respective role in the robustness of the tracker.

Hence, this thesis provides a comprehensive understanding of the KCF tracker and proposes a novel method to implement a long-term tracker with depth data. This thesis also discusses the possibility of using particle filters in the tracker, its challenges, and possible solutions for the future.

## 1.2. Related Works

### 1.2.1. Literature Review on RGB Based Trackers

The visual tracking community has been developing RGB based tracking for a long time. Due to significant research in this field, many frameworks have contributed to increasing the tracking robustness and speed. Some of these frameworks include Kalman Filter [37] [38] and Particle filter [12] [39]. Since the invent of MOSSE [18] in 2010, correlation filter (CF) based trackers gained huge popularity owing to their speed and accuracy.

Correlation Filters are a class of classifiers, which use a designed template to produce sharp peaks (strong correlation) in the correlation output. The peaks correspond to the precise localization of the object/target in scenes. A general framework of correlation filter-based tracking methods can be summarized as follows. A patch is cropped from the target location and used for feature extraction. Features e.g. raw features, HOG features, color features are extracted from this patch. A cosine window is then applied over these features to remove the boundary effects (distortion of features at the edges). Subsequently, efficient correlation operations are performed by replacing the exhausted convolution with element-wise multiplication using Digital Fourier Transform (DFT). The location of the highest correlation response corresponds to the predicted target location in the next frame. This process is iterated for all the frames and at every step. One of the early CF tracker MOSSE [18] was trained on grayscale templates and had an FPS of ~700 making it one of the breakthroughs works in the field of DCF. Since then various improvements have been made to improve the tracker.[19] and [40] introduced new feature representation by proposing multi-dimensional features, to make the tracker learn better. Henrique *et.al.* proposed a kernelized version of the CF tracker [41] which benefitted from the circulant nature of the samples. Scale adaption (when target size changes e.g. when it moves back and forth) of the correlation filter tracker was investigated by [42], [43] who applied correlation filter to scale space. Other improvements include spatial regularization in SRDCF [24], continuous convolution in C-COT [25], max-margin classifiers in [44]. Works like [45] showed a connection to spatio-temporal learning in the image content in these trackers. Most of these trackers did not explicitly account for deformation, part-based occlusions, and model drifts. These issues were addressed in methods proposed using part-based features [46] to make the filter robust. Works like [47] attempt to combine the part-based features and particle filter in correlation filter tracking and get a performance gain of around ~6% in terms of area-under-curve (AUC). Part-based trackers have a disadvantage of computing a high number of parameters, an issue which was addressed in [48] and [49] where the tracker architecture was decomposed to global and local appearance layers. These developments have established that CF-based trackers are significant contributions to visual tracking research. However, recent improvements in CF based trackers have come at a cost of speed and real-time performance. For example, Discriminative Correlation Filter (DCF) [24] using HOG features reached ~6 FPS as compared to some early state-of-arts like KCF [41] which attained ~170 FPS and MOSSE [18] which was ~700 FPS.

Another popular framework is Siamese-based networks which became hugely popular when the results of Siamese based CNN architecture [50] showed great performance in the VOT-2015 challenge [51]. It also established that deeper and wider networks can improve target representation. Later various deep CNN-based models were proposed. Some of these methods addressed the issue of target scale estimation by predicting segmentation masks rather than bounding boxes [52], [53].

Overall, several efforts have been made to address where challenging issues of visual tracking. However, target appearance, if used as the main cue for tracking, is not a very reliable feature when the target suffers from challenging issues of occlusion, out-of-view, and illumination changes. Features like depth data, with its ability to distinguish between foreground and background, can help in making the tracker more accurate. Trackers have been developed in the past which use RGB features augmented with additional features like depth. It is important to discuss how depth can act as a complementary cue and make the tracker more robust.

## 1.2.2. Literature Review of RGB- Depth Based Trackers

Visual tracking combined with depth data has become increasingly popular in recent years since they provide additional information in the form of depth features needed to prevent model drift and model failure in the tracker. It was the availability of low-cost depth sensors e.g. Kinect, Asus, and Intel Realsense depth sensors that allowed the research community to fully exploit the abilities of the sensor. These sensors have different effective ranges and can be used as per the requirements of the desired range. For example, Microsoft Kinect V2 can sense object between 0.5m and 4.5m, Asus XtionPro Live has a range of 0.8m to 3.5m and Intel RealSense Depth Camera D455 has a range from 0.4m to 20m. If the subject moves farther away out of the effective range of the sensor, the quality of the image deteriorates, and the depth accuracy becomes more scattered (i.e. less reliable).

One of the first works that led to the popularity of the depth sensors was the early work by Song *et. al.* [54] where the authors released a large public dataset (Princeton RGB-D dataset) that had both RGB and depth information of the targets. They evaluated the performance of the target model by computing HOG features on both color and depth information. This work used, exhaustive search, optical flow computations, and elaborate color and depth segmentation. It was demonstrated to have a performance of (~0.65 FPS) which outperformed state-of-art RGB trackers proving its effectiveness. Similar work was also proposed in [55] where the authors extended the work of Tracking-Learning-Detection (TLD) [56] and added depth information to filter background pixel. This made the tracker more robust to occlusions and attained performance of ~10 FPS. Since then, various works have focussed on using depth information in addition to other available features. [57] proposed using depth in 3D object tracking algorithm based on the point cloud. They explore the global features of the RGB-D image and convert it to the point cloud (extracted by PointNet) and finally integrate them into the object tracking, to solve the problem of occlusion. Issues of occlusions and scale changes are also addressed in [58] [59] which fuse color and depth cues as features.

CF base trackers like KCF have demonstrated the significance of correlation filter-based trackers. Several works build upon KCF by adding depth information to the RGB data to make the tracker more accurate. [58] build upon color only KCF tracker and adds depth showing a real-time performance of ~35 FPS. [59] proposes a distractor-aware learning method (DLS) with RGB-D data to effectively alleviate the model drift problem.

Some of the best RGB-D trackers provide high accuracy but lag in speed. On the contrary, state of the art in RGB tackers can provide high speed but are clearly inferior to RGB-D trackers in terms of success rate (% of correct predictions). It would be interesting to explore an improved CF tracker which can address some of the existing issue related to robustness e.g. occlusion, model drift, scale changes, color camouflage, etc., yet achieving higher accuracy. Few works have attempted to address this gap of speed in RGB-D trackers. [60] proposes a deep depth-aware long-term tracker that extends deep discriminative correlation tracker (DCF) to embed depth information to deep features. It achieves state-of-the-art RGB-D tracking performance and has better speed performance. Closing the gap between speed and accuracy in an RGB-D tracker is an ongoing problem.

## 1.3. Contribution of Thesis

Motivated by the problem that the feature cues of an RGB image are not sufficient to implement a long-term tracker and that depth data provide additional contextual information (information for a tracker to distinguish between foreground and background), we suggest that RGB-D based kernel correlation tracker will be more accurate and robust to occlusions and out-of-view scenarios. We further study if RGB-D based long-term tracker can be further improved using particle filter framework since it can provide additional information to localize the target to make accurate predictions.

As a result, we propose a.) a novel tracker architecture: for RGB-D based KCF tracker and b.) study of Particle Filter based RGB-D tracker to validate our hypothesis on standard datasets as well as real-time datasets.

The main contributions of this thesis can be summarized as follows:

a) An overview of the KCF tracker which talks about the working of the tracking without going into mathematical details. This is important for the reader to get an intuitive idea of the tracker and its underlying concepts.

b) An In-depth mathematical discussion in Section 2.2 and Section 2.3 which goes into the fundamental issues and challenges associated with the KCF tracker using visualization tools. Each algorithmic detail has been explained step-by-step and has been accompanied by relevant visual analytics, block diagrams, or flow charts. To the best of our understanding, this is the first tutorial work that explains through a more intermediate exposition of the algorithm with its detailed experimental evaluation.

c) Systematic experimental evaluation and performance analysis as shown in Section 3.1 in real-time scenarios under various challenging scenarios like that of illumination, occlusion, speed, etc. The tracker has also been compared with other algorithms on OTB-50 and VOT datasets.

e) Development of a robust correlation-based long-term tracker using depth data. This tracker has the ability to re-detect the tracked target. Section 4.2 details the evaluation results of the proposed tracker on a standard benchmark and user-collected data.

f) Study to improve the performance of the KCF-based tracker by exploring additional features, for example, particle filter, to improve the KCF tracker. The results and observations of this study can be seen in Section 5.2

## 1.4. Layout of Thesis

This thesis is organized into six sections.

• Chapter 1 is the introduction which discusses the motivation of the research undertaken by the author. It is followed by a literature review of various RGB and RGB-D based tracker, their advantages and places of potential improvement. It ends with a brief explanation of the contribution of this thesis.

- Chapter 2 gives an intuitive overview of the RGB based Kernel Correlation Filter tracker. It does not go into any deep mathematical details but explains the terms and concepts from a general understanding. This overview lays the foundation of deep mathematical explanations which has been explained later in the chapter. It is followed by step by step experimental explanation of the chapter using a visual example.

- Chapter 3 builds upon the knowledge of the RGB-based KCF tracker in the preceding chapter and discusses the tracker performance. This chapter discusses how well the KCF tracker performs on a.) standard benchmark dataset and b.) in real-time. A detailed analysis is followed where the author tests the tracking performance of KCF on various evaluation metrics and discuss its strength and shortcoming. Towards the end of the chapter, the author compares its tracking performance with various other trackers on a standard dataset to see the tracker's current standing when compared to other trackers.

- Once we have discussed the tracker's fundamental concepts, its strengths, and shortcomings, readers are introduced to Chapter 4 where we attempt to address few weaknesses of the KCF tracker – the inability to track a target in case of occlusion, out-of-view, and model drifting. This chapter introduces an RGB and Depth based Kernel Correlation Filter which is robust to occlusion and model drift and can re-detect target when the target comes out of occlusion. The chapter ends with an experimental evaluation of the proposed RGB-D tracker on the standard benchmark and dataset collected by the author.

- In Chapter 5, we discuss the potential of adding particle filter as another layer in the dynamic model to track the target in our existing RGB-D based long-term tracker. This chapter discusses the implementation detail of the proposed framework and evaluates the tracker experimentally. It ends with a discussion on the tracking performance of the particle filter-based tracker and demonstrates its strength and weakness.

- The thesis ends with Chapter 6 where we make final remarks on the current status of the KCF tracker, our observation, and the advantages of the novel tracking frameworks we proposed. We also discuss future possibilities to improve the tracker using additional features.

# Chapter 2.

# An Overview of Kernel Correlation Filter (KCF)

There are not many recent works that have surveyed correlation filter tracking techniques. The most recent work is [61] which focuses on the background and current advancement of correlation filter-based algorithms in object tracking. [36] surveys recent development and improvements in CFTs and summarizes their general framework. Similar to this work, our work also discusses the correlation tracking methodology and experiments on standard datasets. We also discuss various challenging scenarios that affect the tracking performance.

However, this chapter differs from others in a way that none of the surveys mentioned above provide such an in-depth explanation of the workings of the tracker. Additionally, readers can benefit from understanding how the mathematical workings of various parts of the tracker can be translated into code and visualized for step by step understanding.

This chapter is organized as follows: Section 2.1 gives a general overview of the Kernel Correlation Filter followed by a detailed mathematical explanation in Section 2.2. This is followed by Section 2.3 where further explains the workings of the tracker experimentally using a real-time data sample (collected by the author). This section demonstrates visually several intermediate steps of the tracker for better understanding. It concludes with Section 2.4 where we discuss how the research community can benefit from these in-depth explanations.

## 2.1. Kernel Correlation Filter

The process of perceiving the physical properties (color, depth, etc.) of an object by our eye and applying semantics to it is a highly complex task. However, the human eye has the innate ability to perform this task in split seconds. This ability, known as visual recognition, also includes the process of understanding the object, previous experiences associated with this object, and its relation to the surroundings. Neurological science and cognitive science have done several pieces of research and developed various theories on understanding the process of human perception. The works go back as early as the 90s, some of the notable ones being hierarchical learning approaches [62] [63] [64]  and the effect of prior knowledge [65]. The former is based on the idea that an image ("scene") can be parsed into hierarchical levels much like a paragraph can be broken down into sentences, phrases and words. The latter talks about how human evolution and its interaction with the surroundings has contributed to a priori knowledge which helps humans perceive objects in a better way.

These ideas and the understanding of the hierarchical approach and effect of a priori knowledge on human visual perception have inspired various modern computer vision algorithms [66], [67]. The importance of certain prior knowledge about the object, here 'image', can help the algorithm discard redundant or unimportant computations. Few examples would be how HOG (histogram of oriented gradients) is used in [68] instead of pixel information thereby making the algorithm illumination invariant, [37] uses sliding window instead of one whole image making it translations invariant, [69] uses depth feature in addition to RGB making the system more robust to color camouflage. Other prior knowledge that can be incorporated into the system could include scale invariance, rotation invariance, etc. The approach for choosing the optimal parameter as well as the process of learning affects how objects are being recognized and detected by the algorithm.

Visual recognition paves way for another very important aspect of research called visual tracking where the object of interest is located over a sequence of time using a camera. Such research has been complimented time and again by tracking using detection. Modern computer vision research has developed several algorithms that use visual detections for tracking, notable ones being: a) finding interest points, followed by Hough voting [70] [71]  and b) sliding window detector, where we slide a box around an image looking for the object and classify the image crop in the box (if the object is found). The idea of using a sliding window is not new and has been consistently used in some of the earlier works [72]. In 2001, Voila-Jones Detector [73] introduced a novel idea of using a sliding window for integral images (in an integral image the value at pixel $(x, y)$ is the sum of pixels above and to the left of $(x, y)$ before performing feature search, making the computation comparatively faster). The use of sliding window helps in searching the scope of an entire image for features or other vectors. This idea of using sliding window was later utilized in HOG  Detector [74] originally used for person detection. It was also used in Deformable Part-Based Model [75], originally developed for face and human detection which is a brute force or exhaustive search method used to localize objects of a certain class across the entire image within a collection of localized windows. Despite various efforts like Viola-Jones Detector [73] and efficient descriptors like HOG [74], the large number of samples traced by the sliding window always led to higher complexities. In 2010, Yichen Wei proposed an efficient method on the histogram-based sliding window [76] which can result in a constant complexity. They achieved this by taking advantage of the spatial coherence of natural images and incrementally computing the objective function. This approach continues to be used today in works like Deep Sliding Shapes [77] in 3D convolution architectures to create a 3D bounding box of the target object.

A successful attempt in implementing a computationally fast tracking-by-detection algorithm using a sliding window came in 2015 when researchers developed Kernelized Correlation Filter (KCF) tracking algorithm [21]. *Tracking-by-Detection* breaks down the task of tracking into detection-learning and then tracking where tracking and detection complement each other i.e. results provided by the tracker are then used for the algorithm to learn and improve its detection. Oher recent works include multi-sensor 3D tracking [78], adaptive visual object tracking [14], and visual tracking which are robust to color changes and deformations [79] KCF differs from the ones mentioned above in a way that it performs better with respect to computational complexity, speed, and accuracy. KCF leverages a certain structure (explained later) of the image data. This structure is useful when the subject being tracked, undergoes displacement in the subsequent frame, allowing the 'structure' to be similar (because of similarity in image properties) in both frames. It further utilizes a classical signal processing technique, fast Fourier transformation, in order to make the tracking algorithm computationally very efficient which is well suited for real-time applications. KCF is a semi-supervised learning algorithm since the location of the target (region of interest) is provided in the first frame. However, region of interest can also be intelligently located using various other methods like change detection [80] and background subtraction [81]. Since, our proposed tracker is based on KCF, we use the semi-supervised approach similar to KCF, for tracking in our proposed modified architectures.

KCF uses a large number of (all the possible) translations of an image patch (window size of the ROI in the entire image). These image patches can be extracted by providing the dimensions of the window size to the image at the time of computation.

KCF uses these patches, their extracted features (e.g. HOG), and computation done on this data (features) to detect the location of the target in the subsequent frame of the image. This detection at subsequent frames can eventually be stored and used to track the target over time. It is to be noted that there is a good rationale behind using this large amount of stored data in KCF. This data is relevant for any learning algorithm because it represents various ways in which the samples (here features) can be encountered by the learning algorithm, thus making the algorithm more robust. However, if not used efficiently, this data (high dimensional features), can also become a bottleneck in the case of real-time processing. If we can, somehow, represent the high dimensional feature space in a form that captures the translational properties of the object between the frame and still accomplish the learning using a linear model (using some algorithmic trick which will be discussed later), we can reduce the computation significantly. A linear model is preferred because it gives us a simpler linear discrimination relationship between the classes in the feature space, any independent and associated dependent variable. The above-mentioned ways also bring with themselves the certain advantage of element-wise multiplication in the Fourier domain and their resemblance to fast correlation filters, making computation faster (discussed in Section 2.2).

## 2.2. Mathematical Exposition of Kernel Correlation Filter

For a better understanding of the mathematical representations, authors feel the need to make certain clarification in the representations. $x$ and $x'$ will be used as independent variables with no relation to each other. Transpose of $x$ will be defined as $x^T$. The authors will use bold fonts in a variable to represent vectors. For example, $\boldsymbol{x}$ represents a concatenated vector with each element /entry in the vector represented as $x_i \in \mathbb{Z}$ i.e. in space of all integers ($\mathbb{Z}$).

KCF is based on the discriminative method, which formulates the tracking problem as a binary classification task and distinguishes the target from the background by using a discriminative classifier [82]. Any modern framework, in addressing recognition and classification problems, in the field of computer vision, is usually supported by a learning algorithm. The objective is to find a function $f$ which can classify through learning from a given set of examples. It should be able to classify an unseen data representation of an image patch from an arbitrary distribution within a certain error bound. One such linear discriminator is Linear Ridge Regression (LRR).

Mathematically it is defined as:

$$\min_{w} \sum_{i}^{n} (w^T x_i - y_i)^2 + \lambda \|w\|^2 \tag{1}$$

where $x_i$ is the input variable (here image feature), $w$ is the weight vector, $(w^T x_i - y_i)^2$ is squared error between the actual and predicted variable, $y_i$ is the desired prediction and $\lambda \|w\|^2$ is a squared norm regularizer with $\lambda$ as the regularizer to prevent overfitting. In simple words, in determining the weight vector $w$ for minimizing the square of the model error $(w^T x_i - y_i)^2$, it is bounded to a certain limit with the help of a regularizer $\lambda$.

The above step is more an overview of the learning step i.e. provided with learning with some limited number of samples $(x_i)$, how well can it classify the unseen data? Traditionally, researchers have used positives samples (samples that are similar to the target object) for learning, however, it has been found negative samples (samples that don't resemble the target object) are equally important in discriminative learning. Hence, the more the number of negative samples, the better the discriminative power of the learning algorithm. If we can use a higher number of features (and as a result more negative samples), it will be beneficial for the learning algorithm. Though effective in helping with improved accuracy; the availability of a large number of samples, can create a bottleneck in terms of complexity and speed.

Kernelized Correlation Filter tracker is an attempt to use these large numbers (high dimensional data) of negative samples, however, with reduced complexity. It achieves this using: a) the dual space for learning high dimensional data using kernels trick and b) the property of diagonalizing the circulant matrix in the Fourier domain. It makes the tracking algorithm computationally inexpensive and faster respectively. We will discuss the concept of circulant matrix, kernel trick, and Fourier domain in more detail in upcoming sections.

Circulant matrix derives their properties from mathematical science (we point the readers to [83] and [84] which have discussed the role of circulant matrices in mathematics). They have been explored in varied different contexts like sensor placement and motion coordination in visual tracking [85], fast algorithm for reconstructing signals from incomplete Toeplitz and Circulant measurements for compressive sensing [86], acoustic noise cancellations [87], and in some image processing applications [88]. A detailed review of Circulant Matrices is provided in [89]. Since the work of Henriques et al [90] on the use of Circulant Matrix, several other works have explored its use case on classical methods like correlation filters [91], [92], and even deep learning [93]. KCF benefits from the use of Circulant Matrix since it can be used to generate a large number of negative samples. We will now discuss the structure and the idea behind the Circulant Matrices.

The structure of the circulant matrix is very simple to understand. Consider an image base patch (for simplicity, we will assume a 1-D image), a vector $x \in \mathbb{R}^s$ with elements $x_i$ where $i \in \mathbb{Z}$. It is possible to use this base image patch vector $x$ to generate additional samples of $x$ (using all the possible translations of it) by using a permutation matrix $P^u | u = 0,1,2, \dots s$. For simplicity, let's say, $s = 3$, then we can define the permutation matrix $P^u$ as shown:

$$P = \begin{bmatrix} 0 & 0 & 1 \\ 1 & 0 & 0 \\ 0 & 1 & 0 \end{bmatrix} (= P^1) \tag{2}$$

Let us define 1-D vector $x \in \mathbb{R}^s$ such that $x = [\ x_1 \ x_2 \ x_3]^T$. The permutation matrix $P^u$ has the effect of permuting elements of a vector, here $x'$ , in a cyclic shift as shown in Table 2.1:

**Table 2.1.** **The table shows the effect of the permutation matrix is a 1-D vector for three permutations**

| | $P^u \ (u = 1,2, \dots s)$ | | $x$ | $P^u x$ |
|---|---|---|---|---|
| $P^1 x =$ | $\begin{bmatrix} 0 & 0 & 1 \\ 1 & 0 & 0 \\ 0 & 1 & 0 \end{bmatrix}$ | | $\begin{bmatrix} x_1 \\ x_2 \\ x_3 \end{bmatrix} =$ | $[x_3 \ \ x_1 \ \ x_2]^T$ |
| $P^2 x =$ | $P^2 = \begin{bmatrix} 0 & 1 & 0 \\ 0 & 0 & 1 \\ 1 & 0 & 1 \end{bmatrix} (= P^1 . P^1)$ | | $\begin{bmatrix} x_1 \\ x_2 \\ x_3 \end{bmatrix} =$ | $[x_2 \ \ x_3 \ \ x_1]^T$ |
| $P^3 x =$ | $P^3 = \begin{bmatrix} 1 & 0 & 0 \\ 0 & 1 & 0 \\ 0 & 0 & 1 \end{bmatrix} (= P^2 . P^1)$ | | $\begin{bmatrix} x_1 \\ x_2 \\ x_3 \end{bmatrix} =$ | $[x_1 \ \ x_2 \ \ x_3]^T$ |

Since at each cycle the resultant vector is of size $s$, the power will be periodic i.e. for $s = 3$, after every three permutations, we will get the same vector. Mathematically, $P^{3n+1} = P^1$ for $n \in \mathbb{Z}$ or in other words, we will get the same vector $\boldsymbol{x}$ for every s uniquely shifted version. Also, it is observed that $P^3 = P^0 = I$. The transpose (shown in the last column of Table 2.1) is only necessary to turn a column vector into a row. Now, the vectors in the last column of Table 2.1 together form a data matrix of the form:

$$\boldsymbol{X'} = \begin{bmatrix} [P^0(\boldsymbol{x})]^T \\ [P^1(\boldsymbol{x})]^T \\ [P^2(\boldsymbol{x})]^T \end{bmatrix} = \begin{bmatrix} x_1 & x_2 & x_3 \\ x_3 & x_1 & x_2 \\ x_2 & x_3 & x_1 \end{bmatrix} \tag{3}$$

(Gentle reminder that $x$ and $x'$ will be used as independent variables. Transpose of $x$ will be denoted by $x^T$) This form of the data matrix is called Circulant Matric $C(\boldsymbol{x})$ and it contains all the cyclically shifted versions of $\boldsymbol{x}$. Numerically it can be represented as shown in Figure 2.1

### 6. Final Circulant Matrix

```
X_matrix = [P1x' ; P2x' ; P3x']
```

```
X_matrix = 3×3
        9     7     8
        8     9     7
        7     8     9
```

**Figure 2.1.** **Creating circulant matrix using permutation matrix as well as MATLAB inbuilt function verifies our calculation, demonstrated in circular_shifts_permutation_live.mlx[1]**

Mathematically, $C(x)$ can now be defined as:

$$C(\boldsymbol{x}) = \{P^{u-1}\boldsymbol{x} \mid u = 1,2,\dots s\} \tag{4}$$

Hence, using solutions obtained in Equation (3) and Equation (4), we can now define our circulant matrix for $\boldsymbol{x} = [\ x_1\ \ x_2\ \ x_3\ ]^T$ where $\boldsymbol{x} \in \mathbb{R}^3$ as:

---

[1] The code files mentioned can be found at https://github.com/copperwiring/KCF_tutorial_code

$$C(\boldsymbol{x}) = \{P^{u-1}\boldsymbol{x} \mid u = 1,2,3,\dots s\} = \begin{bmatrix} (P^0\boldsymbol{x})^T \\ (P^1\boldsymbol{x})^T \\ (P^2\boldsymbol{x})^T \end{bmatrix} = \begin{bmatrix} x_1 & x_2 & x_3 \\ x_3 & x_1 & x_2 \\ x_2 & x_3 & x_1 \end{bmatrix} \tag{5}$$

An equivalent visualization for the 1D image vector can be plotted using $BCM\_1D\_live.mlx$[2]. The visualization would look as shown in Figure 2.2.

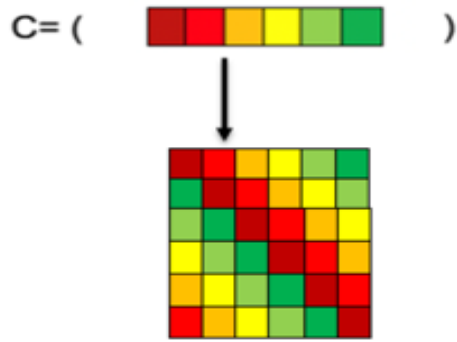

**Figure 2.2.** **Visual Representation of Circulant Matrices. plotted using BCM_1D_live.mlx file**

The KCF algorithm exploits this circulant structure in the learning algorithm by using a sample 2D image patch of the target and generating additional samples (using all the possible translations of it leading to circulant structure). This results in several virtual negative samples which can be used in the learning stage to make our discriminative classier efficient. Though the above mathematical explanation was for a 1-D image, this can also be generalized for a 2-D image. In 2D images, we obtain circulant blocks using a similar concept of circulant blocks. Let us assume we have a 2D image represented as a data matrix in the following form:

$$A = \begin{bmatrix} a_{11} & a_{12} & a_{13} & a_{14} \\ a_{21} & a_{22} & a_{23} & a_{24} \\ a_{31} & a_{32} & a_{33} & a_{34} \\ a_{41} & a_{42} & a_{43} & a_{44} \end{bmatrix} \tag{6}$$

[2] The code files mentioned can be found at https://github.com/copperwiring/KCF_tutorial_code.

where $a_{ij}$ represents the image features. Now, construct four rows of $\boldsymbol{A}$ denoted by $\boldsymbol{a_i}$ as below:

$$\boldsymbol{X_0} = circ(\boldsymbol{a_0}) = circ(a_{11}a_{12}a_{13}a_{14}) \tag{7}$$

$$\boldsymbol{X_1} = circ(\boldsymbol{a_1}) = circ(a_{21}a_{22}a_{23}a_{24}) \tag{8}$$

$$\boldsymbol{X_2} = circ(\boldsymbol{a_2}) = circ(a_{31}a_{32}a_{33}a_{34}) \tag{9}$$

$$\boldsymbol{X_3} = circ(\boldsymbol{a_3}) = circ(a_{41}a_{42}a_{43}a_{44}) \tag{10}$$

$$\boldsymbol{X'} = \begin{bmatrix} X_0 & X_1 & X_2 & X_3 \\ X_3 & X_0 & X_1 & X_2 \\ X_2 & X_3 & X_0 & X_1 \\ X_1 & X_2 & X_3 & X_0 \end{bmatrix} \tag{11}$$

This structure $\boldsymbol{X'}$ is called Block-Circulant Circulant Matrix (BCCM), i.e., a matrix that is circulant at the block level, composed of blocks themselves circulant. For each block of size 2 x 3 (for clear visualization), we can get visualize the block circulant matrix as shown in Figure 2.3

```
 0.5939  -1.3270   0.4018  -2.0220   0.6125  -1.1187   1.5163   1.6360   0.5894  -0.3268   0.5455   0.3975
-2.1860  -1.4410   1.4702  -0.9821  -0.0549  -0.6264  -0.0326  -0.4251  -0.0628   0.8123  -1.0516  -0.7519
-0.3268   0.5455   0.3975   0.5939  -1.3270   0.4018  -2.0220   0.6125  -1.1187   1.5163   1.6360   0.5894
 0.8123  -1.0516  -0.7519  -2.1860  -1.4410   1.4702  -0.9821  -0.0549  -0.6264  -0.0326  -0.4251  -0.0628
 1.5163   1.6360   0.5894  -0.3268   0.5455   0.3975   0.5939  -1.3270   0.4018  -2.0220   0.6125  -1.1187
-0.0326  -0.4251  -0.0628   0.8123  -1.0516  -0.7519  -2.1860  -1.4410   1.4702  -0.9821  -0.0549  -0.6264
-2.0220   0.6125  -1.1187   1.5163   1.6360   0.5894  -0.3268   0.5455   0.3975   0.5939  -1.3270   0.4018
-0.9821  -0.0549  -0.6264  -0.0326  -0.4251  -0.0628   0.8123  -1.0516  -0.7519  -2.1860  -1.4410   1.4702
```

```
imagesc(C);
```

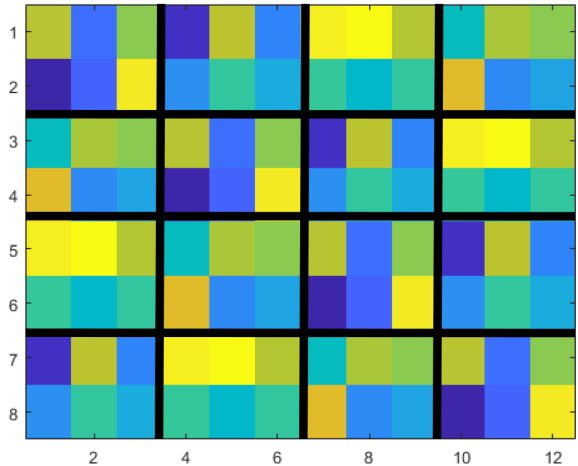

**Figure 2.3.**   **Visualization of block circulant matrix plotted using BCCM_2D_live.mlx file**

One can visualize a circulant matrix using random sample values (where 2 x 3 is the size of each individual matrix block and we want a 4 x 4 circulant pattern) from the file $BCCM\_2D\_live.mlx$[3] as shown in Figure 2.3.

Visualization of a BCCM on a sample 2D image patch of the target taken from Kinect RGB camera (Appendix A) can be seen in Figure 2.4. This visualization can be plotted using the MATLAB file $block\_circulant\_kinect\_image\_live.mlx$

---

[3] The code files mentioned can be found at https://github.com/copperwiring/KCF_tutorial_code

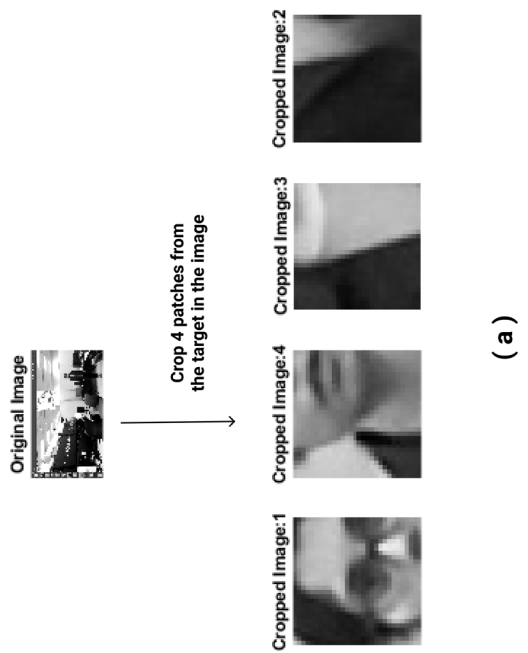

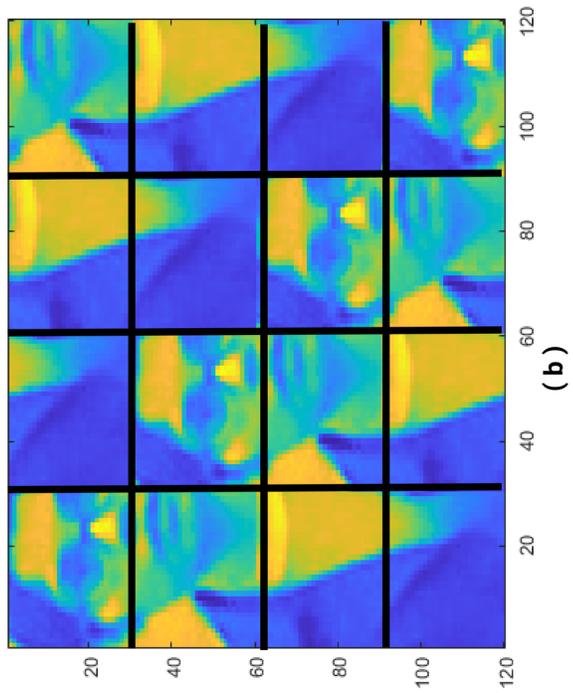

**Figure 2.4.** **(a) sample 2D greyscale image and the cropped samples of the target patch. The dimension of each of the cropped sample is 30x30 (b) Color-coded Block-Circulant Circulant Matrix for the cropped 2D image patch shown in (a) as visualized using block_circulant_kinect_image.mlx**

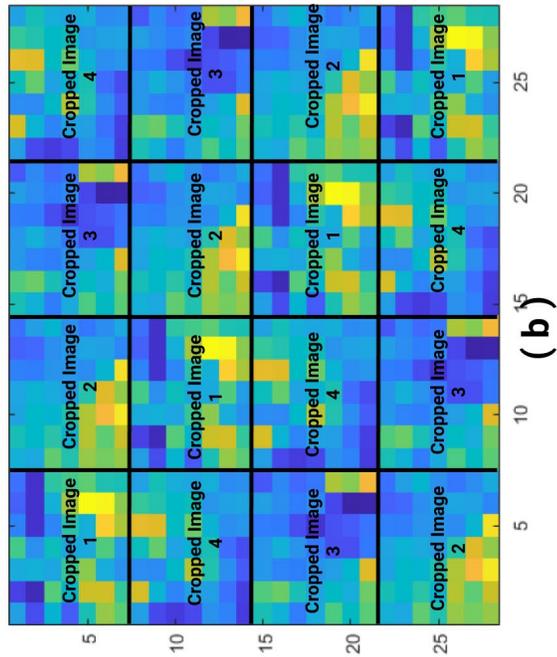

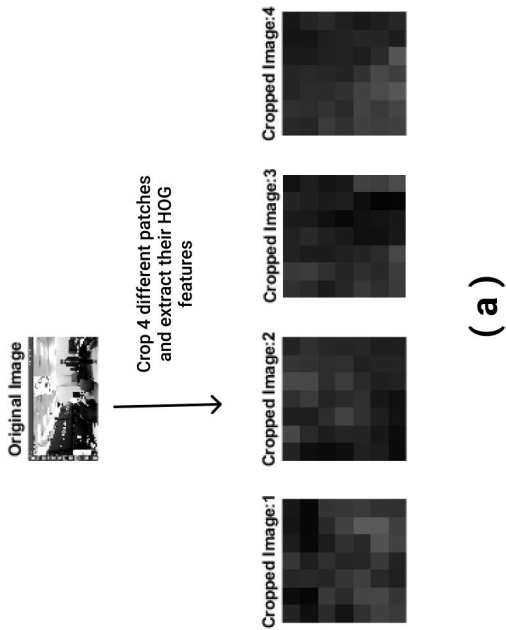

**Figure 2.5.** **(a) HOG features of the cropped samples of target patch from an image frame. For convenience, we extracted the HOG features from the same grayscale cropped patches as shown in Figure 5 (a). It is to be noted that we chose one (out of 32) dimensions[4] of the HOG features for simplicity in visualization (Best viewed in color). The dimension of HOG features for each of the four samples shown is 7x7 per orientation  (b) HOG features as color-coded circulant blocks as visualized using block_circulant_kinect_image.mlx. The edges and labels have been added for better clarity**

A similar visualization can be done on features, like HOG features, of the target image patch we used in our visual tracking experiment, as shown in Figure 2.5. Since we know that a) a large amount of data is beneficial for our learning algorithm and b) translating samples in a circulant manner creates a circulant structure of a large amount of data; we can generate and exploit the circulant nature of, for example, HOG features for the tracking advantage. Hence, we can now define $C(x)$ as a very high dimensional data which represents all the possible translations of the base sample HOG features (in case of 1D image) or base patch HOG features (in case of 2D image). Such a data matrix (with shifted translations) is relevant for a learning algorithm because it represents various ways (distribution of inputs) in which the samples can be encountered by the learning algorithm. These samples are called virtual samples used in the training.

Now, with features in hand, we can use our learning algorithm (e.g. linear ridge regression (LRR) (Equation 1). The goal of the objective function is to find a weight $w$ which will help in classifying the data. Mathematically, the optimal solution of LRR is given by:

$$w = (X^T X + \lambda I)^{-1} X^T y \qquad (12)$$

where $X$ is a data matrix (also called design matrix). In our case, the data matrix is the circulant matrix. If $\in \mathbb{R}^{nxm}$ , we have $n$ examples (samples) and $m$ features as shown in Equation 13.

$$\begin{bmatrix} x_1^1 & \cdots & x_m^1 \\ \vdots & x_j^i & \vdots \\ \vdots & \ddots & \vdots \\ x_1^n & \cdots & x_m^n \end{bmatrix} \qquad (13)$$

where each row represents one example (samples) and column represent one/many features associated with that sample i.e. $x_j^i$ implies $i^{th}$ example and $j$th feature. If there is one feature, there will be only one column ( $j = 1$.). Note, from Equation (12), we must take the inverse of $(X^T X + \lambda I)$ which is an $m\ x\ m$ matrix ($X$ is $n\ x\ m$ matrix, so $X^T X$ is $m\ x\ m$). In most cases, $m\ x\ m$ i.e. the feature space is very large making the computation very expensive. It would be helpful if we can somehow reduce the computation complexity of $m \times m$ matrix. This problem can be addressed by solving

Linear Ridge Regression in dual space. Mathematically, the dual solution for Linear Ridge Regression can be written (Appendix B) as:

$$\alpha = (G + \lambda I)^{-1} \boldsymbol{y} \qquad (14)$$

where $\alpha \in \mathbb{R}^n$ is a vector of coefficients that define the solution. Here $G = XX^T$ ($X$ is data matrix) is called the Gram Matrix. Gram Matrix will be helpful later. In general, for any two pair of samples $(x_i, x_j)$, Gram Matrix can be written as:

$$G_{ij} = x_i^T x_j \qquad (15)$$

Hence for $X = n \times m$ where $n$ is the number of examples (samples) and $m$ is the number of features, dual space allows us to compute the inverse of $n \times m$ dimensional data (which returns an $n \times n$ dimension data), unlike $m \times m$ in case of equation (12). It is advantageous in scenarios where $n \lll m$. It is also helpful when, with the increase in the size of the base image sample/patch, the size of this data matrix will increase quadratically; hence any computation done with this high dimensional data will adversely affect the primary goals of any tracking algorithm such as computational power and time complexity.

In the linear learning model (i.e. dual LRR) the objective is to find a hyperplane that can linearly separate our training data. However, there is an issue. Our preferred feature space i.e. HOG is non-linear in nature (our implementation of HOG has high (32) feature dimensions[5] which add to its non-linearity). One way to resolve this issue is to investigate some non-linear hyperplane pattern classifiers (i.e. higher-order surfaces). This is not a practical approach in comparison with a simple linear one. Also, as discussed before, the linear approach can be relatively inexpensive using dual space formulation. Hence, if we want to use our linear techniques on non-linear features, we will need to somehow transform our features from a non-linear space to a linear one. There are several such transformation techniques, one such being the powerful kernel trick [94] and their efficient applications in correlation filter based tracking [90], [95]. By applying the kernel trick on non-linear feature space, we would be able to do the computation in the non-linear feature space without explicitly instantiating a vector in the space. We will now explore the kernel trick in more detail.

---

[5] https://pdollar.github.io/toolbox/channels/fhog.html

Mathematically, many non-linear data points can be linearly separable if we can map the original data points to some other *feature space* using a transformation, say $\varphi$. Hence, it is expected that, in this new *feature space* (a high dimensional space), the points will be linearly separable. In an ideal case, we would want the dimensions to be infinite. Unfortunately, higher dimensions bring with themselves higher computational costs. This is where the kernel trick is useful. The idea behind the kernel trick comes from the fact that a major part of the computation in many learning algorithms is accomplished using dot product operation. Hence, for any two training data points $x_1, x_2$ (vectors), if we can somehow find a function $\kappa$ (called kernel function) which can map a non-linear data to a higher dimensional feature space such that:

$$\kappa(x_1, x_2) = \varphi(x_1).\varphi(x_2) \tag{16}$$

then the data in the new feature space is now linearly separable. The function $\kappa$ that makes these transformations is called kernel function. We will explain this using a small example of polynomial kernels. For two arbitrary samples (vectors) $x_i, x_j$ a polynomial kernel function is defined as $\left(1 + x_i^T x_j\right)^p$ i.e.

$$\kappa(x_i, x_j) = \left(1 + x_i^T x_j\right)^p \tag{17}$$

Let us assume that we have a 2D vector as $x = [x_1 x_2]$ i.e. we have two features for each sample. Hence, applying kernel function with $p$ = 2 would give us the following:

$$\kappa(x_i, x_j) = \left(1 + x_i^T x_j\right)^2$$

$$= \left(1 + x_{i1}.x_{j1} + x_{i2}x_{j2}\right)^2$$

$$= \left(1 + x_{i1}^2 x_{j1}^2 + x_{i2}^2 x_{j2}^2 + 2x_{i1}x_{i2}x_{j1}x_{j2} + 2x_{i1}x_{j1} + 2x_{i2}x_{j2}\right)$$

$$= \begin{bmatrix} 1 & x_{i1}^2 & x_{i2}^2 & \sqrt{2}x_{i1}x_{i2} & \sqrt{2}x_{i1} & \sqrt{2}x_{i2} \end{bmatrix} . \begin{bmatrix} 1 & x_{j1}^2 & x_{j2}^2 & \sqrt{2}x_{j1}x_{j2} & \sqrt{2}x_{j1} & \sqrt{2}x_{j2} \end{bmatrix}$$

$$= \varphi(\boldsymbol{x_i}).\varphi(\boldsymbol{x_j}) \text{ where } \varphi(x) = \begin{bmatrix} 1 & x_1^2 & x_2^2 & \sqrt{2}x_1x_2 & \sqrt{2}x_1 & \sqrt{2}x_2 \end{bmatrix} \text{ and the dot}$$
symbol ( . ) is the symbol for dot product.

Hence, a kernel function helps us evaluate the dot product in the lower dimensional original data space without having to transform it to the higher dimensional feature space. In this process, we can benefit from the linearity in the transformed high-dimensional (mapped with $\varphi$) feature space. This is called the kernel trick. There are various types of kernels like linear kernels, polynomial kernels, and gaussian kernels.

Implications of understanding kernel trick can allow their further extensions to their applications in the KCF tracker. For our explanation, we will stick with Gaussian kernels because of their ability to transform data into an infinite feature space.

From Equation (14) and (15) we know that the dual solution of Linear Ridge Regression. is given by:

$$\boldsymbol{\alpha} = (G + \lambda I)^{-1}\boldsymbol{y} \tag{18}$$

where $G = x_i^T x_j$ is called Gram Matrix. If we can map our current non-linear features to a high-dimension feature space using kernel function $\kappa$, we can now use standard machine learning techniques, like linear ridge regression. For simplicity in explanations, let our original data points be denoted by $x_i' \in \mathbb{R}^{m'}$. We will call this space input space. Let us assume that for some kernel function $\varphi$, we can transform our data points from input space to a higher dimensional feature space having data $x_i \in \mathbb{R}^m$ i.e.

$$\varphi: R^{m'} \rightarrow R^m \tag{19}$$

such that $x_i = \varphi(x_i')$. Hence, we can define a solution for the dual case using kernel function $\kappa$ as:

$$\alpha = (\mathrm{K} + \lambda I)^{-1}\boldsymbol{y} \tag{20}$$

where $K_{ij} = \kappa(x_i', x_j') = \varphi(x_i') \cdot \varphi(x_j') = x_i \cdot x_j$. In this case $K$, the gram matrix is called kernel matrix. $K$ is circulant for kernel like Gaussian kernels. Note that we keep stating high dimensional *feature space* brought by $\varphi$ yet we never talk about what these *features* are. For Gaussian kernels, we can obtain the explicit features using the Taylor expansion [96] of these Gaussian kernel function as follows:

$$\varphi(\boldsymbol{x}') = e^{-\frac{\|x'\|^2}{2\sigma^2}} \frac{1}{\sigma^k \sqrt{k!}} \prod_{i=0}^{k} \boldsymbol{x}'_{j_i} \tag{21}$$

where $j$ enumerates overall selections of $k$ co-ordinates of $\boldsymbol{x}'$.

Now, in KCF, the learned function $f(\boldsymbol{z}')$ for some sample $\boldsymbol{z}'$ is given by (Appendix C):

$$f(\boldsymbol{z}') = \left(\sum_i^n \alpha_i \varphi(x_i')\right)^T \varphi(\boldsymbol{z}') = \sum_i^n \alpha_i \kappa(x_i', z') \tag{22}$$

To calculate this learned function, we need the value of $\alpha$. It can be computed using relation between Circulant Matrix $C(x)$, $F$ and sample $x$ [21] such that:

$$C(x)^{-1} = F^{-1}\left(F\left(\frac{1}{x}\right)\right) \tag{23}$$

where $C(x)^{-1}$ is the inverse transform of $C(x)$, $F$ is the Fourier transform and $F^{-1}$ is the inverse Fourier transform. Using Equation (23) and Equation (18), it is possible to diagonalize $K$ and obtain:

$$\alpha = (K + \lambda I)^{-1}\boldsymbol{y}$$

Since $(K + \lambda I)$ is circulant in nature, equation 23 is applicable to $\alpha = (K + \lambda I)^{-1}\boldsymbol{y}$. Hence

$$\alpha = F^{-1}\left(F\left(\frac{1}{K + \lambda I}\right)\right)\boldsymbol{y}$$

$$\text{DFT}(\alpha) = \text{DFT}\left(F^{-1}\left(F\left(\frac{1}{K + \lambda I}\right)\right)\boldsymbol{y}\right)$$

$$\hat{\alpha} = \left(\frac{\hat{\boldsymbol{y}}}{F(K + \lambda I)}\right)$$

$$\hat{\alpha} = \left(\frac{\hat{\boldsymbol{y}}}{\hat{k}^{xx} + \lambda}\right) \tag{24}$$

where $k^{xx}$ is the first row of the kernel matrix $K = C(k^{x_i x_j})$ , a hat $\widehat{\phantom{x}}$ defines the Discrete Fourier Transform (DFT) of a vector, $\boldsymbol{y}$ is the regression label and $I$ is the identity matrix. $\alpha$ can be computed by taking inverse Fourier transform of $\hat{\alpha}$. $k^{xx}$ is also called autocorrelation and the parameter $\alpha$ is learnable and is part of the training stage.

Detection stage, however, does not work with one image patch in isolation. the object of interest to be detection. For training sample $x$ and candidate patch $z$ (where object can be predicted i.e. the new frame), we can use (from Equation (22)) kernelized Ridge Regression to finally evaluate $f(\boldsymbol{z}')$ which represents learned function (detection response) on several image locations as:

$$f(\boldsymbol{z}') = \left(\sum_i^n \alpha_i \varphi(x_i')\right)^T \varphi(\boldsymbol{z}') = \sum_i^n \alpha_i \kappa(x_i', z') = \left(K^{z'}\right)^T \alpha \tag{25}$$

where

$$\alpha = \text{IFFT}\left(\frac{\hat{\boldsymbol{y}}}{\hat{k}^{xz} + \lambda}\right) \tag{26}$$

$k^{xz}$ is called cross-correlation (because of correlation between different samples). Hence, we only need the first row of the Kernel matrix $K$ for our computation. This is one of the main components that can offer much lower computation compared to $m \; x \; m$ or $n \; x \; n$ computation in the original input feature space of data matrix $X_{n \; x \; m}$. Also, from Equation 25, we have $f(\boldsymbol{z}')$ , the full detection response i.e. a vector containing the output for all cyclic shifts of $\boldsymbol{z}$. Equation 25 can be computed more efficiently by diagonalizing it to obtain:

$$\hat{f}(z) = \hat{k}^{xz} \odot \hat{\alpha} \tag{27}$$

where $\odot$ represents dot product. Hence, we can increase the computation speed due to 1.) element-wise multiplication in Fourier Domain and 2.) computation in high dimensional data using kernel trick. Once we are able to locate the target in a subsequent frame, we save this location and then interpolate our model to train the data in a new frame (at this location). This method can be employed for all subsequent frames. Hence, for any instance in the image, with the past target location and current features, the target can be tracked over time. As mentioned in Section 2.1, this type of tracking methodology is called *tracking-by-detection* and has been used in the research time and again.

## 2.3. A Walk-Through Experimental Study

This section provides a detailed overview of the experimental study of the KCF tracking algorithm. Figure 2.6 shows the CF tracking pipeline in detail. It has the following main components: (a) pre-processing stage, (b) feature extraction stage, (c) learning stage, and (e) updating stage. A preprocessing step is performed first on every captured frame of the video before it is passed to the tracker.

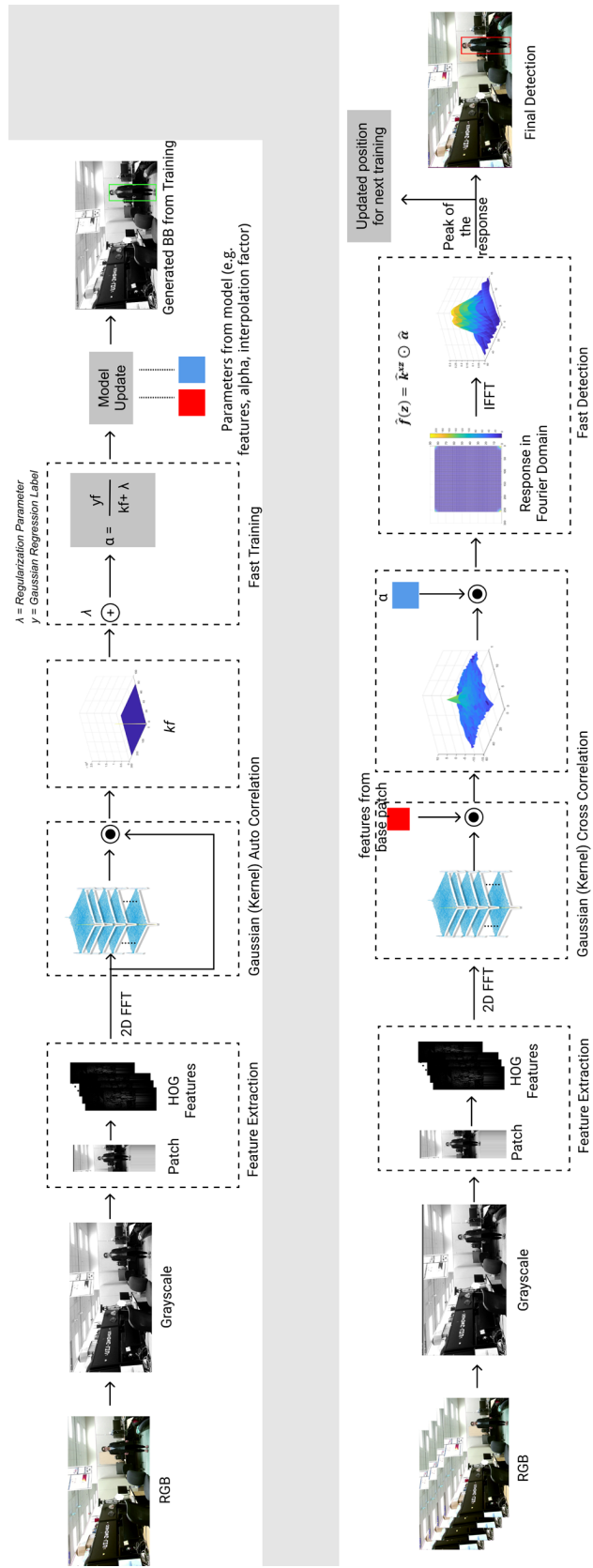

**Figure 2.6    Flow diagram of training and detection pipeline**

In this experimental study, we will be frequently referring to a number of terms where authors feel it will be apt to state the clear definitions:

Ground Truth: Original dimensions of any image frame defined by the region of interest (ROI). They are denoted by $x, y, w, h$ which represent the top-left $x$ co-ordinate, top-left $y$ co-ordinate, width and height of the image frame. Ground truths are used in any algorithm to make comparisons with the predicted/detected results.

Target size: It is defined as the height and width of the subject of interest in the region of interest, in our case the person in this region.

Window size: Target size in the ROI + padding. Padding is added to address the errors at the edges of an image that may be encountered during computation

Patch: A part of an image that is extracted for any computation.

Frame: The entire image used as an input.

Subject: The object of interest intended to be tracked by the tracking algorithm.

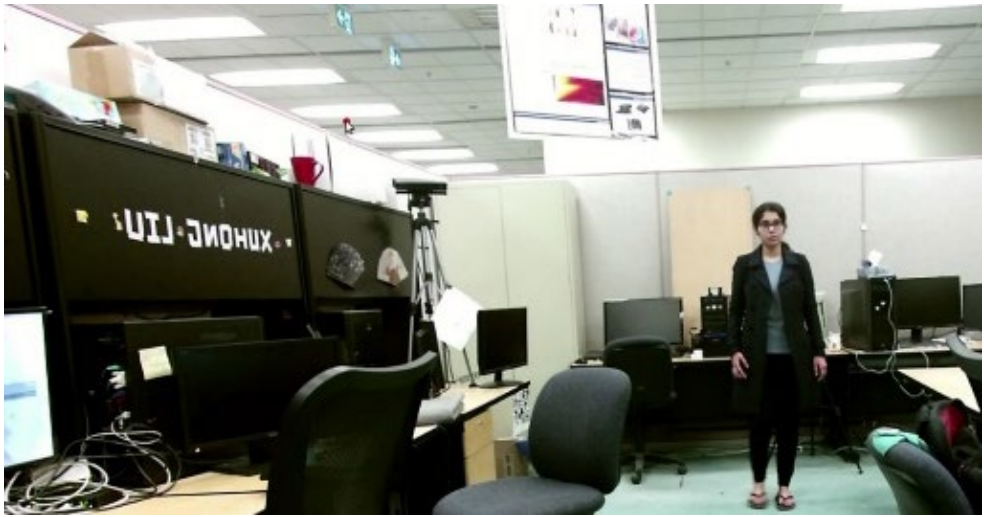

**Figure 2.7**    **A frame from the video provided to the tracker. It has the subject standing in a room. The size of the frame is 1376 x 770**

## 2.3.1. Input Stage

The images used for the experimental analysis have been collected from the Kinect V2 RGB sensor. The subject is a single moving object in a cluttered environment to best replicate a noisy scenario. The algorithm has been implemented and tested on Ubuntu 16.04 with C++ as well as MATLAB R2019b.

## 2.3.2. Preprocessing Stage

### *Image Extraction*

The image captured from the Kinect V2 camera (Appendix A) is converted from RGB to grayscale. For example in Figure 2.8, the RGB to the grayscale conversion of the first frame results in $1376 \times 770$ pixels. It is done for mathematical simplicity because dealing with one channel is simpler than 3 channels and yet represents all aspects of the image like brightness, contrast, edges, shape, contours, texture, perspective, shadows, and so on, without addressing color.

### *Image Resolution*

Despite having a grayscale image, an image in high resolution can be computationally expensive. This would be a bigger issue in tracking since a large portion of tracking performance is defined by its computation speed. If the image being processed is very large or has a very high resolution, it would mean that there are more pixels per inch (PPI), resulting in more pixel information and hence more computation. The image quality and/or size will also depend on the viewpoint of the camera and the location of the subject. For example, images captured from the viewpoint of surveillance cameras (assuming the target is far) will be smaller. However, when a subject is near to the camera, for instance, a robot following a person, the image may be larger. In such situations, it is recommended that the image is resized to lower the resolution thereby improving the computational speed.

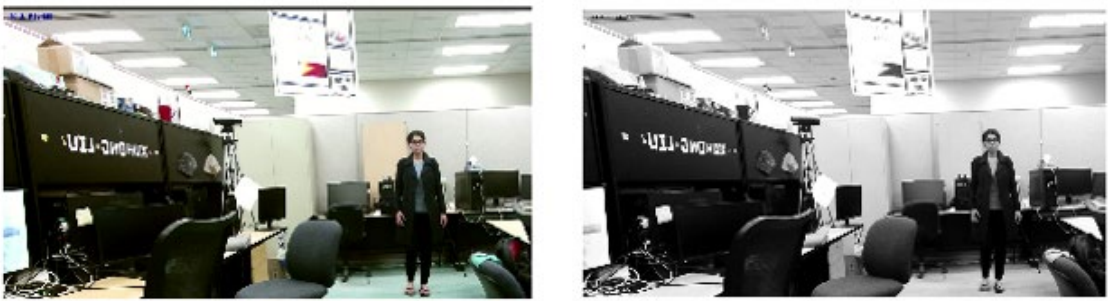

**Figure 2.8**       **RGB (left) format of the first frame is converted to grayscale (right) during tracking pipeline**

Padding will be added to the patches/target-size in ROI allowing space for the wrap-around to occur hence avoiding contamination of actual output pixels. This new size of the window with the added padding is referred to as 'window size'.

### 2.3.3. Training Stage

***Regression labels***

From Section 2.2, it is observed that training samples are composed of shifted versions of base samples. For each of these samples, we will need to specify a regression target. All these regression targets collectively form a vector of regression targets $y$ (expected response) which we had defined in Equation (24). These regression targets can be binary or Gaussian or any other similar function. However, since the Gaussian function is relatively smoothers, we will choose regression targets that are Gaussian in nature. Hence, the Gaussian regression label ($y$) is a Gaussian distribution with peak at the center (Figure 2.9). As we will see later, this will be helpful during training in localizing the object of interest using Equation (24). With the help of the regression target, the algorithm can localize the object since the center peak will denote the more likely probability of the presence of an object.

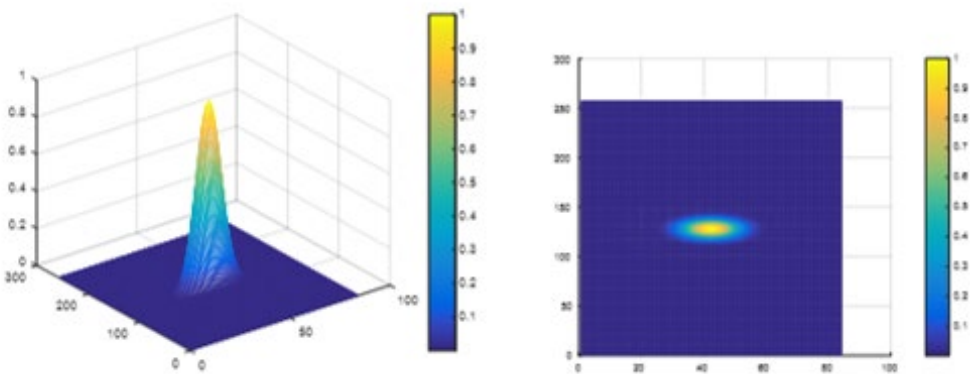

**Figure 2.9**   **Gaussian regression with the peak at the center and color bar on the right of each Gaussian representation. The blue area in the image is the area of the target size under view. The picture on the left side view and the one on right is the top view. The yellow color shows the highest probability (Probability =1) and blue the lowest (Probability= 0).**

### *Patch Extraction*

For the first frame, the target is located at the defined ground truth definition, however, in subsequent frames, it is located from the data provided by the detection stage (explained later). Padding is also added to this target image. Hence, a patch of the size of the window size centered on the subject is cropped from the grayscale format of the frame. Figure 2.10 (b) illustrates a patch of $135\ x\ 412$ pixels (the original target size) (d) illustrates a patch of 337 x 1030 pixels (window size) both cropped from the frame of 1376 x 770 pixels.

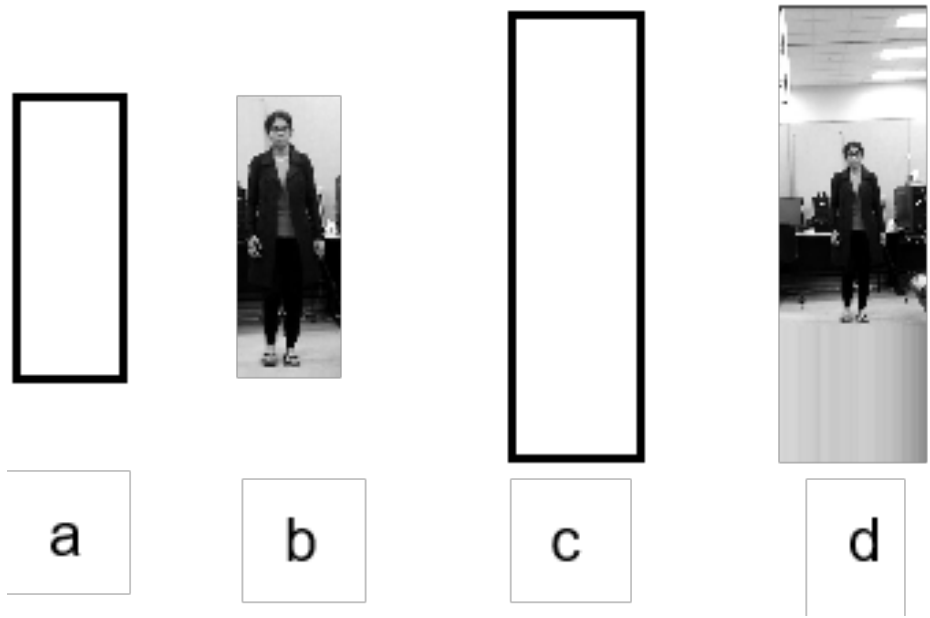

**Figure 2.10.**  **(a) Dimensions of original 'target size' 135 x 412, (b) Patch of the size of the 'target size' extracted using the GT for the first frame, (c) Dimensions of the 'window size' of 337 x 1030 once the padding added, (d) Patch of the size of the 'window size' extracted using the GT for the first frame. The patch shown in (d) is only for the N = 1 where N is the frame number**

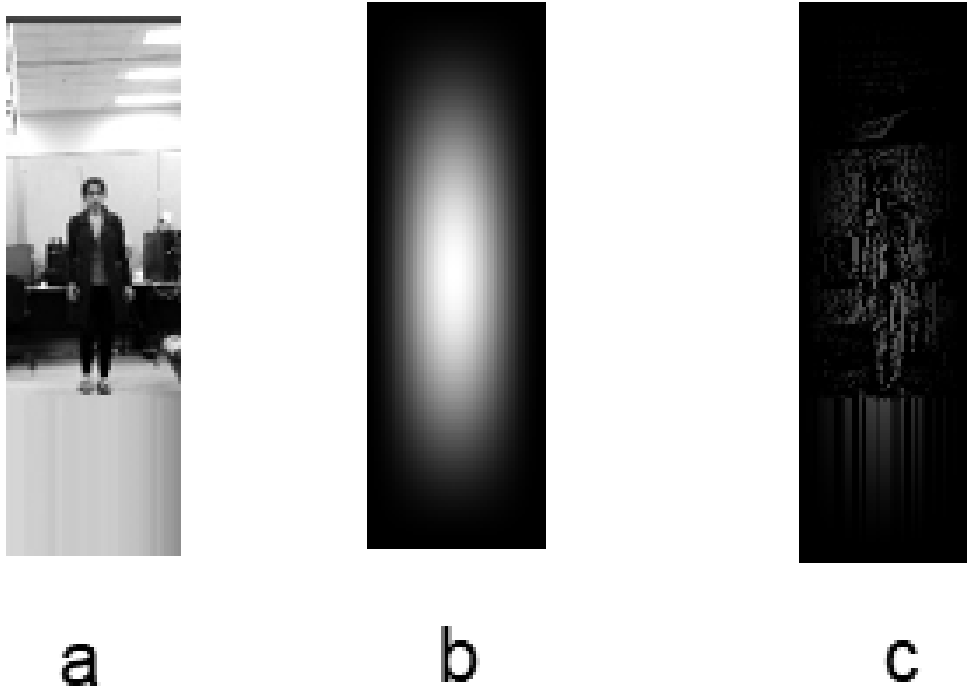

a       b       c

**Figure 2.11.**   **(a) Image patch whose HOG feature is extracted; (b) Cosine window used for smoothening the features; (c) Processed HOG feature vectors after applying cosine window**

### Feature Extraction

The patch is chosen such that it has the object of interest at its center. HOG features are extracted from this image patch. Figure 2.11 shows the HOG features of the patch with the object of interest in the center (zoom in for better clarity). These features extracted from the patch undergoes windowing (here using cosine window) to locate and focus on the object of interest. Wind owing is also useful in eliminating the noise at the edges.

### Auto-correlation response

The features extracted from the previous step are transformed into the Fourier domain. From Equation (24), we calculate $\alpha$ as part of the training process. This computation is part of the kernel trick, and hence correlation associated with it is called kernel correlation. Equation (24) provides us with the equation for fast training as shown below:

$$\hat{\alpha} = \left( \frac{\hat{y}}{\hat{k}^{xx} + \lambda} \right) \qquad (28)$$

where $\hat{k}^{xx}$ are the learned features (autocorrelation of the features), $\hat{y}$ are the regression labels and $\lambda$ is a model parameter. The model parameter $\alpha$ and $\hat{k}^{xx}$(learned features) will be updated at every training stage. Since the training stage only involves dot product hence computation is faster. All the steps above from the KCF tracker can be visualized from file $response\_plots.m$. Figure 2.12 shows the target position estimated by the training stage. This 'position' (coordinates of the bounding box) will be used to make target prediction in the next frame in the detection stage (next section).

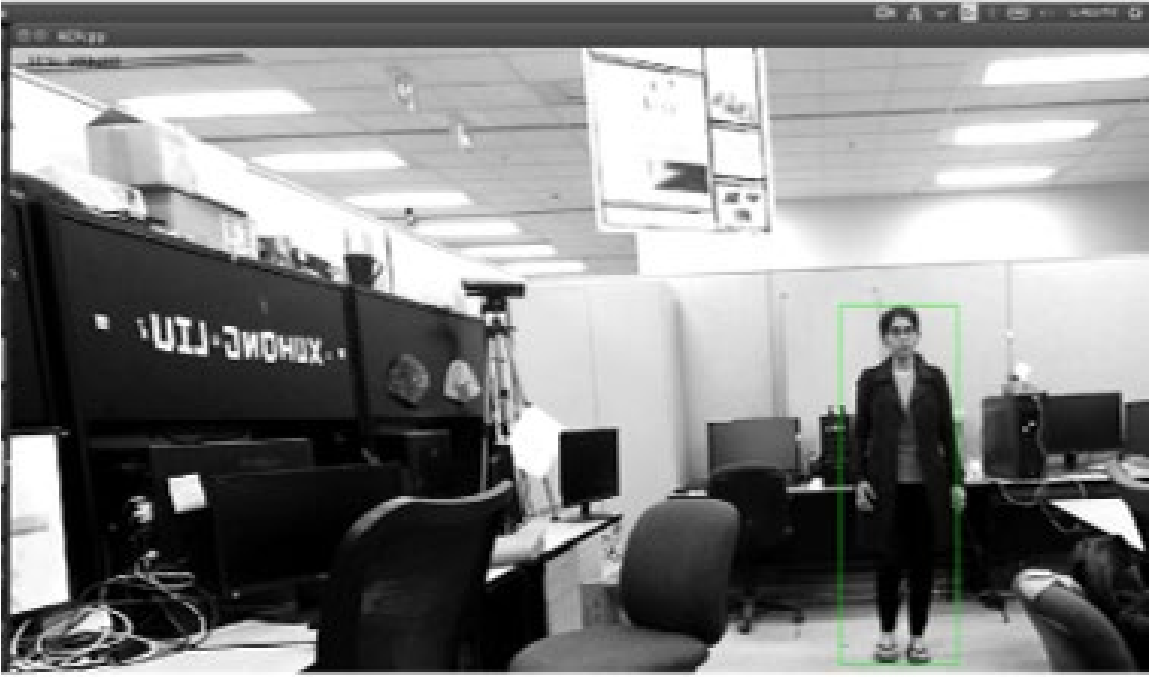

**Figure 2.12** **The position of the target computed at the learning stage. This position will serve as ground truth for the next frame in the detection stage (Best viewed in color)**

## 2.3.4. Detection Stage

### *Patch Extraction*

The algorithm reads the next frame in sequence for the detection of the target. It uses the target location computed at the training stage as the location coordinates for the new frame and extracts a patch/sub-window of the size of the window size for detection. Figure 2.13 shows the patch extracted. Table 2.2 represents it using actual target co-ordinates for the experiment.

***Pre-processing and feature extraction***

**Table 2.2.     Table shows the way target location computed on one frame is used for patch extraction for the subsequent frame**

| Target coordinates format | Computed target location after training on Frame 1 | Target location for patch extraction at Frame 2 (this target location will be used to make final prediction) |
|---|---|---|
| $(x, y, w, h)$ | 1001.50, 353, 135, 412 | 1001.50, 353, 135, 412 |

The pre-processing step and feature extraction for this newly extracted patch are similar to the training process. HOG features are extracted from the grayscale patch and windowing in done to get the processed features.

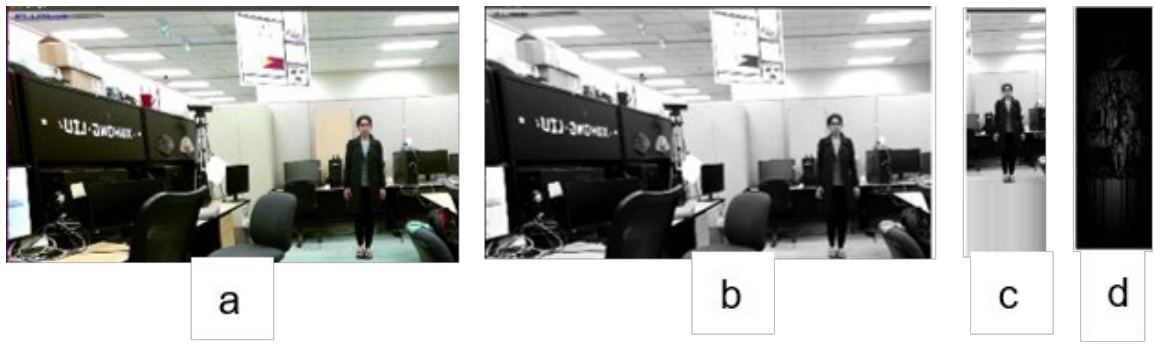

**Figure 2.13     (a) shows the new RGB frame used as an input to the algorithm (b) shows the grayscale image of the same input (c) shows the extracted patch and (d) shows the HOG features extracted**

***Cross-correlation response***

From Equation (25), we know that the detection response for the classifier at all shifts is given by:

$$f(\mathbf{z}') = \left(K^{z'}\right)^{T} \boldsymbol{\alpha} \tag{28}$$

which can be computed more efficiently as:

$$\hat{f}(z) = \hat{k}^{xz} \odot \hat{\alpha} \tag{29}$$

where $\alpha$ is the model parameter, obtained and updated every time at the training stage. Hence, as seen from Equation (27), we will need to compute the correlation between features from the new patch and the updated model features from the last training stage. We call this 'kernel' cross-correlation because of their relevance to the kernel trick. Figure 2.14 shows the output of the kernel correlation.

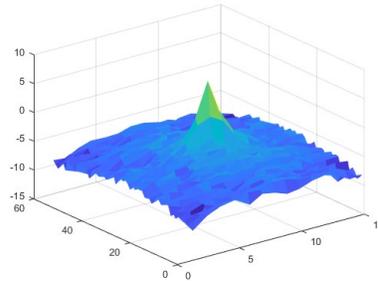

**Figure 2.14.** **The response of the cross-correlation of the features (denoted by $k^{xz}$ in Equation (26))**

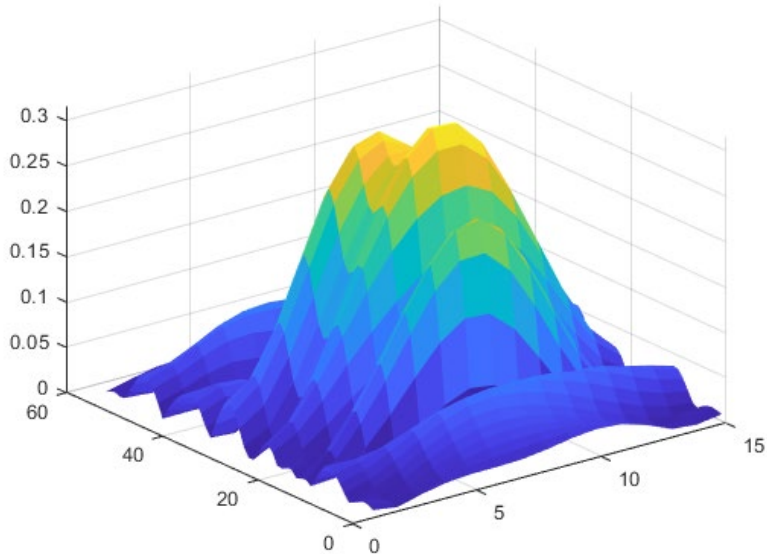

**Figure 2.15.** **Shows the response in the spatial domain. The detected target location is the peak i.e. the highest response**

The peaks correspond to the maximum response (response is shown in Figure 2.15.) and the location of the peak can be used to compute the final detected position of the target in this new frame (i.e. Figure 2.16).

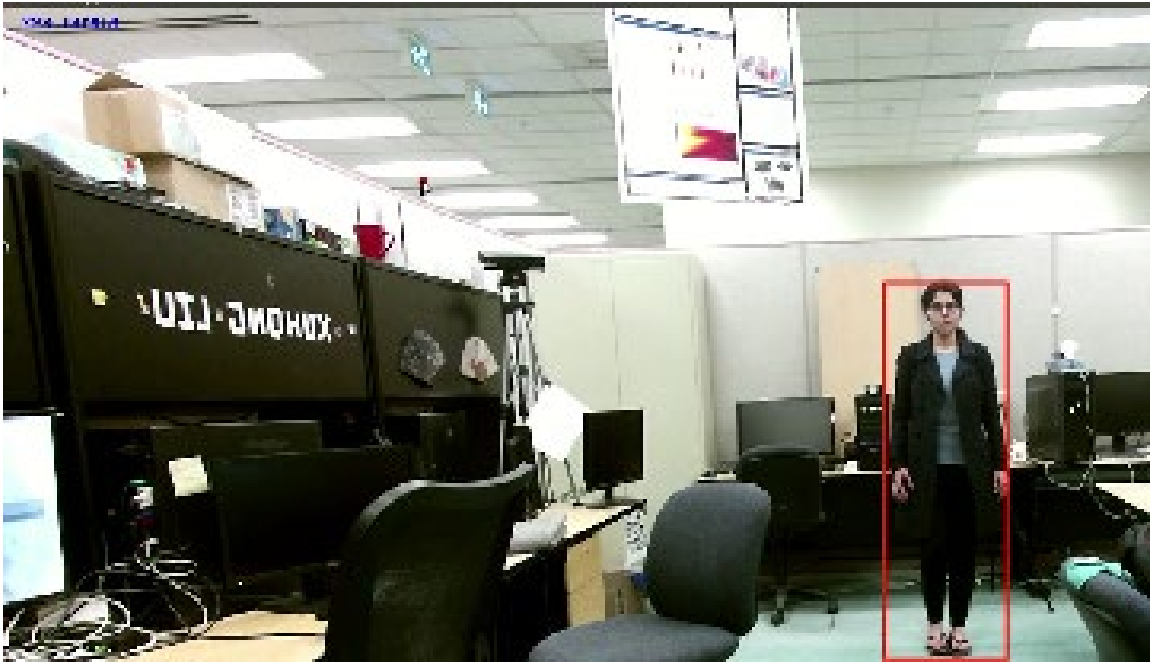

**Figure 2.16    Final target detected using the computed response (Best viewed in color).**

This new target location is interpolated for the next training stage and process continues to give us a series of the detected target in every frame which can be used for tracking and hence the name 'tracking-by-detection' as shown in Figure 2.17.

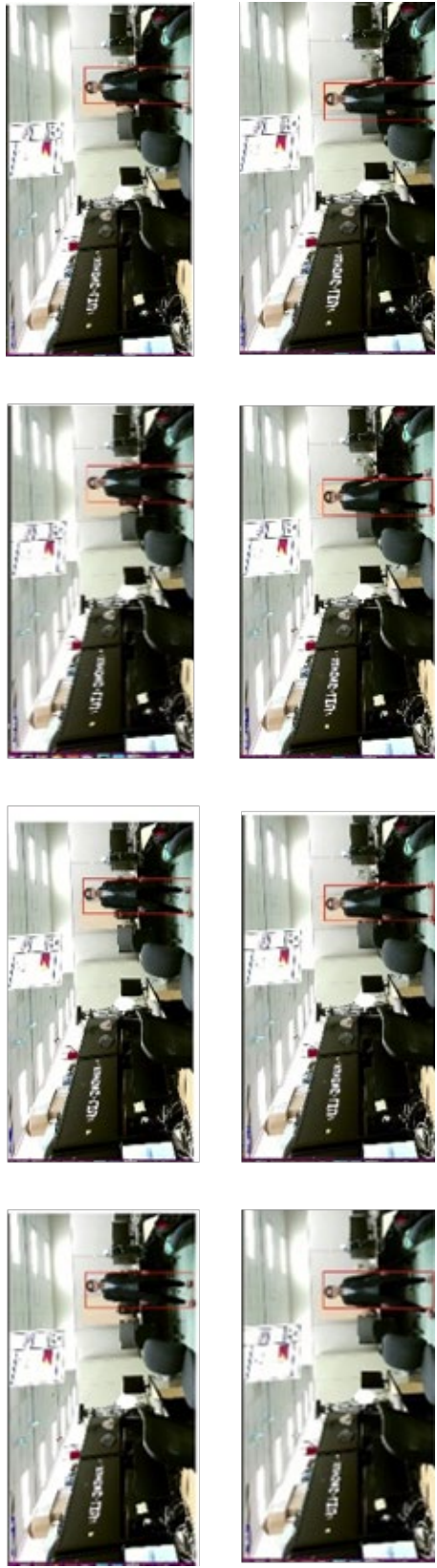

**Figure 2.17    An example of sequential frames from our experimental case studies showing the results of tracking-by-detection**

## 2.4. Discussion

This chapter presents a detailed overview of the KCF tracker which included both analytical and experimental results with various step-by-step intermediate visualization of the results. This is contrary to recent survey papers [35] [36] which summarize the recent developments in correlation filter tracking techniques and tracking community in general. None of the previous publications discuss either of the techniques mentioned in detail. Most importantly, none of the recent surveys have demonstrated how the mathematical implementation of different parts of the KCF tracker can be studied and visualized using demonstrated code for practical understanding. Some of these visualizations and explanations include working of circulant matrices, block circulant matrices, proof of diagonalization of circulant matrices, working of gaussian correlation, response detection in KCF tracker and many more. This Chapter also discusses Kernel Correlation Filter in detail. It is expected that such detailed explanations will benefit research community in adding to their current understanding of this tracker and will potentially encourage them to further investigate any improvements that can be made to it.

# Chapter 3.

# Performance Analysis of KCF Tracker Using Real-time Experimental Setup and Standard Dataset

Motivated by the suggested performance of the KCF tracking algorithm, we realize the need also exists to further analyze and study the experimental performance of the tracker under similar conditions. Commonly such a study is accomplished using standard datasets. However, it was realized the tracker performance needs to be also validated in real-time under different challenging possible scenarios. This chapter builds on this motivation and discusses tracker performance.

This chapter covers performance analysis in two parts. Section 3.1 describes the process of real-time data collection. It is followed by Section 3.2 detailing various scenarios that were taken into consideration while tracking in real-time. Section 3.3 does an in-depth analysis of KCF as a tracker on this real-time dataset collected by the author using the Kinect RGB camera. The analysis is done on various challenging scenarios (clutter, deformation, etc.) and using various standard methodologies. It is followed by a discussion of the performance of various CF trackers, including KCF, on three popular datasets - OTB-50 in Section 3.4, VOT 2015 in Section 3.5, and VOT 2019 in Section 3.6.

## 3.1. Real-Time Analysis

The experimental setup is configured at the Networked Robotics and Sensing Laboratory located at Engineering Science to validate the real-time capabilities of the KCF tracker. During the real-time experimental analysis, we saved the frames and the tracking results (predicted annotations) for quantitative analysis as discussed in Section 3.3. These collected frames include nearly 6000 RGB images.

Our experimental setup consists of a Microsoft Kinect v2 (see Appendix A). It uses a paired infrared projector and camera to calculate depth value. Direct sunlight severely impacts its performance outdoors hence all the datasets collected are from indoor spaces.

We manually annotate the ground truth (the correct target location) of the real-time dataset by drawing a bounding box on each frame as follows: A minimum bounding box covering the target is initialized on the first frame. In the next frame, if the target moves, the bounding box will be adjusted accordingly; otherwise, it remains the same. All frames are manually annotated by an author to ensure high consistency. When an occlusion occurs, the ground truth is defined as the minimum bounding box covering only the visible portion of the target. When the target is completely occluded there will be no bounding box for this frame. We annotate all following frames in this way only

Numerous performance measures have been suggested in the visual tracking community. However, any of them has not been singled out as a de facto standard. In most cases, standard datasets are used for comparative results. However, since our purpose was to analyze the tracker in a real-time environment, we collected our dataset using different subjects from different scenarios to test the robustness against, for example, occlusions, out-of-view, etc. Inspired by the work [97], we define a general definition for the description of the object state in a dataset of N sequences as:

$$S = \{(R_t, x_t, y_t)\}_{t=1}^{N} \tag{30}$$

where $x_t, y_t \in C$ denotes the center of the object. $R_t$ denotes the region of the object at any time $t$.

Hence, for any particular example and at any particular time $t$, the center can be denoted as:

$$C_t = (x_t, y_t) \tag{31}$$

A subject in the algorithm is tested against the ground-truth (the true value of the object location obtained using manual annotation). We define our ground truth annotations as $\Delta_G = x_g, y_g, w_g, h_g$ and tracker's predicted annotation as $\Delta_t = x_t, y_t, w_t, h_t$ where $x, y, w, h$ are the top-left x co-ordinates, top-left y coordinates, width, and height respectively of the subject in question.

## 3.2. Tracking Scenarios Selection

Scenarios used for the analysis were selected based on the major challenges faced by the visual tracking community. These scenarios (a sample scenario is shown in Figure 3.1) include a) normal b) background clutter c) deformation d) occlusion e) fast motion f) out-of-view scenes. The figure shows sample images from the 'occlusion' scenario which was part of our tracker analysis. Figure 3.8 gives a detailed view of all the scenarios used and sample images from each scenario.

A brief description of these scenarios is given below:

**Normal:** Normal scenarios are when the target under observation is under ideal conditions for observation. This would include no occlusion, deformation, illumination, etc. Most tracking algorithms are expected to perform their best in these scenarios.

**Background Clutter**:  When viewing cluttered scenes, observers may not be able to clearly separate the object from the background. Such scenes are composed of several familiar objects with similar colors and textures, making it harder to recognize the target by an algorithm.

**Deformation:**  In the case of deformation, the objects do not always reside in the regular grids of the image space. The location of the object as a whole or object parts, often vary in frames of a video. E.g., when a person extends his arms, the bounding box changes to accommodate for extended hands target. This would make the bounding box larger than in the usual standing scenario. The accuracy of a tracking algorithm is adversely affected due to these appearance variations.

**Occlusion**: In the case of occlusion, the target under observation has their body parts or movements hidden. This can be due to a) obstruction in the path of the target (if the target is moving), b) blockage of view by a moving object (if the target is stationary). Occlusion is one of the most challenging issues of visual tracking since the information of the target is unknown for a short time. Hence, the algorithms make probabilistic predictions of the state of the target using prior available information.

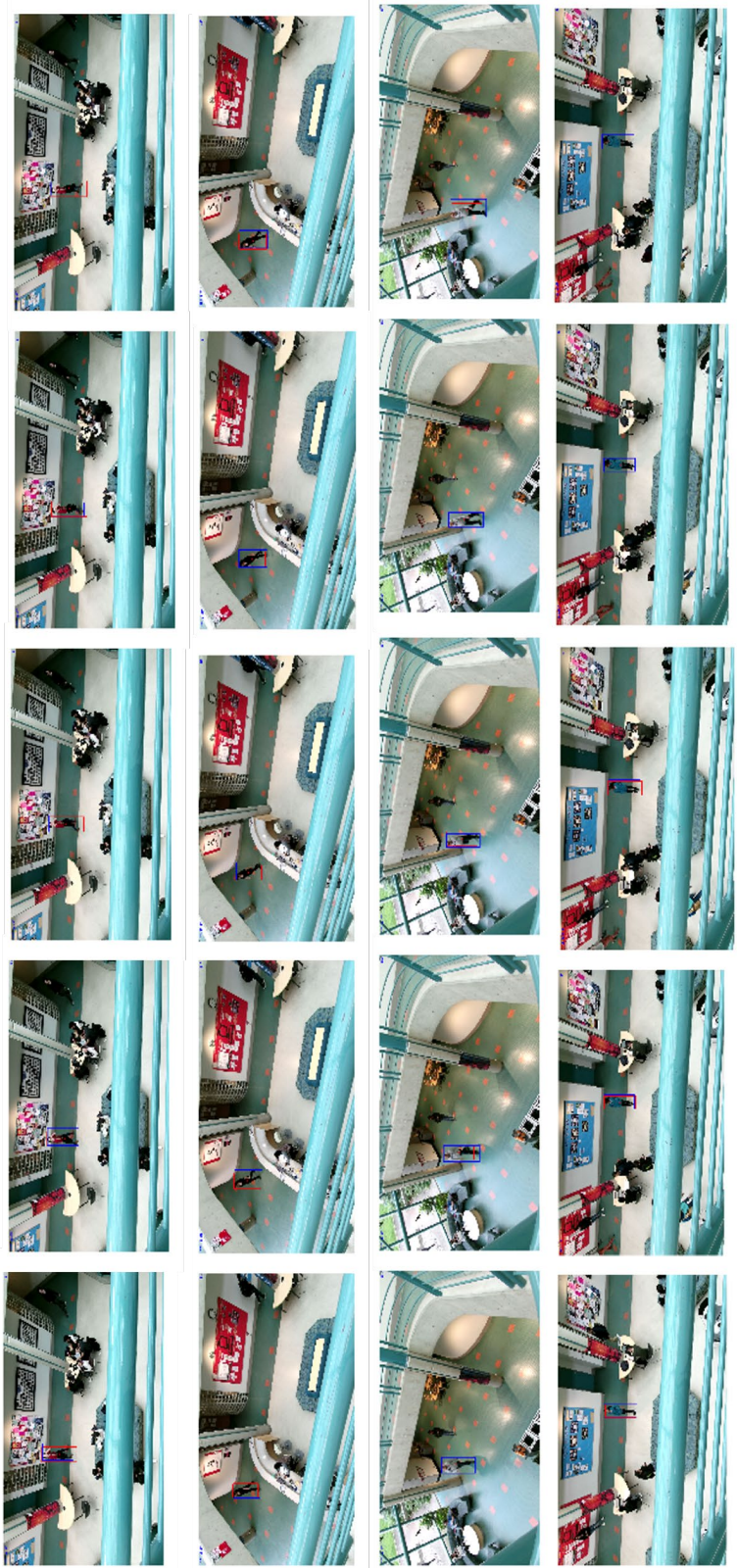

**Figure 3.1.** Sample images from 'occlusion' scene which were used in tracker evaluation

**Motion**: Motion in our context refers to how fast the target is moving. Fast-moving target often leads to motion-blur. Such targets also cause model drift where the position of the target predicted by the model is not smooth in timing thereby affecting future predictions.

**Out-of-view**: Out-of-view is a state when the target, during its observation, moves out of the visible range of the sensor (example: camera). This scenario is different from occlusion since in occlusion target is still in the visible range of the sensor but temporarily hidden. However, in out-of-view, the target disappears from the visible range altogether. In such cases, the model loses information of the target making it harder to re-detect.

These scenarios formed the basis of the real-time dataset which was collected using our experimental setup. The dataset used in the evaluation was collected by the author using the Kinect V2 RGB camera. The environment was chosen to replicate the real-time environment for testing the KCF tracker. Subjects chosen were colleagues and friends and varied in size, the color of the dress, etc. Table 3.1 gives a more detailed number on the number of datasets (all subjects combined) collected for individual scenarios.

**Table 3.1.     Details about the RGB dataset collected using Microsoft Kinect V2**

| Scenarios | Subjects | Data (no of frames) |
|---|---|---|
| Clutter | 6 | 1677 |
| Deformation | 6 | 1494 |
| Normal | 6 | 900 |
| Occlusion | 6 | 663 |
| Motion | 6 | 784 |
| Out of view | 6 | 620 |
| Total | | 6138 |

## 3.3.  Performance of Real-Time Kinect RGB Dataset

Though there are various performance metrics e.g. confusion matrix, success rate, etc., these metrics are better to use when we are comparing two or more trackers. In the present case, we focus on evaluating the tracking performance w.r.t. the ground truth and hence we will use different metrics. Figure 3.3 will show some high peaks. As an observation, we see that y-axis has to be capped at 150 to show clearer graphs.

Some of the peaks are due to missed bounding boxes and some are due to momentarily misplaced bounding boxes. The original image size is around $1370 \, x \, 770$ to give an idea of the big size of the image.

### 3.3.1. Centre Error

This is one of the oldest measures [98][99] and relatively still popular. It measures the difference between the target's predicted center from the tracker and the ground-truth center. Mathematically, it can be calculated using the Bhattacharya method as:

$$\text{Centre Error} = \sqrt{(x_1 - x_2)^2 + (y_1 - y_2)^2}$$

where the center for ground truth bounding box can be defined as $C_G = (x_1, y_1)$ and for the predicted bounding box can be defined as $C_T = (x_2, y_2)$. The resulting error can be plotted against the sequence of frames as a graph as can be seen from Figure 3.3. It requires less annotation effort and hence can be used for quick analysis.

**Observation**: The plots for the center error for all six scenarios can be seen in Figure 3.3. We observe that in cases of occlusion, and out-of-view, the center error was relatively higher as compared to other scenarios, for all datasets. In the case of fast motion, the center errors are low in the beginning but tend to increase after 40% of the frames have been tracked indicating it may be more robust for short-term tracking than the long term. In the case of clutter, one dataset did not perform well indicating that the tracker may be sensitive to additional factors. The tracker performed best for the general scenario (in ideal settings) indicating that given ideal scenarios, KCF works well.

| 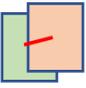 Ground Truth | 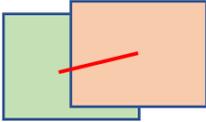 Prediction | Centre Distance ———— |
|---|---|---|
| 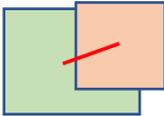 | | |
| Case 1: If the prediction and ground truth annotation have same size, for a smaller subject, center error is less. | Case 2: If the prediction and ground truth annotation have same size, for a bigger subject, center error is more than Case 1 | Case 3: If the prediction and ground truth annotation have different size, for any subject, center error will differ depending on the prediction annotation. |

**Figure 3.2.** The figure shows the reason for the sensitivity of center error because of ground truth and prediction annotations

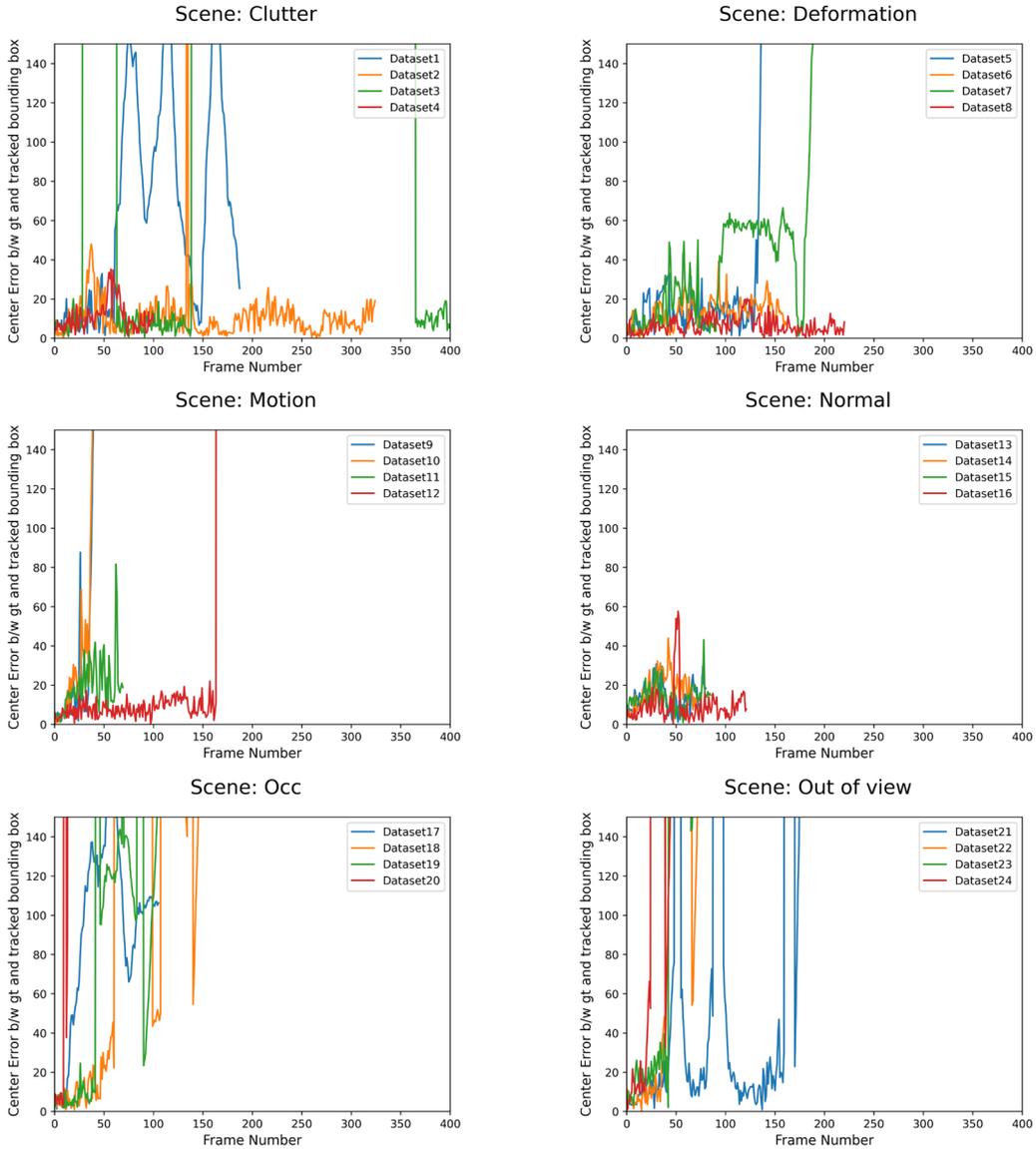

**Figure 3.3.** Plots for center error for (Row1:Top, L-R) Clutter, Deformation, (Row2: L-R) Fast Motion, Normal (Row3:Top, L-R) Occlusion, Out-of-view (Best viewed when zoomed in)

Even though the center error is good to give a quick overview of the subject's performance, the results can be misleading. The center error results are dependent on sensitivity to annotations. Results are independent of the size of the target and hence show drastic differences in center error for the same subject at a similar location at a different size. The results also don't explicitly reflect cases of tracking failure.

### 3.3.2. Centre Co-ordinate Error (X and Y coordinates)

The center error analysis is a good way to look at the performance and observe the cases where the errors are high. However, it would be beneficial to see to know the factors which contribute to these errors. For example, we don't know if it is the variation in height or width which contributes to the most error. For this analysis, we calculate the center errors but for each coordinate of the center. Mathematically, for center $C_t = (x_t, y_t)$ at any time $t$:

$$\text{CentreError}_x = (x_2 - x_1)$$

$$\text{CentreError}_y = (y_2 - y_1)$$

The results for the center error with respect to the x-coordinate is shown in Figure 3.4. The results for the center error with respect to the y-coordinate is shown in Figure 3.5.

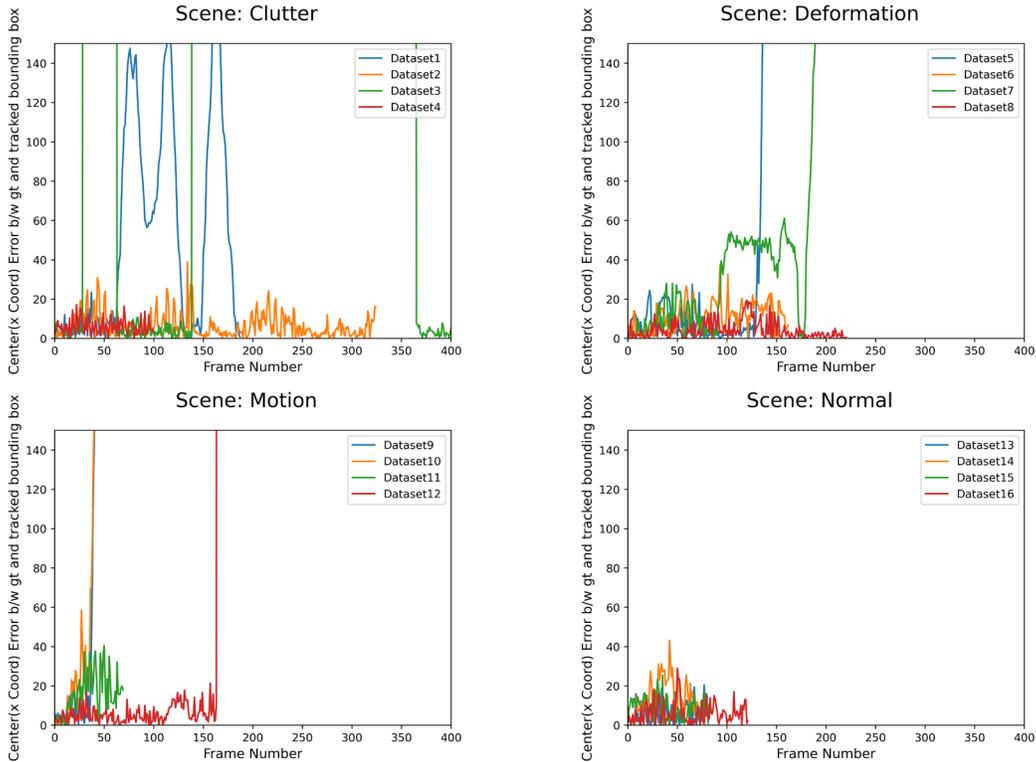

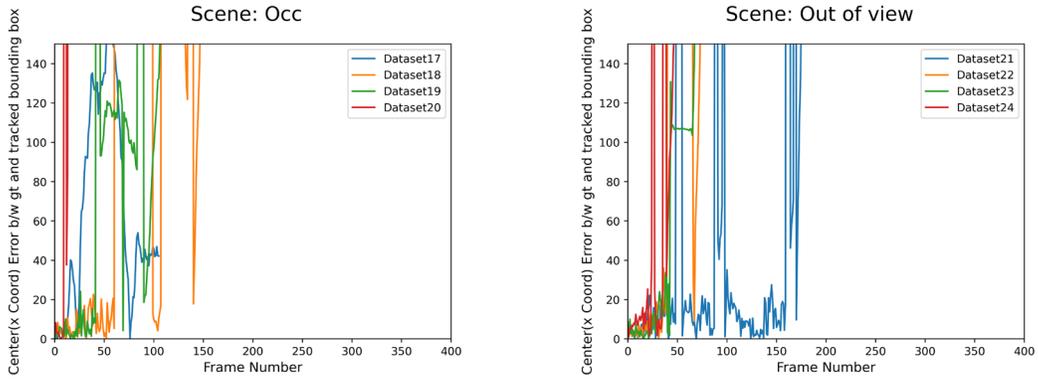

**Figure 3.4.  Plots for center error for x center-coordinates (Row1:Top, L-R) Clutter, Deformation, (Row2: L-R) Fast Motion, Normal (Row3:Top, L-R) Occlusion, Out-of-view (Best viewed when zoomed in)**

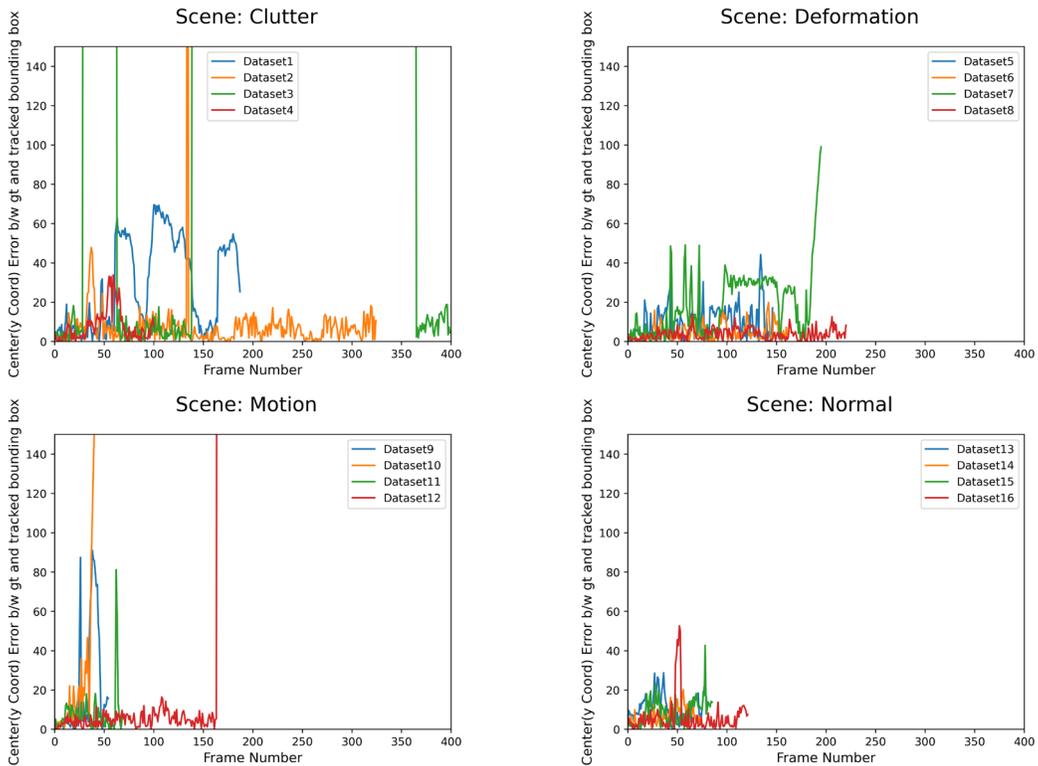

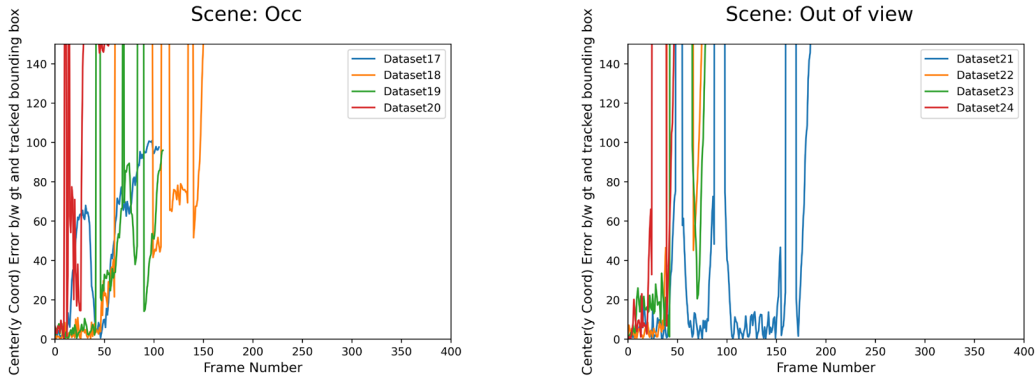

**Figure 3.5.**    Plots for center error for y center coordinates (Row1:Top, L-R)
Clutter, Deformation, (Row2: L-R) Fast Motion, Normal (Row3:Top, L-
R) Occlusion, Out-of-view (Best viewed when zoomed in)

**Observation**: Though most observations were similar to what we observed in
center error for all scenarios, the occlusion scenario stands out. As can be seen from the
plots in Figure 3.5, the error in y-coordinates of the center tends to show a higher error
as compared to its x-coordinate for similar scenarios for all datasets. Hence, either the
height changes more rapidly than the width, or the tracker is highly dislocated in the
vertical direction as the subject is being tracked in consecutive frames over time. Hence,
we can improve the predictions such that the tracker maintains the target height (its state
in y-dimension) within reasonable error limits, we expect that tracker will become more
accurate and stable to changes in the environment.

### 3.3.3. Intersection of Union

The intersection of Union (IoU) is an evaluation metric used for calculating the
match between the ground truth and predictions. In other words, a higher value of IoU
implies that the coordinates $x_t, y_t, w_t, h_t$ of the tracked bounding box (prediction) are
going to closely match the coordinates $x_g, y_g, w_g, h_g$ of the subject location (ground truth).
It computed using:

$$IoU = \frac{Area\ of\ Overlap}{Area\ of\ Union}$$

IoU can be seen as a metric to measure the accuracy of the tracker. IoU rewards
more to the prediction parameters (co-ordinates) which overlap heavily with the ground
truth parameters (coordinates). The results for the IoU for each scenario for all datasets are shown in Figure 3.6. The values of the y-axis denote the IoU where 0 denotes no overlap and 1 denotes a 100 % overlap. A threshold for 0.5 is set to make a distinction between desired and undesired values of IoU. Any value above the threshold would imply that there is a 50% or more overlap between the ground truth and the trackers' prediction, and hence the tracker is performing well.

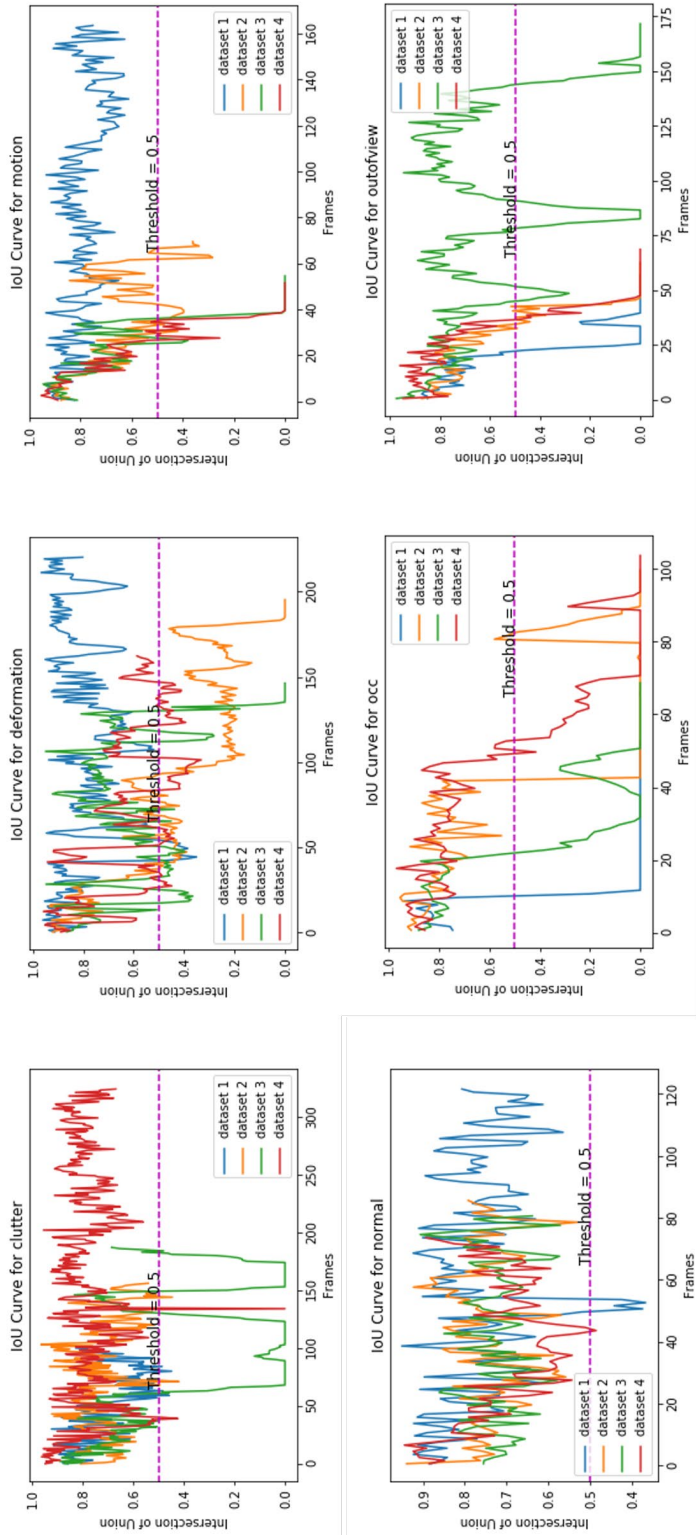

**Figure 3.6.** Plots for Intersection of Union for (Top, L-R) Clutter, Deformation, Motion (Bottom, L-R) Normal, Occlusion, Out-of-view (Best viewed when zoomed in)

**Observation:** Ignoring few inconsistent outliers due to datasets, it was generally observed that the tracker performs best in ideal scenarios (first graph in the second row). It was most unstable in case of occlusion and out-of-view. In the case of deformation, the performance was initially better, however, it degraded as the subject continues to undergo deformation over time. In the case of out-of-view, the tracker failed abruptly indicating that once the subject goes out of view of the camera, it is never recovered.

### 3.3.4. Precision Curve

Precision is another evaluation metric that gives information about what proportion of positive identifications was actually correct. Formally, it is defined as:

$$Precision : \frac{TP}{TP + FP}$$

where True Positive (TP) is an outcome where the model/algorithm correctly predicts the positive class i.e. in our case the algorithm correctly predicts the subjects where the subject exists. On the contrary, False Positives (FP) in an outcome where the model/algorithm incorrectly predicts the positive class i.e. algorithm predicts the subject where it does not exist. Hence, the higher the precision, the better the prediction. In general practice, a precision value below a certain threshold is not desired.

**Table 3.2.      Precision over the sequence of datasets for each scenario/environment**

|           | Clutter | Deformation | Motion | Normal | Occlusion | Out-of-view |
|-----------|---------|-------------|--------|--------|-----------|-------------|
| Dataset-1 | 0.9808  | 0.9168      | 1.0    | 0.9718 | 0.4958    | 0.7480      |
| Dataset-2 | 0.9841  | 0.6005      | 0.9123 | 1.0    | 0.7857    | 0.8717      |
| Dataset-3 | 0.9843  | 0.7731      | 0.8852 | 0.9927 | 0.8436    | 0.8577      |
| Dataset-4 | 0.7093  | 0.7274      | 0.8884 | 1.0    | 0.6784    | 0.8798      |
| **Average** | 0.9038 | 0.7544     | 0.9213 | 0.9911 | 0.7008    | 0.8393      |

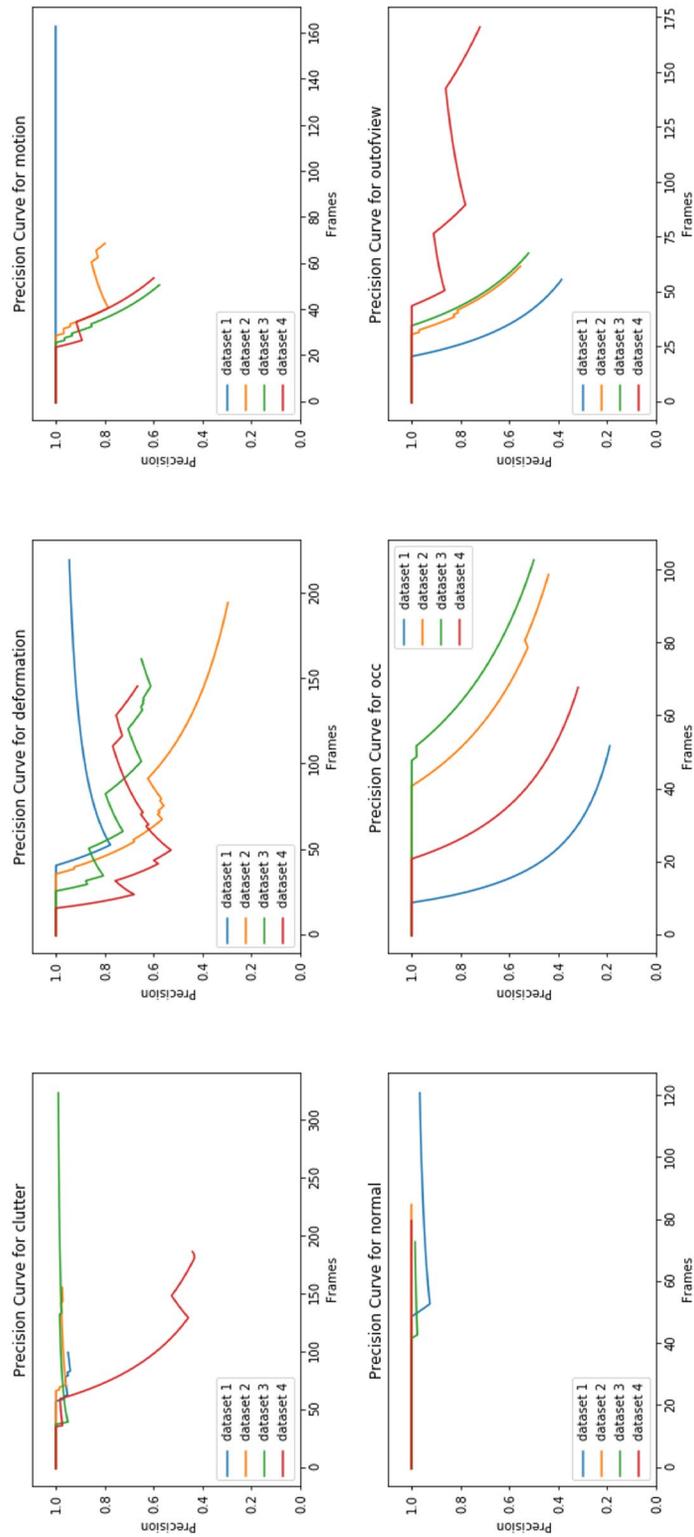

**Figure 3.7.** Plots for Precision for (Top, L-R) Clutter, Deformation, Motion (Bottom, L-R) Normal, Occlusion, Out-of-view (Best viewed when zoomed in)

**Observation:** From Figure 3.7, we observe that precision is lowest for occlusion, although depending on the dataset and when the subject was occluded, the precision may decrease at different times. The value of precision can also be seen in Table 3.2 where average precision is 70%, the lowest among all scenarios. It is closely followed by deformation at 75 % where depending on subject movements it may or may not recover over time. The out-of-view scenario performs poorly too. It is better than occlusion and deformation because the subject is tested under normal scenarios initially and it's only towards the end where goes out of the view of the camera (last given in Figure 3.7). As can be seen from the plot in the figure, we can see that if the target never recovers, it indicates that the tracker does not have the ability to re-identify the subject once it goes out of the view and returns. For scenarios like clutter and ideal cases, the tracker seems to perform well.

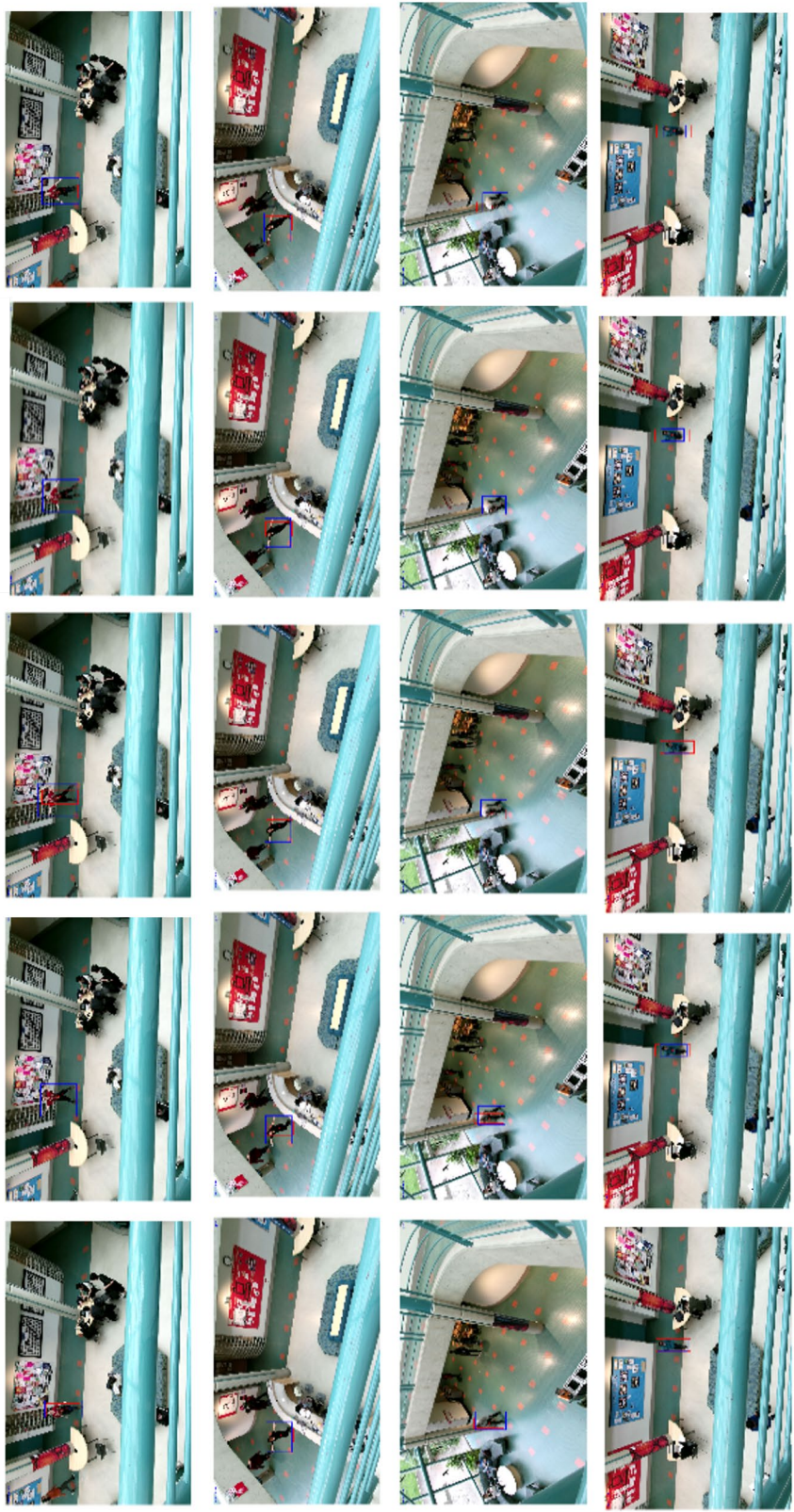

**(a) Clutter (blue is the ground-truth, red is the result from the KCF tracker). Best viewed in color.**

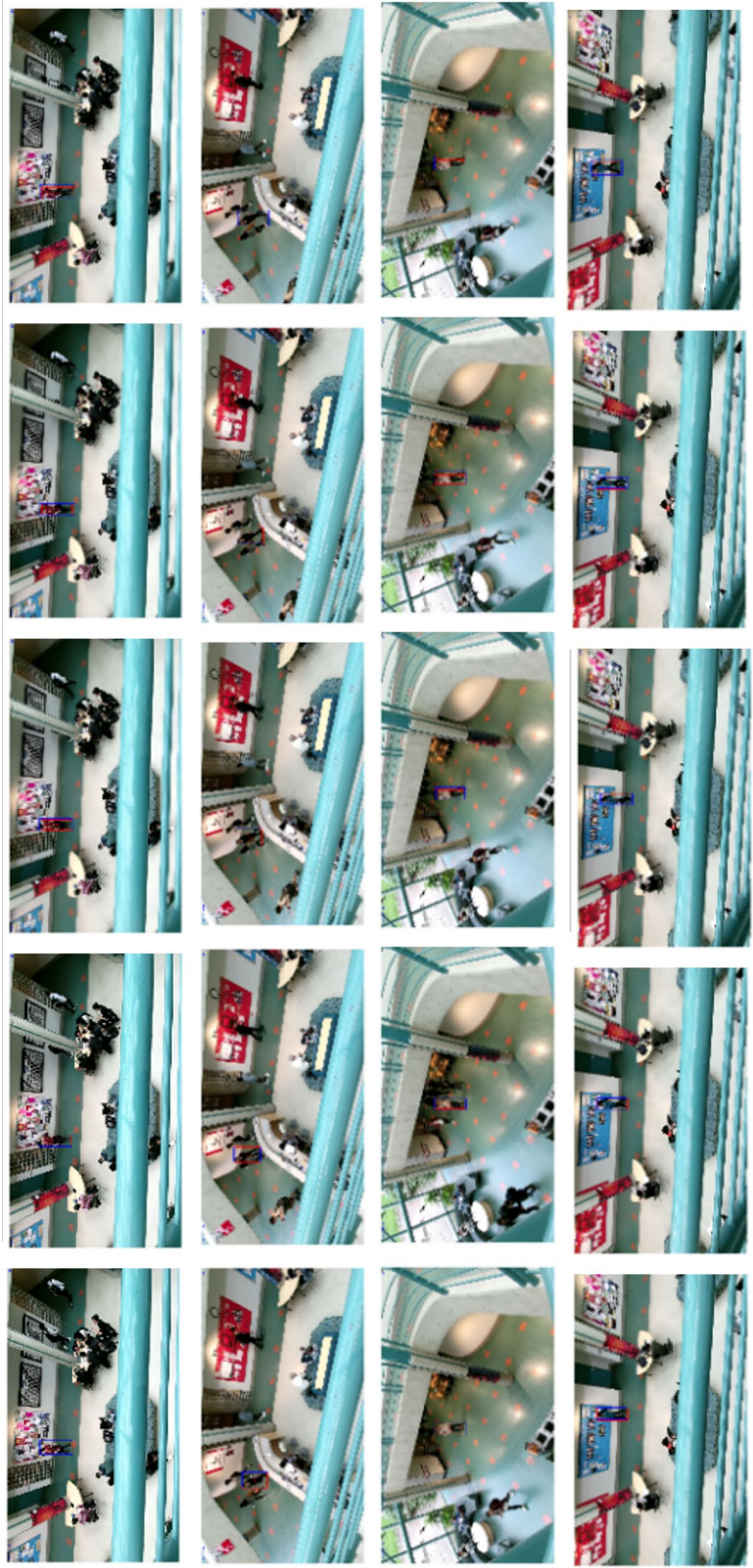

**(b) Deformation (blue is the ground-truth, red is the result from the KCF tracker). Best viewed in color.**

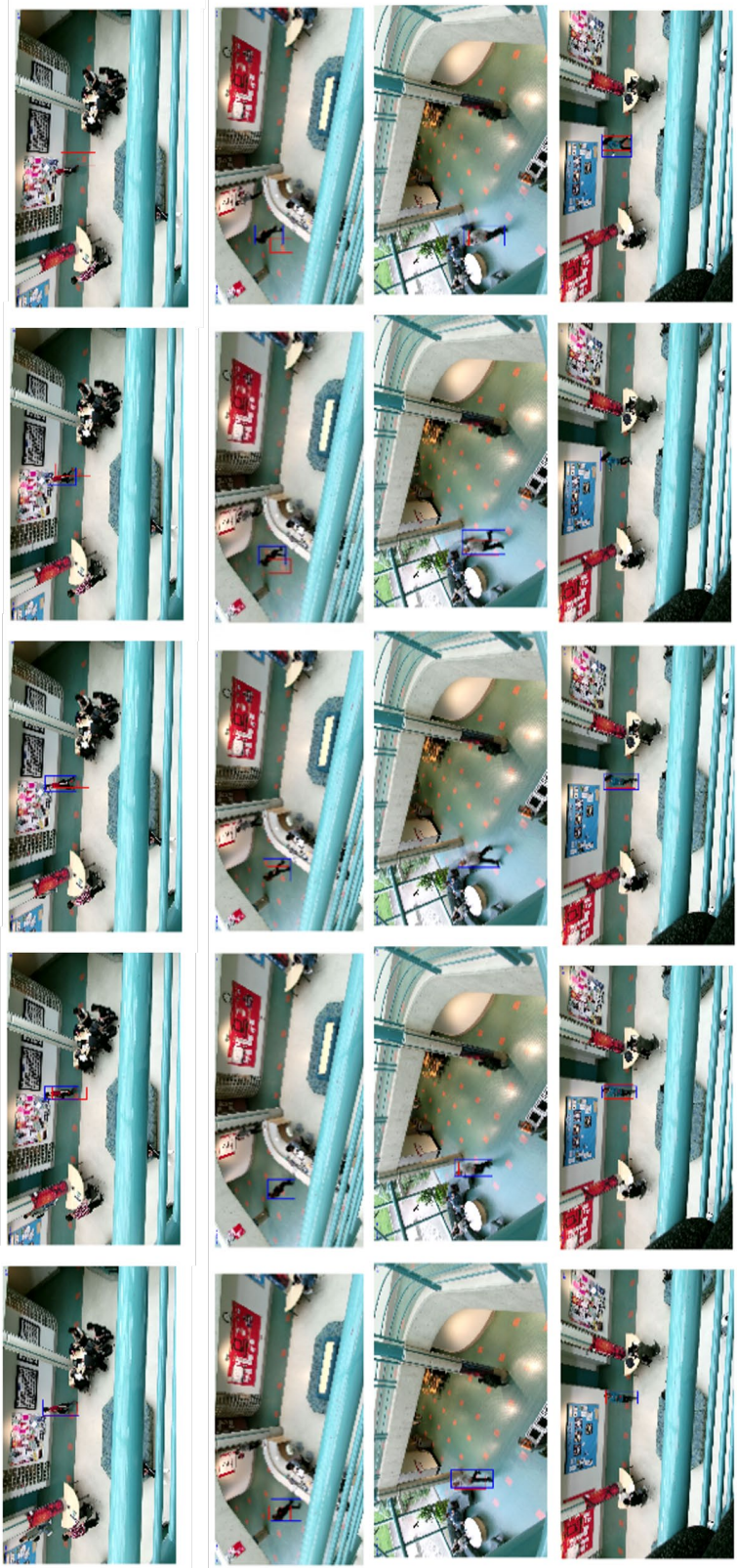

(c) Motion (blue is the ground-truth, red is the result from the KCF tracker). Best viewed in color.

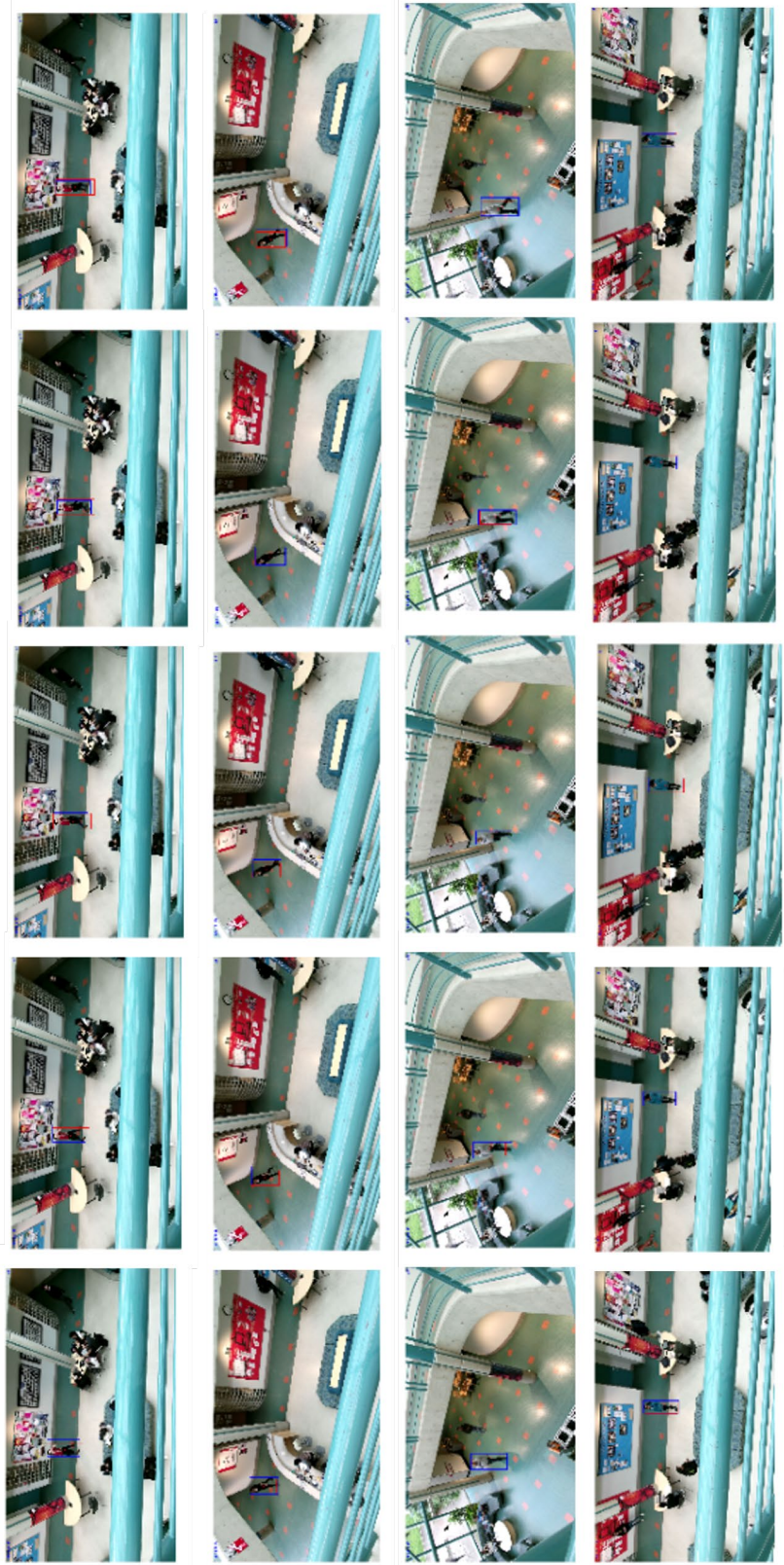

**(d) Normal (blue is the ground-truth, red is the result from the KCF tracker). Best viewed in color.**

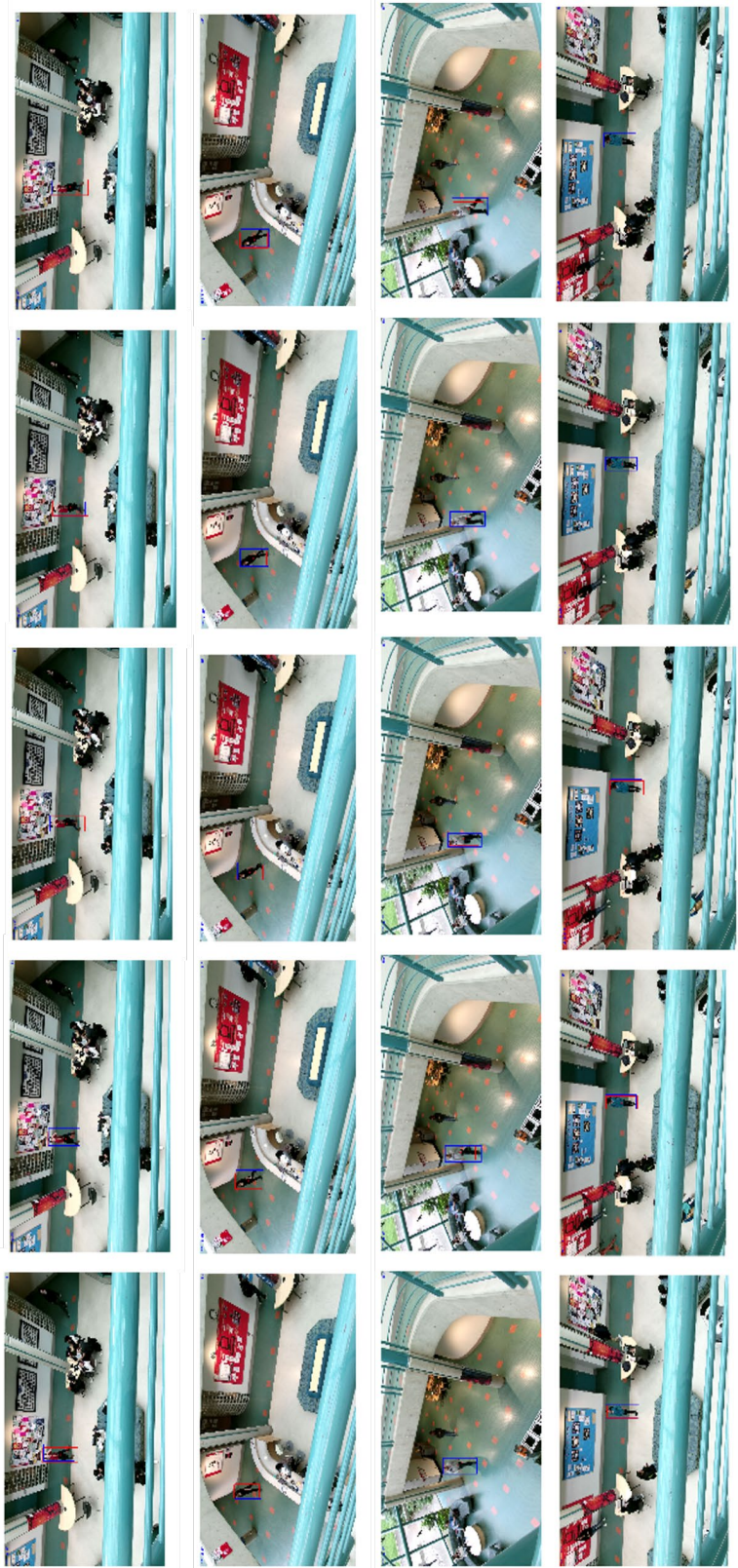

**(e) Occlusion (blue is the ground-truth, red is the result from the KCF tracker). Best viewed in color.**

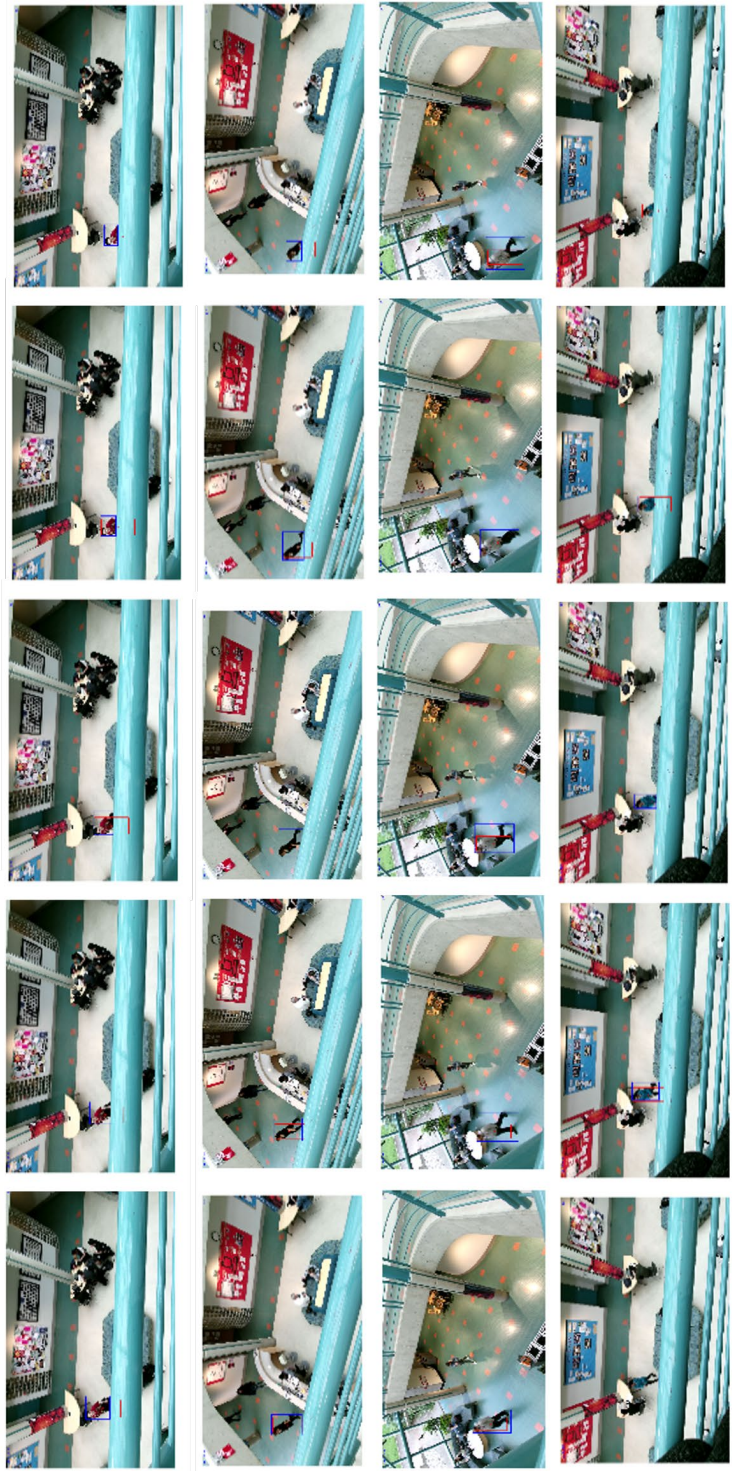

**(f) Out of view (blue is the ground-truth, red is the result from the KCF tracker). Best viewed in color.**

**Figure 3.8.** Overview of the sequences used in the experiment. A total of 6 scenarios experimented a.) Clutter b.) Deformation c.) Motion d.) Normal e.) Occlusion f.) Out-of-view

## 3.4. OTB-50 Benchmark Analysis

The OTB [26] methodology applies a no-reset experiment in which the tracker is initialized in the first frame and it runs unsupervised until the end of the sequence. KCF is compared with other existing top-performing trackers available for comparison on OTB50 [26] sequences that contain more than 26000 frames. A standard evaluation protocol is used to evaluate the 11 appearance variants. The results are presented in the form of a Success Rate (SR). For tracker bounding box $r_t$ (most tight fitted area which encapsulates the target) and the ground truth bounding box $r_0$ of the target, success rate (SR) is given by:

$$SR = \frac{|\ r_t \ \cap \ r_0|}{|\ r_t \ \cup \ r_0|} * 100$$

where ∩ and ∪ represents the intersection and union operators. We can read SR as IoU metrics but in %. A high success rate implies that the tracker successfully tracks the target and vice versa. The 11 appearance variants or categories are: illumination variation (IV), scale variation (SV), occlusion (OCC), deformation (DEF), motion blur (MB), fast motion (FM), in-plane rotation (IPR), out-of-plane rotation (OPR), out-of-view (OV), background clutter (BC), and low resolution (LR). For comparison purposes, we take state-of-trackers as mentioned in Table 3.3. KCF [21] and DSST [43] are the most popular correlation filter-based trackers, while trackers like Struck [14] are the frequently used tracking-by-detection based algorithms. There are other CF trackers that address many model issues like unwanted boundary and model degradation (DDCF [100]), scale change of target objects (SAMF [22]), and drifting and long-tern failure (ELMACF [101]). With the increased popularity of attention mechanisms, trackers like AFCN [102] utilized the dynamic properties of targets for best performance.

 **The SR score of the several state-of-art trackers, including KCF, on the 11 challenging scenarios. The best performance in each tracker has been highlighted**

|        | IV   | IPR  | LR   | OCC  | OPR  | OV   | SV   | MB   | FM   | DEF  | BC   |
|--------|------|------|------|------|------|------|------|------|------|------|------|
| ELMACF | 55.7 | 59.1 | 38.1 | 62.5 | 62.2 | 51.3 | 52.8 | 58.1 | 54.0 | 64.0 | 59.6 |
| ACFN   | 55.7 | 56.5 | 35.2 | 60.4 | 60.0 | 62.4 | 59.2 | 52.1 | 52.7 | 63.2 | 54.6 |
| FCN    | 59.8 | 55.5 | 51.4 | 57.1 | 58.1 | 59.2 | 55.8 | 58.0 | 56.5 | 64.4 | 56.4 |
| DDCF   | 63.8 | 48.8 | 44.8 | 44.4 | 46.1 | 44.3 | 48.3 | 53.3 | 52.3 | 41.2 | 54.5 |
| SAMF   | 46.3 | 45.8 | 44.0 | 47.8 | 48.1 | 39.5 | 44.6 | 44.0 | 42.8 | 44.0 | 43.8 |
| DSST   | 51.7 | 44.0 | 37.3 | 43.2 | 40.2 | 32.3 | 41.7 | 40.5 | 36.6 | 40.9 | 49.1 |
| KCF    | 43.3 | 38.9 | 28.5 | 39.6 | 39.6 | 32.7 | 35.3 | 40.8 | 38.9 | 40.0 | 41.7 |
| Struck | 33.0 | 37.6 | 31.9 | 33.2 | 33.5 | 32.7 | 35.9 | 39.9 | 40.4 | 32.3 | 35.6 |

# 3.5. VOT 2015 Performance Analysis

VOT 2015 [51] challenge saw various competitive trackers tested against a diverse dataset. Here, many variants of correlation filter-based trackers and some specifically around KCF tracker proved their competitiveness, robustness, and speed. VOT 2015, three performance measures were used due to their understandability and interpretability (a) accuracy (b) robustness, and (c) speed. The accuracy measures how well the bounding box predicted by the tracker is similar to the ground truth annotation for the dataset. On the other hand, robustness measures how many times the tracker loses the target. In particular, the raw accuracy and raw robustness (number of failures per sequence) were computed for each tracker on each sequence. They were then quantized into the interval [0, 9]. Raw robustness was clipped at nine failures to calculate the quantized robustness and the quantized accuracy was computed by $9 - [10\emptyset]$, where $\emptyset$ is the VOT accuracy.

It was observed that though CNN based correlation trackers did achieve great accuracy, however, traditional correlation filter (especially KCF variants) performed best in speed. The effectiveness of KCF based trackers can be understood from the fact that the speed of traditional KCF trackers (sKCF [51] and KCFv2 [51]) are almost 70 – 100 % faster than CNN based CF trackers. Table 3.4 shows various CF trackers and their results on the VOT 2015 challenge.

In VOT 2015, among other trackers, several were from the class of KCF tracker namely SRDCF [24], DeepSRDCF, LDP [48], NSAMF [51] RAJSSC [51] and MvCFT [51]. RAJSSC is a KCF based framework which focuses on making tracker rotation invariant, NSAMF is an extension of VOT 2014 top-performing tracker that uses color in addition to edge features, SRDCF is a regularized KCF that reduces boundary effects in learning a filter and DeepSRDCF is its extension of [24] that uses CNN for feature extraction. MvCFT address the problem of multiple object views using a set of correlation filter. LDP, on the other hand, addresses the non-rigid deformation by applying a deformable part-based correlation filter.

**Table 3.4.**    **Raw accuracy and raw robustness (defined by the average number of failures) and speed of various CF trackers. The best performance is highlighted in bold.**

| Tracker | Accuracy | Robustness | Speed |
|---------|----------|------------|-------|
| RAJSSC | **0.57** | 1.63 | 2.12 |
| SRDCF | 0.56 | 1.24 | 1.99 |
| DeepSRDCF | 0.56 | **1.05** | 0.38 |
| NSAMF | 0.53 | 1.29 | 5.47 |
| MKCF+ | 0.52 | 1.83 | 0.21 |
| MvCFT | 0.52 | 1.72 | 2.24 |
| LDP | 0.51 | 1.84 | 4.36 |
| KCFDP | 0.49 | 2.34 | 4.80 |
| MTSA-KCF | 0.49 | 2.29 | 2.83 |
| sKCF | 0.48 | 2.68 | 66.22 |
| KCF2 | 0.48 | 2.17 | 4.60 |
| KCFv2 | 0.48 | 1.95 | **10.90** |

## 3.6.  VOT 2019 Performance Analysis

VOT 2019 [103] was different than previous challenges as evaluation included the standard VOT and other popular methodologies for both short term tracker and long term tracker in a diverse dataset. The metrics for analysis were similar to VOT 2015. Table 3.5 shows the performance results of the correlation filter-based trackers on the VOT 2019 challenge. Expected average overlap (EAO) is an estimator of the average overlap a tracker is expected to attain on a large collection of short-term sequences with the same visual properties as the given dataset. Readers are referred to [51]  for further details on the average expected overlap measure.

LSRDFT [103] and TDE [103] are CNN based deep tracking approach equipped with deep features. They differ in the way that LSRDFT uses a shortened interval of updating in the correlation filter, TDE utilizes an adaptive spatial selection scheme to learn a robust model. SSRCCOT is a correlation filter based tracking method that proposes selective spatial regularization while training continuous convolution filters. CSRDCF [104] (and its C++ implementation CSRcpp) improves DCF trackers by introducing spatial and channel reliability. Its variant CISRDCF differs from the CSRDCF by independent feature channel calculation and iterative regularization process. Another CFT tracker ECO [105] saw two variants FSC2F and M2C2F in the comparative analysis. Where former employs a motion-aware saliency map to address robustness, later adaptively utilizes multiple representative models of the tracked object for robustness. The issue of boundary effect was attempted to be addressed by WSCF-ST [103]. An ensemble of CF trackers also proved its robustness. TCLCF [103] which is an ensemble CF tracker uses a different correlation filter to track the same target. Being computationally faster, it is great for embedded systems. There are other trackers that are either variant of KCF (Struck [14]) or use KCF in some way (DPT [49]) in their algorithm.

**Table 3.5.** The table shows expected average overlap (EOA), raw accuracy (A), and raw robustness (R) for the experiments in the VOT 2019 challenge. The best performance is highlighted in bold.

|  | baseline | | | real-time | | |
|---|---|---|---|---|---|---|
|  | EOA | A | R | EOA | A | R |
| LSRDFT | **0.317** | 0.531 | **0.312** | 0.087 | 0.455 | 1.741 |
| TDE | 0.256 | **0.534** | 0.465 | 0.086 | 0.308 | 1.274 |
| SSRCCOT | 0.234 | 0.495 | 0.507 | 0.081 | 0.360 | 1.505 |
| CSRDCF | 0.201 | 0.496 | 0.632 | 0.100 | 0.478 | 1.405 |
| CSRpp | 0.187 | 0.468 | 0.662 | 0.172 | 0.468 | **0.727** |
| FSC2F | 0.185 | 0.480 | 0.752 | 0.077 | 0.461 | 1.836 |
| M2C2F | 0.177 | 0.486 | 0.747 | 0.068 | 0.424 | 1.896 |
| TCLCF | 0.170 | 0.480 | 0.843 | **0.170** | 0.480 | 0.843 |
| WSCF-ST | 0.162 | 0.534 | 0.963 | 0.160 | 0.532 | 0.968 |
| DPT | 0.153 | 0.488 | 1.008 | 0.136 | **0.488** | 1.159 |
| CISRDCF | 0.153 | 0.420 | 0.883 | 0.146 | 0.421 | 0.928 |
| KCF | 0.110 | 0.441 | 1.279 | 0.108 | 0.440 | 1.294 |
| Struck | 0.094 | 0.417 | 1.726 | 0.088 | 0.428 | 1.926 |

# Chapter 4.

# Depth Augmented Target Re-Detection

A significant improvement over existing CF-based trackers came with data augmentation using depth sensors [22], [58]. Additionally, the popularity of affordable depth sensors (Intel RealSense [106], LiDAR [107]), has made depth acquisition easily available. Using depth data in RGB based visual tracking is not a new method, however, past works [69] [54] have shown that depth feature can complement RGB image features and significantly improve tracking results with robust occlusion and model drift handling. Most of these trackers are holistic in nature, depending on how the feature space is expressed and captured by a single patch. Sliding window is a common technique to localize the object in the search space. Despite the progress made in tracking robustness, most works like [57] [60] [108] mostly focus on scale adaptation, occlusion detection, or shape change. A big part of long-term tracking is not only detection of occlusion but also re-detection of the target once it is out of the occlusion and track It in real-time continuously.

Motivated by the analysis done in Chapter 2 which shows the need for a robust tracker to address the issue of occlusion and informed of the advantages of using depth information in re-detecting the target, this chapter discusses this challenge of re-detection of a single occluded target. Our work specifically focuses on human targets that are occluded by objects (e.g. chairs) or by other humans (e.g. a tall person). It discusses the implementation of architecture that can help overcome this challenge. The tracking methodology proposed considers a single Kinect RGB-D camera, single-target, and is model-free, applied to long-term tracking. The model-free property means that the only supervised training example is provided by the bounding box in the first frame. The long-term tracking means that the tracker learns to re-detection after the target is lost. i.e. to infer the object position in the current frame.

A key advantage of the proposed tracker is that the depth information used by the tracker is intelligently adapted to avoid boundary issues. Our tracker, similar to KCF, uses the ROI specified in the first frame to initialize the tracker. However, KCF only uses image features of the RGB image within the ROI but our proposed tracker uses both image features and depth information of the ROI. Due to the inclusion of depth information, this approach can lead to boundary issues since the depth data is expected to change at the edge when the tracker is in motion, adversely affecting the tracking performance. However, the depth information in the proposed tracker is intelligently adapted to incorporate correct depth information of the moving tracker as discussed in Section 4.1.3. The tracker was validated on the Princeton RGB-D dataset as well as the real-time dataset collected by the author. The results show that the tracker can successfully detect when the human target is occluded and re-detect it after occlusion.

The chapter is organized as follows: Section 4.1 is the tracking pipeline of our proposed tracker. It details the data augmentation process, tracking, and detection methodology. It is followed by two experimental evaluation sections; Section 4.2 discusses the results of our tracking algorithm as computed on the Princeton RGB-D dataset and Section 4.3 which discusses our results as observed on the dataset collected by the author. This chapter concludes in Section 4.4 where we discuss the advantages of using depth information in the tracker and its potential for improving the tracker.

## 4.1. Tracking Pipeline

### 4.1.1. Data Augmentation using Depth Features in KCF

KCF tracker was originally implemented for RGB data. Hence, the dataset on which KCF was originally evaluated can not be used for testing the proposed RGB-D algorithm. Hence, to evaluate our algorithm, we use the Princeton RGB-D dataset. The dataset is also very diverse including examples of a) occlusion b) speed c) and size d) deformability which helps test the algorithm over a wide range of possible challenges.

In the Princeton RGB-D dataset, images are in 16-bit PNG format. Values at each pixel are the distance from Kinect to the object in mm. A sample image with RGB and its depth data can be seen in Figure 4.1 a) and b). We can create a depth view of the sample image using the RGB-D data where $x, y$ define pixel location and $z$ the distance from the sensor. Depth is when an image is viewed as seen from $z$ axis as shown in Figure 4.1 b) shows the depth image available in the dataset and Figure 4.2 c) shows it color-coded w.r.t. the depth value of each pixel. Depth view image helps us visualize the images as they are seen from the depth sensor.

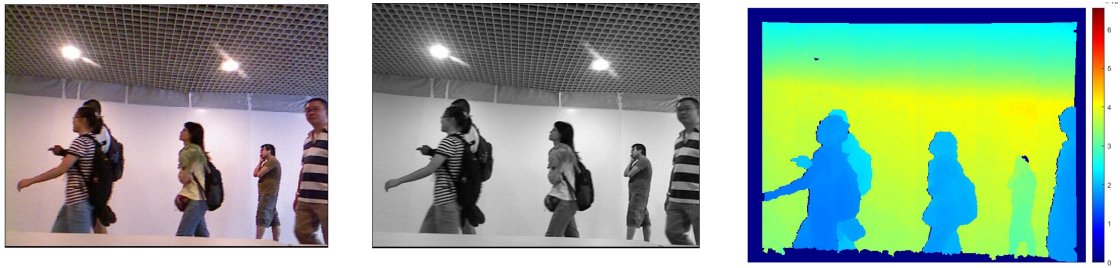

**Figure 4.1.** **a.) Sample RGB image from Princeton Dataset b.) Corresponding depth image c.) depth colormap from the Depth data**

Apart from benchmark comparison (on Princeton dataset), we collected our own RGB-D dataset for further evaluation. This dataset is collected using Microsoft Kinect V2. Figure 4.15 (at the end of Section 4.3) gives detailed information on the type of scenes and their respective samples

## 4.1.2. Combining RGB and Depth Features

As discussed in Chapter 2, KCF tracker achieves faster throughput by substituting convolutions in the spatial domain with element-wise multiplication in the frequency domain using RGB color features. However, these color features do not encode complete spatial information of the target. Depth information can provide additional spatial information in the form of its distance from the camera. This additional information forms the basis of our occlusion detection and re-detection framework. It is based on the assumption that an occluding object will have a different depth information as compared to the target being occluded. The current occlusion knowledge is computed using the information from around the target's center (as explained in Section 4.1.3). It is so because the information at the edges of the bounding box tends to include small false information but the target center is expected to be constant. Since the computation is made on a small section (the area around the target center) of the target bounding box area, it improves computational efficiency. The RGB information helps the tracker to update the model template (updated in the Fourier domain) and depth information helps it to decide when to track. Our proposed tracker stops tracking at the moment the target is occluded. Hence, it adds to the tracker's robustness as the model and its coefficients are interpolated correctly.

### 4.1.3. Creating Self-Adjusting Depth Patch

In any tracking algorithm involving depth data, ensuring that the correct depth of the target is included is very important. KCF with depth data takes the RGB image features and depth information of the target patch using the bounding box used for detection. As the target moves and changes its position or scale, there is a high probability that the bounding box (of the target) starts including more depth data of the background from the edges. This would falsely provide higher background depth, as compared to (only) target depth, making it harder for the model to make the right predictions.

Our tracker proposes a self-adjusting depth patch as a solution to this problem. Here, it saves the depth information of the previous (target) patch and the new (target) patch and then locates the possible center of these depth patches by taking the area around the center of the patch (calculated using position coordinates). Figure 4.2 (a) and (b) show the difference between a frame and a patch. Figure 4.2 (c) shows how the depth patch around the center of two patches are extracted.

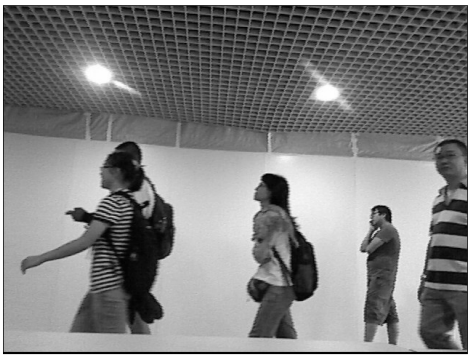

(a) Frame (grayscale)

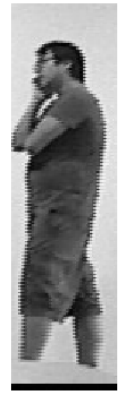

(b) Patch from the frame

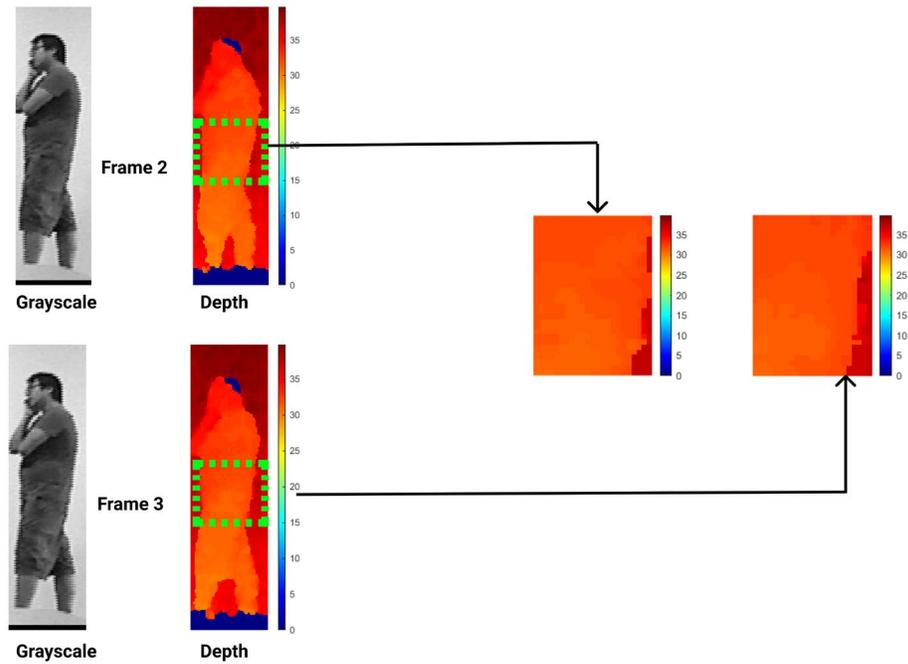

(c) Depth patch around the center of two patches of two subsequent frames

**Figure 4.2.** **Figure shows (a) grayscale frame (b) patch from the grayscale frame (c) the depth of the patch around the center of two patches (of two subsequent frames.**

The tracker further computes the difference in the depth information of these two parts of patches and looks for any high peaks at the edge of this value as shown in Figure 4.3. The edges are defined as few columns on the extreme end.

**Peaks due to depth from the background which need to be removed from computation**

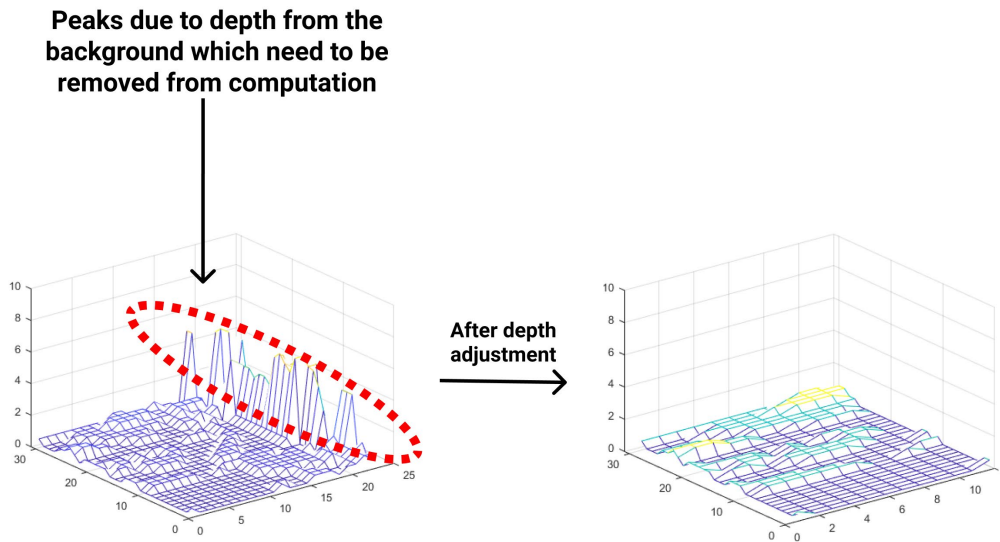

**After depth adjustment**

(Left) Before depth adjustment: Mesh graph of difference of centre depth patch for frame 2 and frame 3, x = width, y = height, z = depth value

(Right) After depth adjustment: Mesh graph of difference of centre depth patch for frame 2 and frame 3; x = width, y = height, z = depth value

**Figure 4.3.    The figure shows how the difference of center depth patch is computed for two subsequent frames when the target is not occluded. The image frame and the patch considered here are same as in Figure 4.2**

For a non-occluding object, any abnormally high peak at the edges would indicate the inclusion of background depth information (which is undesired). This is based on the assumption that two depth images in two subsequent frames will have similar depth data around their center, hence their difference should be minimal or close to zero. If there is a high peak, we decrease the size of the depth patches from the edges such that abnormally high peaks are removed as shown in Figure 4.3. After the depth has been adjusted, the difference of the depth is minimal as shown in Figure 4.4. This ensures that we have correct depth information for the target.

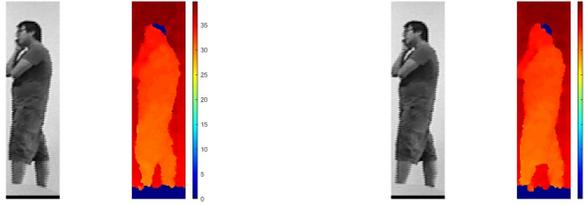

Frame 2: Grayscale and Depth    Frame 3: Grayscale and Depth

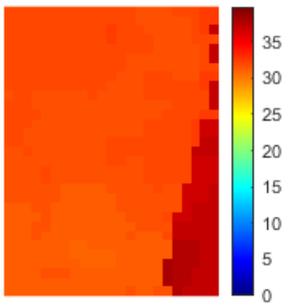
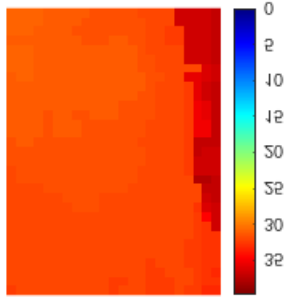
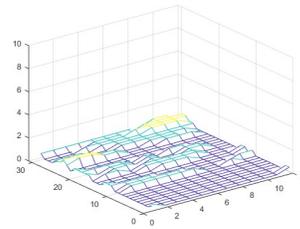

(a)

Centre depth patch:
Depth of the area around target patch center) extracted from the depth data of the target patch of frame 2

(b)

Centre depth patch:
Depth of the area around target patch center) extracted from the depth data of the target patch of frame 3

(c)

(After depth adjustment) Difference for center depth patches of two (non-occluding) frames (Frame2 and Frame 3). We see no peaks in their difference after depth adjustment.

**Figure 4.4.**    **The figure shows how difference of center depth patch changes for two subsequent frames when the target is not occluded**

However, in the case of occlusion, even after adjusting this difference in depth patch of subsequent frames, peaks will exist because the peaks will cover far more than area at extreme edges (with few peaks almost towards the center of the patch) as shown in Figure 4.5. The existence of these peaks helps the tracker identify the occurrence of occlusion in the frame.

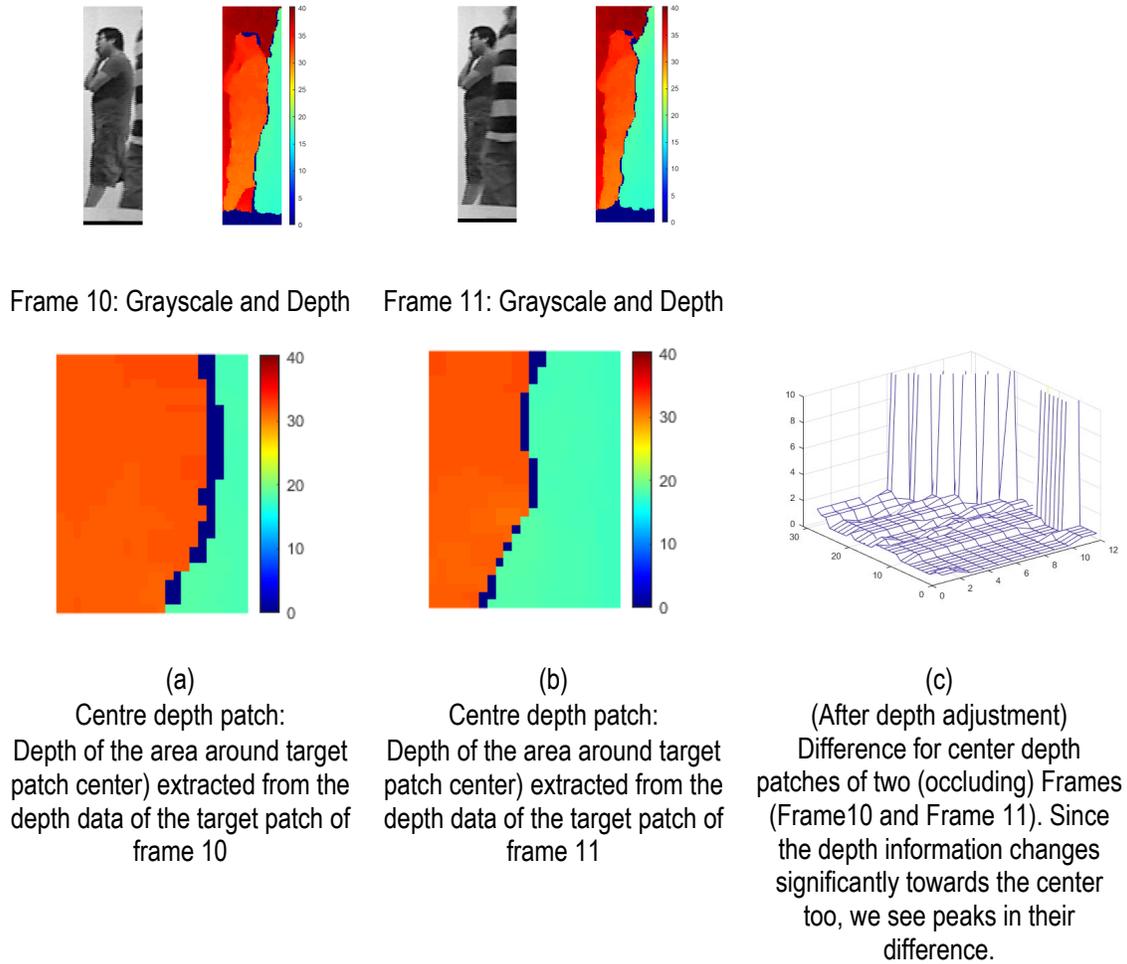

Frame 10: Grayscale and Depth    Frame 11: Grayscale and Depth

| (a) | (b) | (c) |
|---|---|---|
| Centre depth patch: Depth of the area around target patch center) extracted from the depth data of the target patch of frame 10 | Centre depth patch: Depth of the area around target patch center) extracted from the depth data of the target patch of frame 11 | (After depth adjustment) Difference for center depth patches of two (occluding) Frames (Frame10 and Frame 11). Since the depth information changes significantly towards the center too, we see peaks in their difference. |

**Figure 4.5.    The figure shows how difference of center depth patch changes for two subsequent frames when the target is occluded**

This computation is performed for all images and detection of occlusion helps the tracker make decisions in the detection pipeline as discussed in the next section, Section 4.1.4.

## 4.1.4. Training, Detection, and Re-detection of Target under Occlusion

Our proposed RGB-D tracking with re-detection algorithm builds upon the KCF tracker. KCF uses image data features to locate and detect trackers. In our proposed tracker, we add depth information to provide the tracker with contextual information of the target and the background, for it to be able to distinguish between the two.

## Training Pipeline

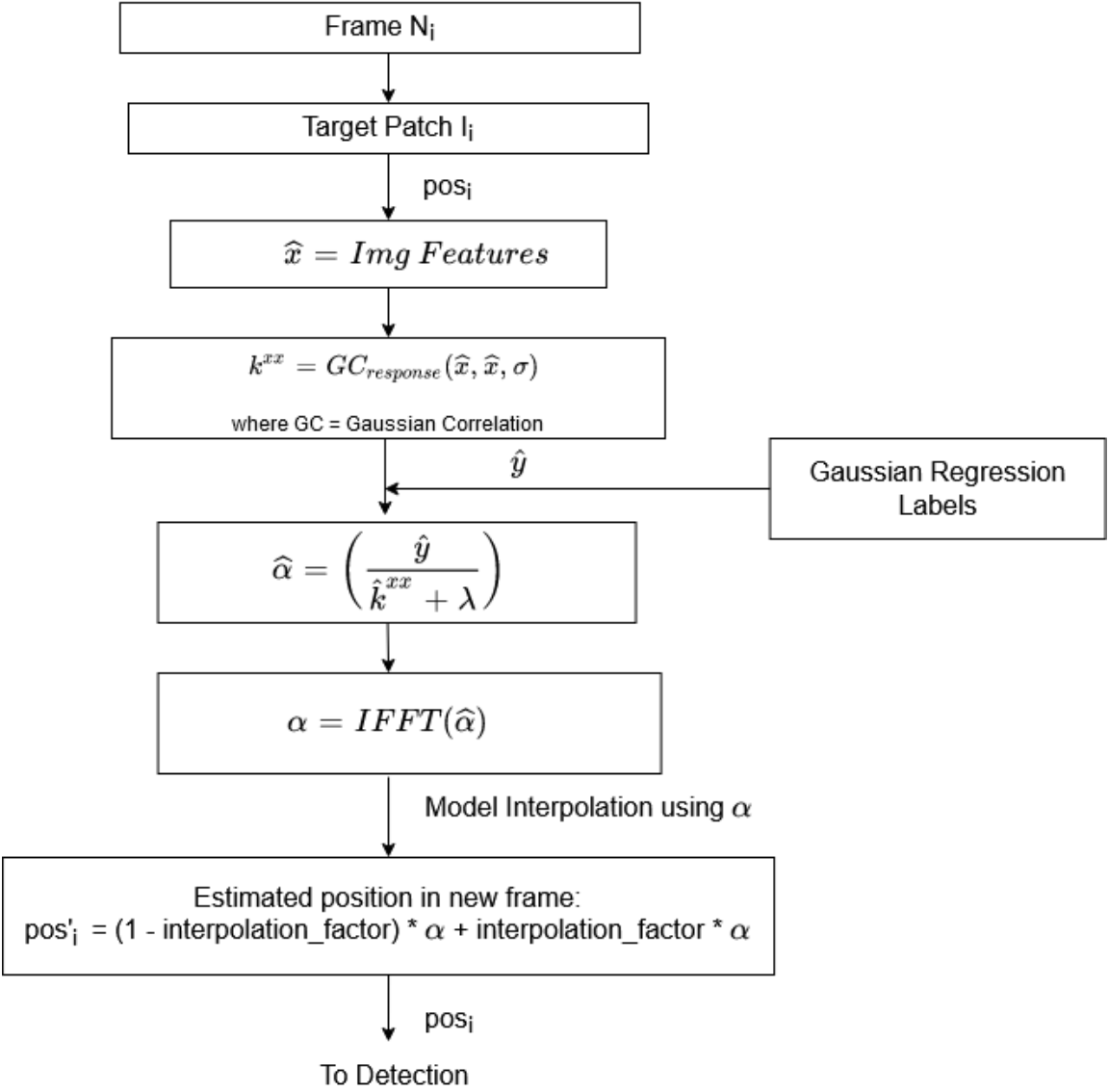

(a) **Tracking pipeline of the RGB-D tracker**

**Detection Pipeline**

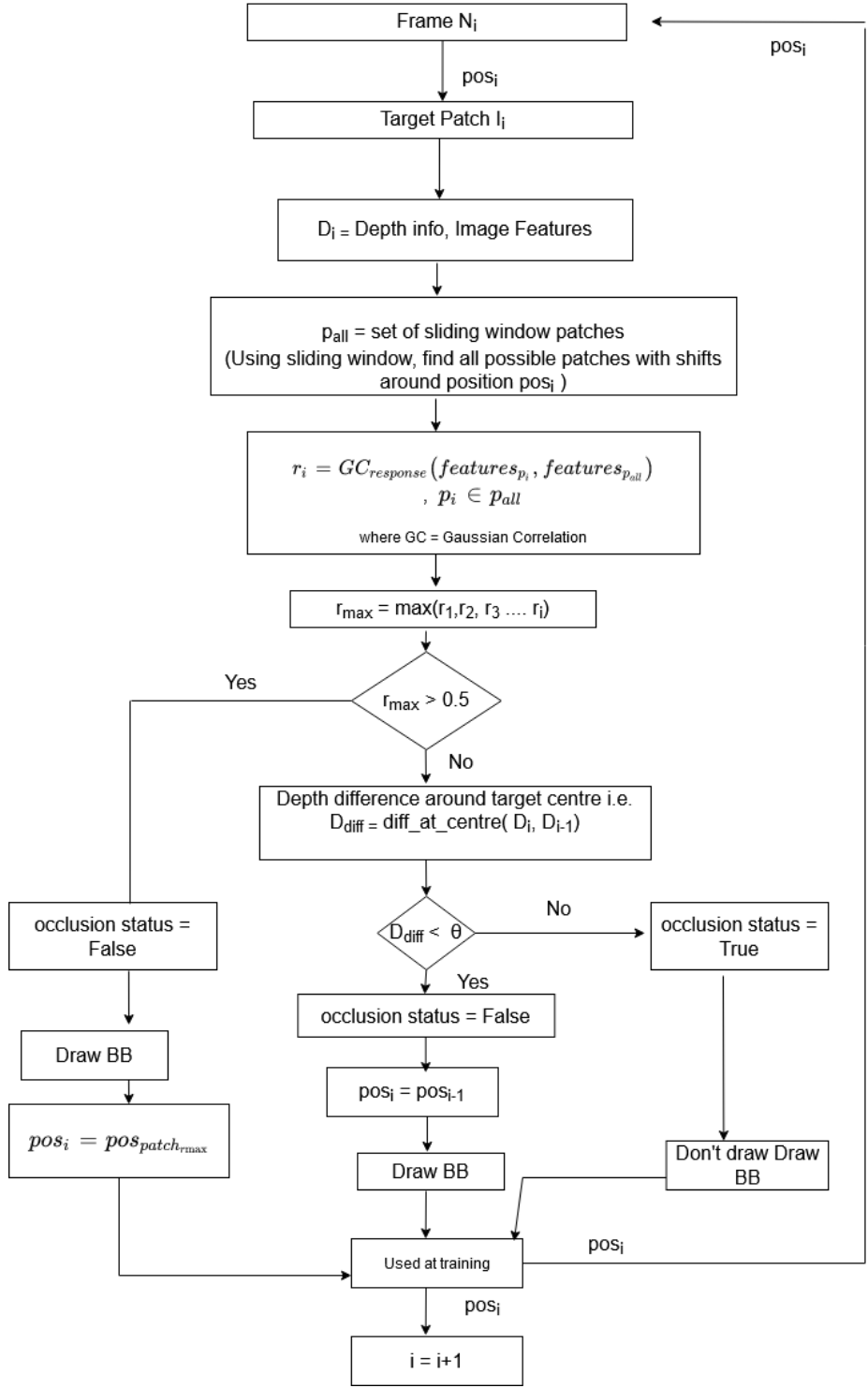

**(b) Detection pipeline of the RGB-D tracker**

**Figure 4.6.     Flowchart of the training and detection pipeline of the RGB-D tracker**

The tracking pipeline for the proposed RGB-D tracker is shown in Figure 4.6 and is also detailed in Algorithm 1. In Figure 4.6, the symbols defined are as follows:

| | |
|---|---|
| $\sigma$ | standard deviation |
| $\lambda$ | regularizer |
| $\alpha$ | learnable parameter |
| $f(z)$ | The response of kernel ridge regression |
| $patch_{rgb}$ | part of the RGB image which has the target, as shown in Figure 4.2b |
| ^ | FFT of a variable |
| $Occlusion_{status}$ | status of the tracker which informs if the tracker is occluded (True) or not occluded (False) |

| kernel_correlation( $x$ , $z$, $\sigma$) |
|---|
| 1  Calculate $\hat{k}^{xz} = FFT\left(\exp\left(-\frac{1}{\sigma^2}\left(\|x\|^2 + \|z\|^2 - 2F^{-1}\left(\hat{x}^* \odot \hat{z}\right)\right)\right)\right)$ |
| 2  Return: k = IFFT ($\hat{k}^{xz}$) |

(a)

| Training (x, y, $\lambda$ ) |
|---|
| 1  kernel_correlation(( $x$ , z, $\sigma$) |
| 2  Compute $\hat{\alpha} = \left(\frac{\hat{y}}{\hat{k}^{xz} + \lambda}\right)$ |
| 3  Return $\alpha = IFFT(\hat{\alpha})$ |

(b)

| Detection ($x, z$, $\lambda$) |
|---|
| 1  $\hat{k}^{xz}$ = kernel_correlation(( $x$ , z, $\sigma$) |
| 2  Calculate $\hat{f}(z) = \hat{k}^{xz} \odot \hat{\alpha}$ |
| 3  Response = $f(z) = IFFT\left(\hat{f}(z)\right)$ |
| 4  Return max ($f(z)$), position of max ($f(z)$) |

(c)

| Algorithm 1: Tracker ($x_{RGB}, x_{Depth}$) |
|---|
| 1  Extract patch $patch_{rgb}$ from image $I_i$ |
| 2  Calculate $D_i$ = Depth of image $I_i$ |
| 3  *Training* ( $features_x$ , features$_x$, $\lambda$) where $features_x$ is the hog features of the target at $pos_i$. For frame 1, $pos_i$ is provided by ground truth, for subsequent images, is it it output of detection pipeline |

| Algorithm 1: Tracker ($x_{RGB}, x_{Depth}$) | |
|---|---|
| 4 | Interpolate the model with training output. Estimate target position at frame $i = i + 1$ |
| 5 | $i = i + 1$ . At $pos_i$, find all possible patches $p_{all}$ using a sliding window with shift $s$ around the target |
| 6 | Compute $gaussian\_correlation\big(\,\text{features}_{p_i}, \text{features}_{p_{all}}\,\big)$ for $p_i \in p_{all}$ |
| 7 | Find responses $r_i$ of the classifier at all shifts $s$ for $p_{all}$ |
| 8 | Find maximum response $r_{max}$ $\;s.t.\; r_{max} \in r_s = \{r_1, r_2, r_3, \dots r_{s_i}\}$ |
| 9 | Save ( $r_{max}$ , $patch_{r_{max}}$ ) where $patch_{r_{max}}$ is the patch which gives max response $r_{max}$ |
| 10 | **If**    $r_{max} > 0.5$ |
| 11 |          Target detection using Detection($\text{features}_{r_{max}}, \text{features}_{\text{pos}_1}, \lambda$) where $\text{features}_{r_{max}}$ = features of the patch which gives maximum response , $\text{features}_{\text{pos}_1}$ = features of the original patch |
| 12 |          Save position of max response |
| 13 |          $Occlusion_{status} = False$ |
| 14 | **else**    Calculate $D_{diff} = diff\_at\_centre\,(D_i, D_{i-1})$ where $diff\_at\_centre$ computes difference of depths at target's center (and adjust the depth as required) |
| 15 |          **if**    $D_{\text{diff}}$ < threshold |
| 16 |                   $Occlusion_{status} = False$ ; continue tracking (it may be a partly visible target), Go to Step 5 |
| 17 |          **else**    $Occlusion_{status} = True$ (i.e. the target is occluded), no target detection and no bounding box, Go to Step 5 |

(d)

**Figure 4.7.**    **Figure showing the RGB-D based tracking algorithm. (d) shows the full algorithm which uses modules shown in (a) the kernel correlation computation, (b) the training using kernel correlation module, and (c) detection using kernel correlation module**

For the first frame, the tracker is provided with the ground truth. The tracker trains on this frame using the ground truth data (target position) to interpolate this position to the next frame. With this knowledge of target location (from the previous frame), the tracker defines its search space around this target location, in the new frame. Search space is the area in which the tracker will attempt to locate the target as shown in Figure 4.8.

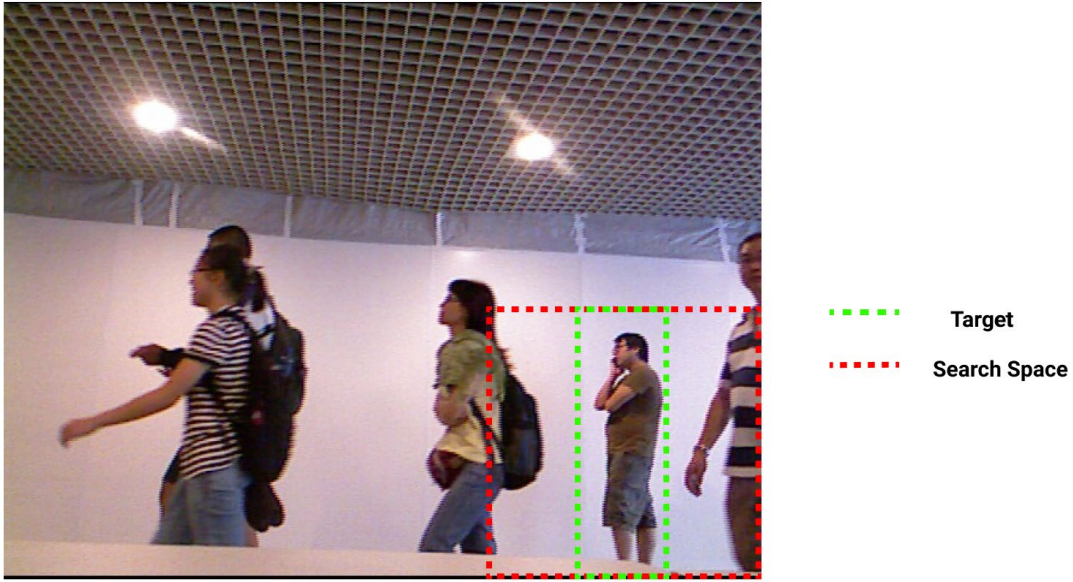

**Figure 4.8.** The figure shows the target patch and associate search space of the tracking pipeline. Search space is dependent on a.) the target location in the previous frame b.) width of the target

The tracker extracts all the possible patches from the space. The search space is dependent on the width of the tracker. If the tracker width is large, search space is larger, and vice versa. If we keep constant width of the search space, we might end up storing much higher number of patches for smaller target sizes and very few target patches for large target sizes, making further computations difficult. The search space is also limited to the horizontal plane since we don't expect the target to move vertically in space as shown in Figure 4.9.

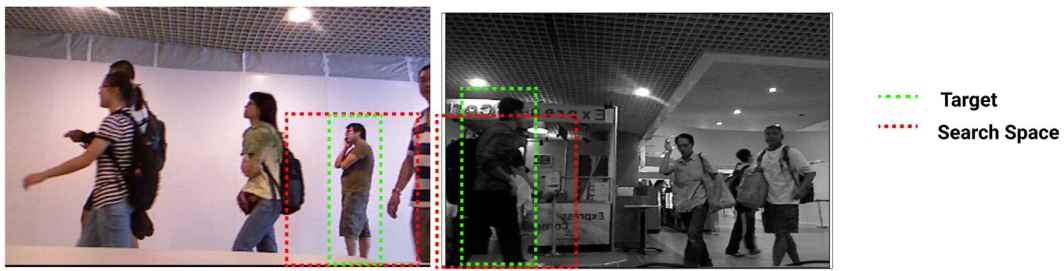

**Figure 4.9.** The figure shows the issue if the search space is kept consistent across different images, it would lead to incorrect search space when target size will vary

For detection in the new frame, the tracker now correlates these patches (with gaussian correlation similar to KCF) with the initial target patch, to get all possible correlation response. Mathematically, if $z$ and $x$ represent the features of the extracted patch and original target patch respectively, then we need to calculate $\hat{k}^{xz}$ and $\hat{\alpha}$ for fast training as represented by Equation 29 (re-written below):

$$\hat{f}(z) = \hat{k}^{xz} \odot \hat{\alpha}$$

where $\hat{\alpha} = \left( \frac{\hat{y}}{\hat{k}^{xx} + \lambda} \right)$ is the model parameter used for model interpolation, obtained and updated every time at the training stage as shown in Figure 4.7. For gaussian correlation, we can write $\hat{k}^{xz}$ as:

$$\hat{k}^{xz} = FFT \left( \exp\left( -\frac{1}{\sigma^2} \left( \|x\|^2 + \|z\|^2 - 2F^{-1} \left( \hat{x}^* \odot \hat{z} \right) \right) \right) \right) \tag{32}$$

where $\odot$ is the dot product

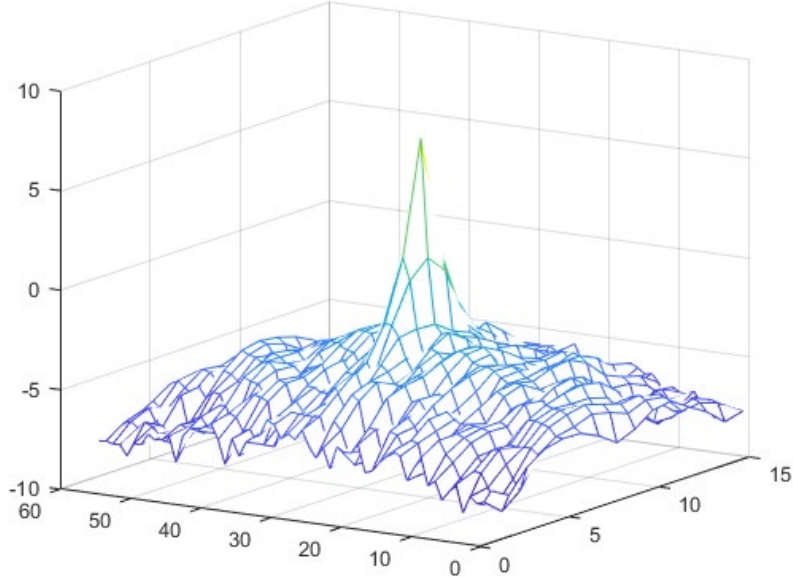

**Figure 4.10.**  **Gaussian correlation of the base patch of the current frame with the patch from previous frame**

$f(z)$ gives us the detection response and the maximum peak of this correlation response (highest correlation response) is observed as shown in Figure 4.11.

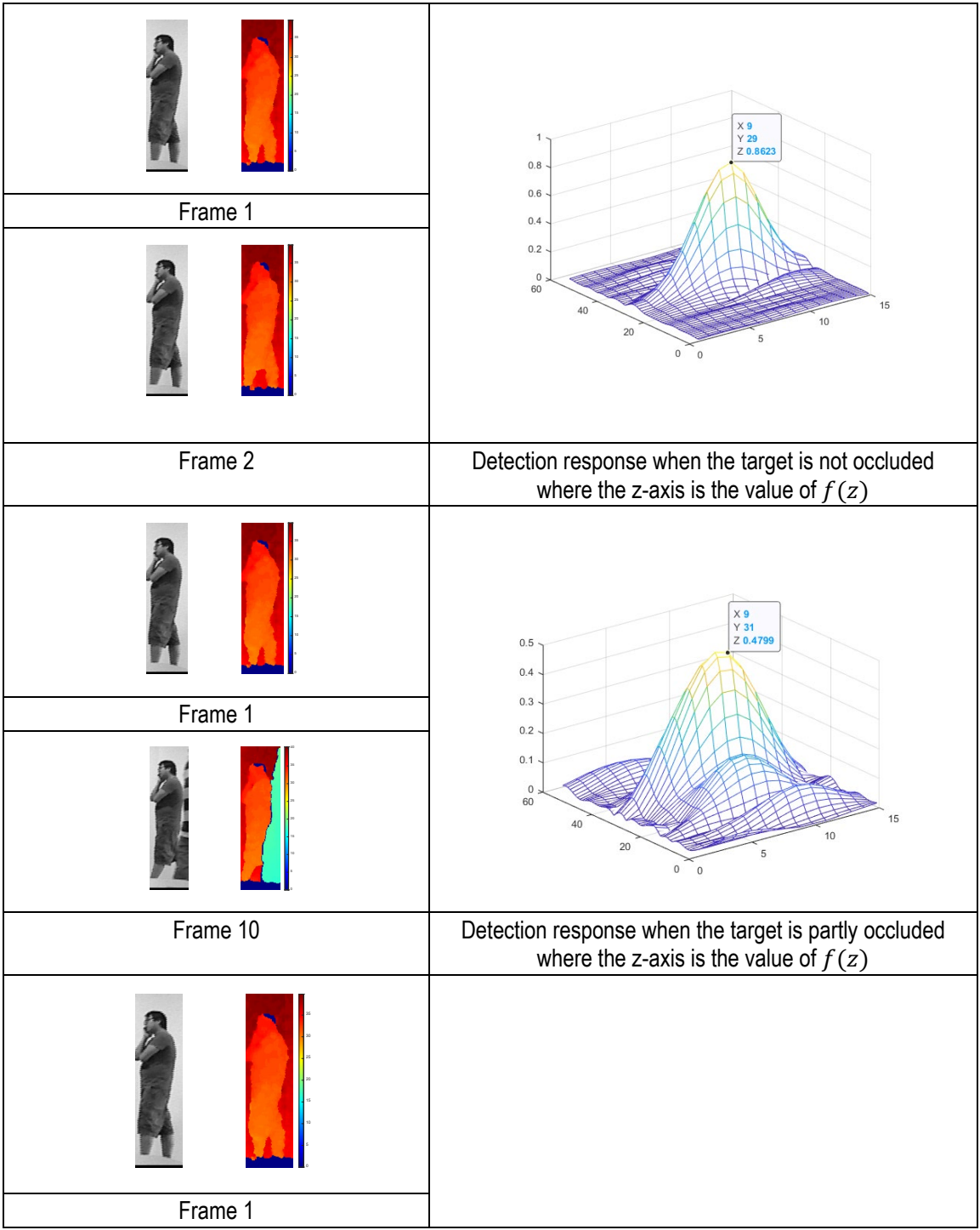

| | |
|---|---|
| Frame 1 | |
| Frame 2 | Detection response when the target is not occluded where the z-axis is the value of $f(z)$ |
| Frame 1 | |
| Frame 10 | Detection response when the target is partly occluded where the z-axis is the value of $f(z)$ |
| Frame 1 | |

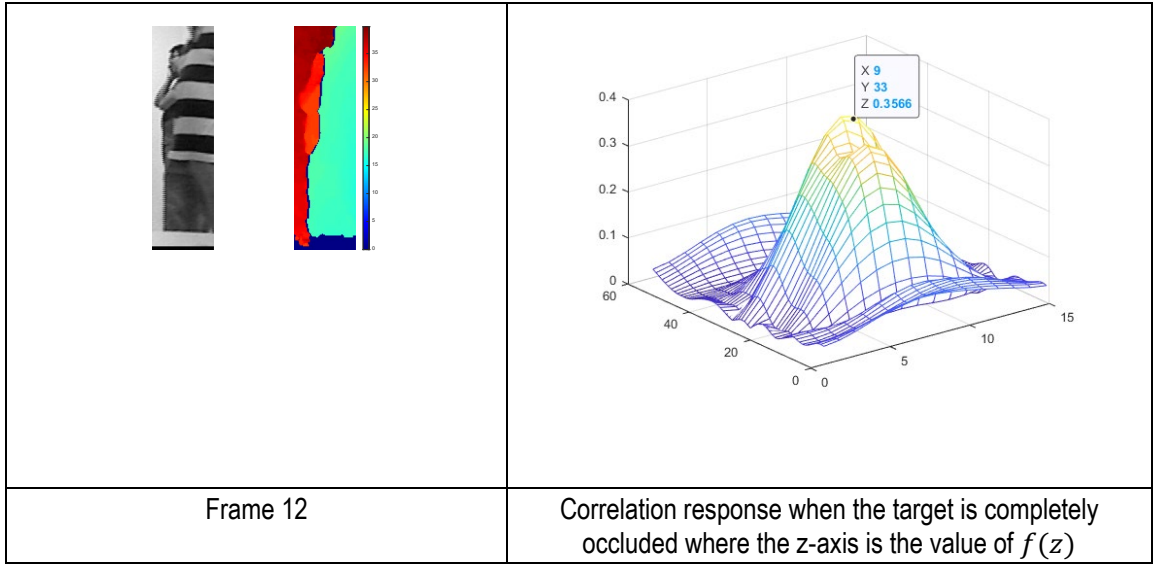

| Frame 12 | Correlation response when the target is completely occluded where the z-axis is the value of $f(z)$ |

**Figure 4.11.  Different detection response of two subsequent frames with the non-occluding and occluding target. Note the peak value changes from 0.8 (i.e. 80%) when it is not occluded to 0.4 (40%) when it is partly occluded to 0.3 (i.e. 30%) when it is fully occluded**

If the maximum detection response is less than 50%, there are two possible scenarios: a) tracker is partly visible b) tracker is occluded as shown in Figure 4.11. To confirm the status of the tracker, we compute the difference of the depth patches at frame $n-1$ and $n$ (at the target centre), as discussed in Section 4.2.3 and shown in Figure 4.5. If the difference of depth is minimal or close to zero, we know that target is not occluded, and the tracker will continue to track at the updated position computed using the detection score (maximum correlation response). However, if the difference of depth is high, we know the target is occluded. The tracker will, hence, stop tracking the target and will keep searching for the target in the search space. Once it locates the target (when a patch from the search space gives a correlation response > 50%), it will re-detect the target and continue to track it.

## 4.2. Evaluation of Princeton RGB-D Dataset

In this evaluation system, we use the Princeton data benchmark [54] to compare our tracker with KCF (our base tracker) and other trackers. This dataset uses 95 videos for evaluation. These datasets were originally captured using standard Microsoft Kinect 1.0. Due to Kinect's limit on the minimum and maximum distance for accurate depth accuracy, these videos contain an indoor environment with object depth values ranging from 0.5 to 10 meters.

This benchmark dataset has high diversity including single tracking subjects like humans, animals, faces, balls, etc. Figure 4.12 shows a few sample images from the Princeton dataset.

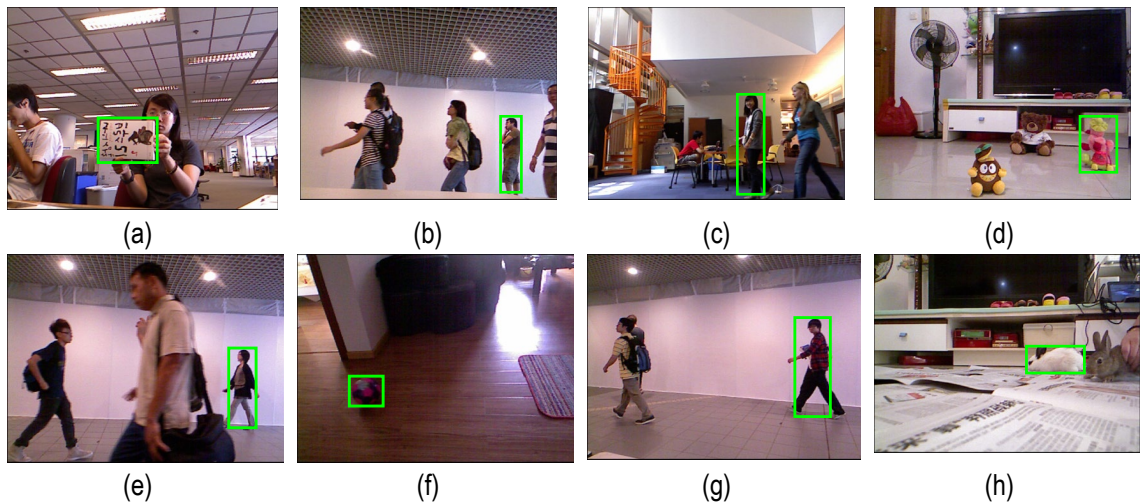

**Figure 4.12.** The figure shows few sample images with the target to be tracked (in green) from the Princeton Dataset. Best viewed in color

The various aspects of the dataset which show the diversity of samples is summarised as follows:

**Target Type**: The dataset contains three types of objects: human, animals, and rigid objects (example: toys and human faces which have the freedom to translate or rotate). Animal movements have out-of-plane rotation and some deformation. Tracking difficulty is expected o be slightly difficult for humans since the degree of freedom for human body motion is very high.

**Scene type**: Each scene has a different type of background. Some scenes like that of café and school are more complex with a lot of people moving around. Others like a turtle in a living room have a more static background.

**Occlusion**: Different possible occlusion scenarios are considered like, how long is the target occluded, whether the target moves while being occluded or is static when something occludes it, etc.

Other criterion that was considered was the bounding box distribution over all sequences and over time. Hence, a target in a sequence does not necessarily be in the center but can be anywhere in the frame at any time. Readers are directed to [54] for a more detailed analysis of the sequence distribution of the dataset.

## 4.2.1. Evaluation Metrics

To evaluate the overall performance, we test it on Princeton data. It doesn't explicitly provide ground truth bounding box values but provides a script to evaluate results as detailed in [54]. We report the success rate provided by Princeton data evaluation (criterion used in the [109]) which is the ratio of overlap between the outputs and true bounding boxes:

$$r_i = \begin{cases} area\left(ROI_{T_i} \cap ROI_{G_i}\right) & if\ both\ ROI_{T_i}\ and\ ROI_{G_i}\ exist \\ +1 & if\ neither\ ROI_{T_i}\ and\ ROI_{G_i}\ exist \\ -1 & otherwise \end{cases}$$

where $ROI_{T_i}$ is the target bounding box in the $i$-th frame and $ROI_{G_i}$ is the ground truth bounding box. By setting a minimum overlapping area $r_t$, they calculate the average success rate $R$ of each tracker as follows:

$$R = \frac{1}{N}\sum_{i=1}^{N} u_i \quad where \quad u_i = \begin{cases} 1 & if\ r_i > r_t \\ 0 & otherwise \end{cases},$$

Where $u_i$ is an indicator denoting whether the output bounding box of the $i$-th frame is acceptable, $N$ is the number of frames, and $r_t$ is the minimum overlap area defining whether the output is correct. Since some trackers may produce outputs that have a small overlap ratio overall frames while others give large overlap on some frames and fail completely on the rest, $r_t$ must be treated as a variable to conduct a fair comparison Table 4.1 shows the success rate of our tracker as compared to KCF and other trackers under different categorizations.

## 4.2.2. Results & Observation

**Table 4.1.** The table shows the success rate (%) of our tracker on Princeton Dataset. The evaluation compares the performance of our tracker against the KCF tracker and few other trackers. Trackers marked with (*) use RGB and Depth data; others use only RGB data.

| | Target Type | | | Target Size | | Movement | | Occlusion | | Motion Type | |
|---|---|---|---|---|---|---|---|---|---|---|---|
| | Human | Animal | Rigid | Large | Small | Slow | Fast | Yes | No | Passive | Active |
| Ours * | 41 | 37 | 42 | 46 | 40 | 56 | 35 | 47 | 51 | 52 | 46 |
| KCF [41] | 39.7 | 49.4 | 54.6 | 40.1 | 52.5 | 57.8 | 42.9 | 35.2 | 63.6 | 56.4 | 43.7 |
| Dhog* [54] | 43.3 | 48.3 | 55.9 | 47.2 | 50.3 | 52.7 | 47.5 | 38.4 | 63.5 | 54.3 | 46.9 |
| Struck [14] | 35.4 | 47 | 53.4 | 45 | 43.9 | 58 | 39 | 30.4 | 63.5 | 54.4 | 40.6 |
| VTD [110] | 30.9 | 48.8 | 53.9 | 38.6 | 46.2 | 57.3 | 37.2 | 28.3 | 63.1 | 54.9 | 38.5 |
| RGB [54] | 26.7 | 40.9 | 54.7 | 31.9 | 46 | 50.5 | 35.7 | 34.8 | 46.8 | 56.2 | 33.7 |
| CT [111] | 31.1 | 46.7 | 36.9 | 39 | 34.4 | 48.6 | 31.5 | 23.3 | 54.3 | 42.1 | 34.2 |
| PCflow* [54] | 35.2 | 29.1 | 43.6 | 42.2 | 33.2 | 47.2 | 33.1 | 32.4 | 43.5 | 41.3 | 35.5 |
| TLD [56] | 0.29 | 0.35 | 0.44 | 0.32 | 0.38 | 0.52 | 0.30 | 0.34 | 0.39 | 0.50 | 0.31 |
| MIL [112] | 32.2 | 37.2 | 38.3 | 36.6 | 34.6 | 45.5 | 31.5 | 25.6 | 49 | 40.4 | 33.6 |
| SemiB [113] | 0.22 | 0.33 | 0.33 | 0.24 | 0.32 | 0.38 | 0.24 | 0.25 | 0.33 | 0.42 | 0.23 |
| OF [54] | 18 | 11 | 23 | 20 | 17 | 18 | 0.19 | 0.16 | 0.22 | 0.23 | 0.17 |

Table 4.1 shows the success rate of our tracker as compared to other RGB and RGB-D trackers as evaluated on the Princeton dataset. From the evaluation, we observe that our tracker performs best for human targets that are large in size and for scenarios where the target is occluded.

When it came to target type, it performs best for humans, with a success rate of 41%, close to Dhog (43% success rate) and KCF (39.7% success rate). The tracker performs worse for animals. One possible explanation is that tracker is unable to extract unique features from animals, especially since the color features of the animals in the dataset are very similar to the background (example: a white rabbit moving on a white floor as shown in Figure 4.12). These animals are also very small making it harder for sufficient features to be extracted.

One of the main objectives of our tracker is to be able to track when the target is occluded. Our tracker shows positive results in the 'Occlusion' category. We observe that, when the target is occluded, the success rate of our tracker is significantly better than KCF. KCF (RGB tracker) has a success rate of 35.2% whereas our tracker (RGB-D tracker) has 47%, a jump of approximately 12%, validating our hypothesis that depth data can significantly improve our results. Our trackers also perform well from a few other RGB-D-based trackers like Dhog which has a success rate of 38.4% and PCflow with a success rate of 32.4%.

In target size, our tracker performs best for larger targets (success rate of 46%) very close to Dhog, another RGB-D based tracker (success rate of 47.2%). Both of them are significantly better than KCF which uses only image features for tracking. When the object is large, not only features are easily and sufficiently captured, the depth information is also large enough to track any significant changes in the depth of the target. Both these factors combined, make depth-based trackers better for larger objects.

It is observed that for 'Movement' and 'Motion Type', depth data can negatively affect the performance depending on the scenario. Depth information of an object will change significantly if an object rotates or move very fast. Image features are also adversely affected since a) fast-moving objects will be blurry and b) rotating objects will have different features at each frame (relative to the previous frame). Also, our tracker is not scale-invariant making the false positives even higher than other depth-based trackers in both these categories. Due to this reasoning, our current implementation of the tracker can not be generalized well on rotating objects and fast-moving objects (35% success rate as compared to 42% in KCF). Depth does help in identifying the target better, however, when the target is hidden or lost, the tracker attempts to locate the target in surrounding areas adding to a few false positives. For these reasons, during passive and active motions of the target, the performance of our tracker can be better or poor depending on a scenario as can be seen in Table 4.1. Current state of art for occlusion detection on Princeton dataset is 3D-T [114] which provides a success rate of approximately 70% for occlusion scenarios and works best for humans (81%) and weakest for animals (64%). It uses 3D based detection that exploits sparse representation, object parts as well as adaptive particle sampling and pruning, all in a unified framework

## 4.3. Evaluation on Real-time Kinect Dataset

To further test our tracker, we tested it on the data we collected in real-time. Princeton data has a mix of subjects with difference scales, color changes. We collected out data on less complex scenarios (almost an ideal testing ground) with human data, minimal speed, and occlusion cases. Since we propose a long-term tracker that is expected to perform better in occlusion and out-of-scene scenarios, we focus particularly on such scenes. To accommodate for diversity in the dataset, we choose subjects which have different size and different types of occluding objects (chair, box, and human).

**Data collection & Hardware Setup**: The data is collected using Microsoft Kinect V2 in an indoor home environment. The subjects vary in size and gender to accommodate for diversity in subjects. The objects used for occlusion are things like a chair, large box which is easily found in day to day life. The data collected assumes that the target is moving unidirectionally or bidirectionally in the horizontal plane. We do not consider scenarios where the target may tend to move towards and away from the sensor. They are useful in scenarios where the sensors are mounted in hallways (at a height at an angle) or mounted on stationary mobile robots which for majority part observe target moving from left to right or right to left. Hence, our data is a fair representation of various scenarios which are likely to occur in day to day lives. Due to COVID-19, there were few restrictions on who we can invite as subjects and where can we do the experiments. Despite these challenges, we made our best efforts to collect data in a reasonable setup. This data was collected using both Kinect for Linux and Kinect for MATLAB on Windows to accommodate for any software differences. There is a total of ~500 scenes with these different subjects and scenarios. Figure 4.13 shows few sample RGB images and their corresponding depth images from our dataset.

**Data Annotation**: All the data were manually annotated by the author to collect the ground truth. Each ground truth bounding box depicts the most tightly fitting box that can be drawn to have the target within the box.

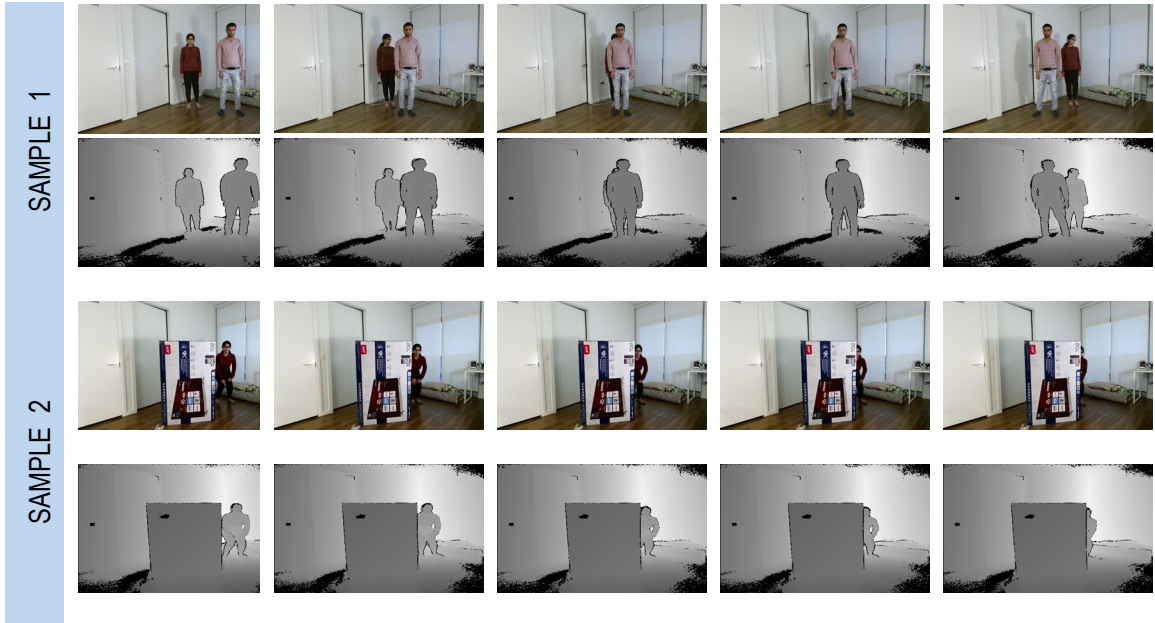

**Figure 4.13.** Sample RGB and Depth image collected for real-time analysis of the proposed long-term tracker

## 4.3.1. Evaluation Metrics

The evaluation metrics used for this analysis are different from the ones mentioned previously in Section 3.3 (which also uses dataset collected by the author). Previously, we wanted to observe the how KCF tracker performs individually as a tracker when compared to the ground truth. However, with the implementation of a new RGB-D tracker, we aim to compare how well does new RGB-D tracker performs as compared to KCF tracker (which only uses RGB).

We take our inspiration for evaluation metrics from the original work of KCF which used average precision as their evaluation metrics. Additionally, we also use confusion matrices for two trackers (our RGB-D tracker and KCF RGB tracker) to discuss the tracker's ability to distinguish when the target is present (True Positives i.e. TP) and when it is absent (True Negatives i.e. TN).

**Average Precision:** The first metric we use for comparison is average precision (AP). Precision can be stated as 'when the model guesses, how well does it guess correctly". Average Precision (AP) as the name suggests is calculated by taking the mean of the precisions (calculated for each dataset). If we define $n$ as the number of scenes in the dataset (e.g. 5 scenes in our dataset) and $s$ as the number of samples(images) in each scene in the dataset (e.g. 150 samples for scene 1, 180 samples for scene 2, etc.) such that total number of test samples is $N = n * s$, then,

$$Precision\ (P) = \frac{TP}{TP + FP}$$

$$Average\ Precision\ (AP) = \frac{1}{N} \sum_{i=1}^{N} P\ , where\ N = number\ of\ samples\ in\ the\ dataset$$

**Confusion Matrices:** Confusion matrix are a performance measure of a classification algorithm. It is a 2D array that compares the true labels against the predicted labels. It shows how well the predictions are made for any category. Confusion matrices can be defined in two ways as mentioned below.

Method 1: We can use a method where TP, FP, FN, TP are dependent on the actual 'number' of classifications w.r.t to the total number of classifications. Hence when $Total\ =\ TP + FN + FP + TN$, elements of confusion matrix can be defined as shown in Table 4.3

**Table 4.2.** **The table shows how elements of the confusion matrix (TP, FP, TN, FN) are computed for Method 1**

| True Positives (TP) | False Positives (FP) | True Negatives (TN) | False Negatives (FN) |
|---|---|---|---|
| The outcome when a model correctly predicts the positive class i.e. models says the target is present when the target is present | The outcome when a model incorrectly predicts the positive class i.e. models says the target is present when the target is absent | The outcome when a model correctly predicts the negative class i.e. models says the target is absent when the target is absent | The outcome when a model incorrectly predicts the negative class i.e. models says the target is absent when the target is present |
| $TP\ (in\ \%) = \dfrac{TP}{Total}$ | $FP\ (in\ \%) = \dfrac{FP}{Total}$ | $TP\ (in\ \%) = \dfrac{TN}{Total}$ | $TP\ (in\ \%) = \dfrac{FN}{Total}$ |

This representation though correct won't help much in our evaluation because for our base tracker KCF (RGB tracker) TN and FN will always be 0. It is because KCF can not identify the absence of the target. Intuitively, if the target does not have the ability to predict the absence of a target, our metrics should show that TN and FN can not be computed and hence '*can not be determined*'.

Method 2: Motivated by the belief that Method 1 may not be adequate, we modify our confusion matrix. We know that at any given instance of time, the target will be either predicted or not predicted. So, we can look at our predictions at every scene to see how well it matches our ground truth. We can now define our confusion matrix as:

**Table 4.3.    The table shows how elements of the confusion matrix (TP, FP, TN, FN) are computed for Method 2**

| For every time the target is predicted in the scene | How many times is the prediction correct? $$TP_{new}(in\ \%) = \frac{TP}{TP + FP}$$ | How many times is the prediction incorrect? $$FP_{new}(in\ \%) = \frac{FP}{TP + FP}$$ |
|---|---|---|
| | | |
| For every time the target is not predicted in the scene | How many times is the prediction incorrect? $$FN_{new}(in\ \%) = \frac{FN}{FN + TN}$$ | How many times is the prediction correct? $$TN_{new}(in\ \%) = \frac{TN}{FN + TN}$$ |

This metrics is also helpful when comparing and discussing datasets where the number of samples in each dataset varies. For example, consider the following scenario:

**Table 4.4.    The table shows the usefulness of Method 2 over Method 1 for two datasets**

| | Total no. of samples | No. of samples where the target was absent | No. of samples when the target was predicted absent | TN (%) (Using Method 1) | TN (%) (Using Method 2) |
|---|---|---|---|---|---|
| Dataset 1 | 100 | 20 | 20 | 20% | 100% |
| Dataset 2 | 200 | 20 | 20 | 10% | 100% |

Method 2 seems to be a fair comparison to answer *'how many false negatives were correctly predicted'* between two datasets with varying samples. Given the above motivation, we use Method 2 as the confusion metrics for evaluation and discussion.

Hence, for the KCF RGB tracker, TN and FN will always be '*undetermined'*. However, if our RGBD tracker performs well, including occlusion scenarios, we expect to have TN and TP values, both high in number.

### 4.3.2. Observations and Discussion

Table 4.5 shows the comparison of average precision (AP) computed on our dataset. Figure 4.14 shows the confusion matrix for the test results for KCF RGB tracker and our RGB-D tracker. As discussed previously, we consider that TP + FP = 100% and FN + TN =100%. We know that KCF RGB tracker gives no TN and FN, and hence, it can't make a decision on it. Figure 4.15 gives a detailed view of the samples in each scene along with the prediction made on them.

**Table 4.5.** **Table shows the comparison of average precision (AP) in % when the two trackers are evaluated on the dataset**

| Tracker | AP |
|---------|-----|
| Ours | 66.35% |
| KCF | 39.99% |

Figure 4.14. shows the confusion matrix for the test results for KCF RGB tracker and our tracker. As discussed previously, we consider that $TP + FP = 100\%$ and $FN + TN = 100\%$. We know that KCF RGB tracker gives no $TN$ and $FN$, and hence, it can't make a decision on it

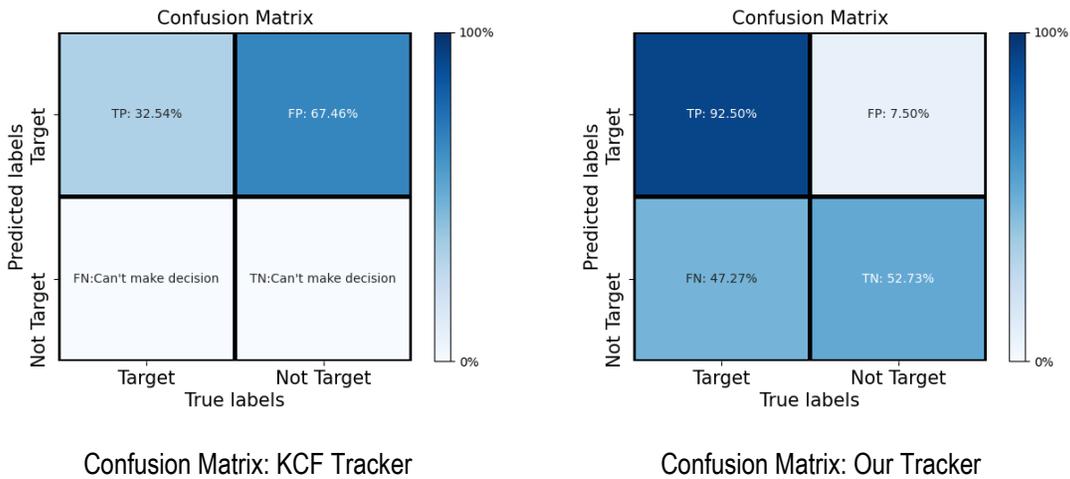

Confusion Matrix: KCF Tracker          Confusion Matrix: Our Tracker

**Figure 4.14.** **Comparison of confusion matrix of test results between our RGB-D tracker and KCF RGB tracker. Note, each row of the confusion matrix adds up to 100% i.e. TP + FP = 100% and FN + TN = 100%. See the related section for an explanation.**

Hence by augmenting image features with the depth information, we make our tracker more robust to occlusions. It is able to detect occlusions and can also detect the absence of the target and stops tracking. This not only increases the $TN$, but also decreases $FP$. From Figure 4.14. we can see that $TN$ are 52.7 % as compared to KCF where we cannot have this value. $TP$ in our tracker is at 92.50% which is significantly higher than KCF at 32.54%. Figure 4.14 shows some sample results from our training method. However, we do have some failure cases which add to our $FP$ and $FN$ as we will now discuss.

Adding depth information to existing features makes the tracker very robust to occlusions. However, when the target is absent, the tracker attempt to locate the target in the vicinity. Since the tracker keeps looking in the neighboring space, there are locations where it falsely predicts the target momentarily adding to False Positives. Also, the proposed tracker cannot detect partial appearance. Hence, in scenes where the target is coming out of occlusions, there are no clear detections adding to False Negatives. When the target is clearly out of the occlusion, the target is tracked again. Figure 4.16 shows some of the failure cases of the proposed tracker.

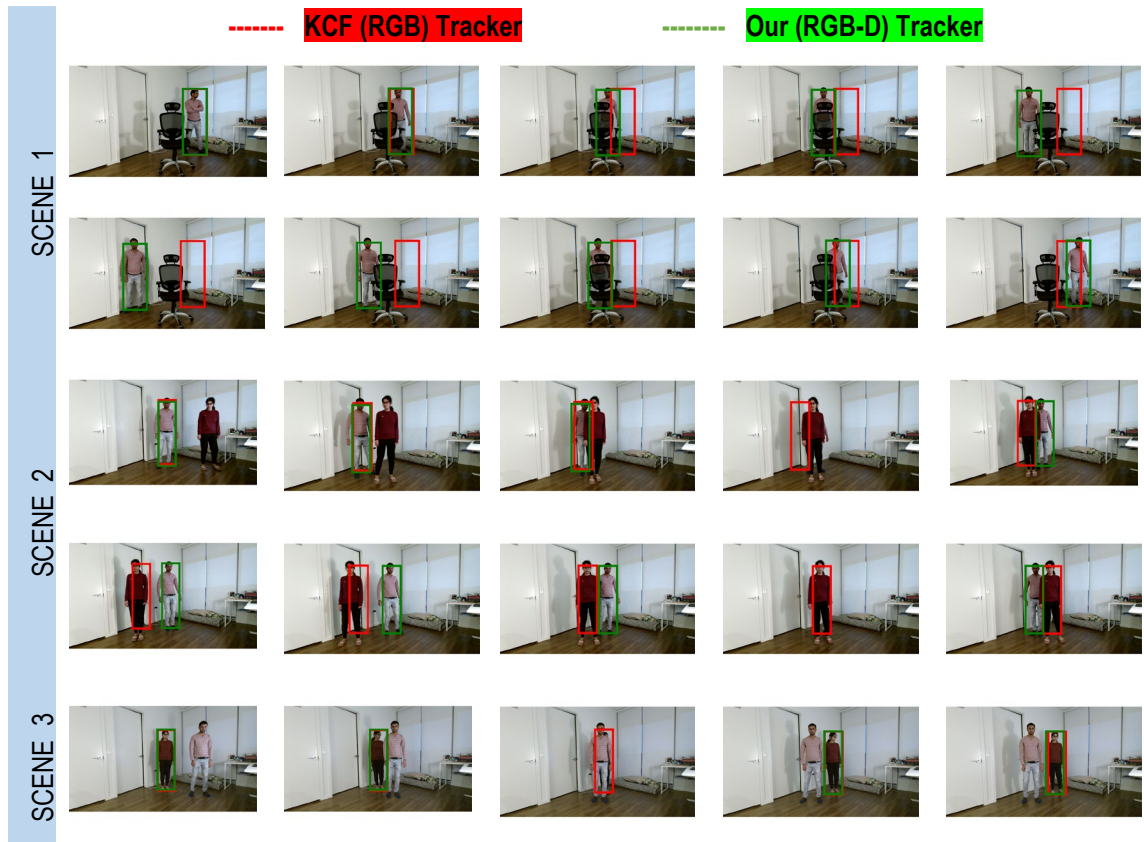

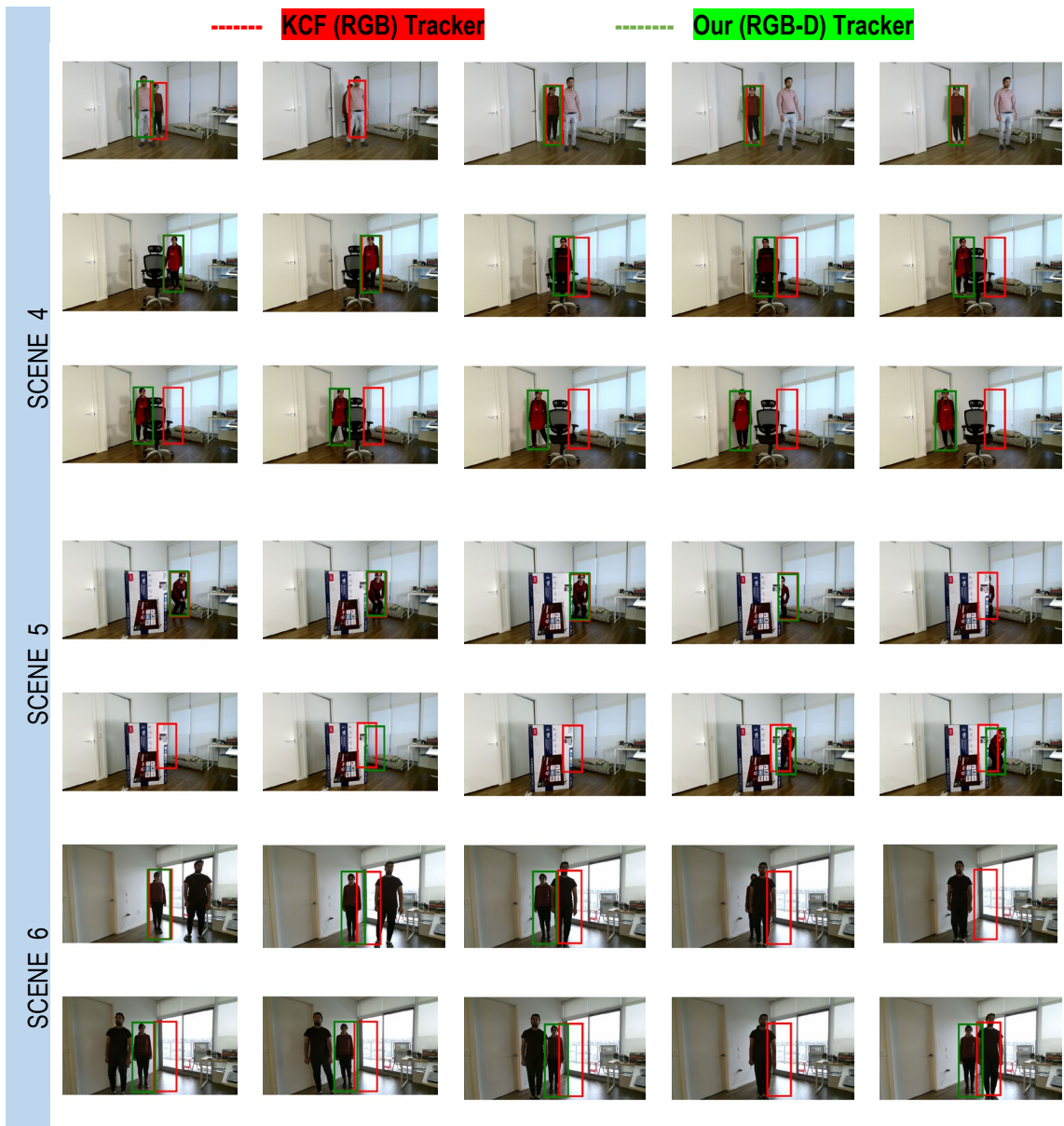

**Figure 4.15.** Qualitative results for the proposed RGB-D tracker, compared to Kernelized Correlation Filter (KCF) tracker. Images show the tracking bounding box on test data. The red color denotes the KCF tracker and the green color denoted our tracker.

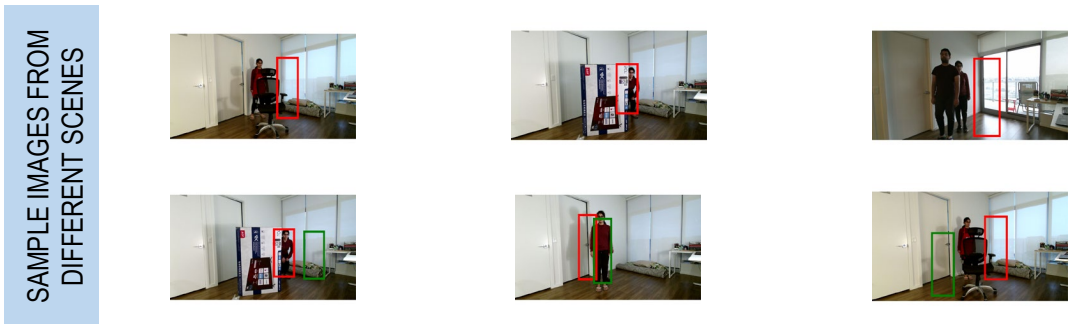

**SAMPLE IMAGES FROM DIFFERENT SCENES**

**Figure 4.16.   Failure cases of our RGB-D tracker**

## 4.4.  Discussion

In our proposed implementation, it is shown that depth cues when used with image features in KCF tracking can significantly improve the performance of the tracker. We observe that KCF tracker loses the target information once it is occluded adding to its inability to track a subject for a long time. Depth information, however, provided an additional layer of information that informs the tracker of the status of occlusion (true or false). Hence, it can stop tracking the target when the target is hidden. Using similar information, when the tracker is informed that the target is no more hidden, using the sliding window approach, it is able to search exhaustively in a large search space around the target to relocate the target. We are assuming the tracking happens in an indoor environment; hence, the nominal walking speed is considered. The experiments prove the usefulness of depth information as additional information in making the tracker robust. Experiments also show few failure cases showing that at a certain time, the model may drift slightly from the expected path, possibly in cases a.) when the target is partly hidden (just coming of occlusion) b.) failed to correctly estimate the depth information. From our experimental observations, we see that the tracking performance of the tracker is improved compared to our KCF RGB tracker but also has the potential to improve even further.

# Chapter 5.

# Study and Experimental Observations in Using Particle Filter

In Chapter 4 we propose a tracker that was an improvement over the existing RGB based kernel correlation tracker. It used depth information as well as existing RGB based features (hence called RGB-D tracker) to address the issue of occlusion. In the previously proposed approach, we used a sliding window in the search space defined around the last known position of the target to locate the target, when it re-appears. In Chapter 4, Section 4.2 we validated this approach and concluded that it does help in making the tracker robust to occlusions. However, the sliding window approach is a slow process since it has to extract all possible patches in a large search space and correlate (find similarity) it with original target features, to find the location of the target in a new frame. Searching most of this search space is redundant since the target can not be present at many locations (e.g. in a common stationary sensor location), if the target is moving from right to left, searching in the space above the target head, or searching on way extreme ends of target search space will not yield any positive results most of the time). If we know the current state of the target (i.e. position, velocity), we can anticipate that the target can possibly move only in certain directions. If we can use such a model-based approach which, for a current state of the tracked object can help to identify possible locations where the target will be present, it can allow the tracking algorithm to search much narrower and accurate space. Motivated by this idea, we explore the Bayesian framework like particle filter. We hypothesize that we can use RGB based KCF tracker to track the target, depth to identify occlusions, and particle filter to better localize the target, we can improve predictions. We study this hypothesis and see if it can help the current RGB-D tracker in improving its tracking performance.

## 5.1.  Particle Filter Framework

Particle filter framework is helpful in continuous (target's position and velocity can smoothly vary over time) and non-linear behaviors (two target's colliding with each other). It is based on the Bayesian formulation where the particles are propagated overtime to maintain multiple hypotheses of the possible location of the target. It then uses a model to predict the next state (e.g. location) at the next available instance (e.g. time). It is based on Monte Carlo methodology which provides a probabilistic framework for tracking target over time by sampling *posterior* density using *prior* information of the target. Each such sample of the state vector is referred to as a particle. The basic idea of resampling is to eliminate particles that have small weights and concentrate on the ones which have a large weight. This sampling is based on different techniques, most commonly being mean and variance of a location (so that samples are not always distributed at random locations). However, if the number of particles propagated is too high, these dense samplings if used throughout the tracking process, add to the computational complexity of the tracker which increases linearly with time. Optimizing *how* to propagate the particles and simultaneously use a *smaller number of particles* is an entire research domain in itself which we will leave for future research.

Particle filter algorithm for state estimation can be presented as:

1. Initialization (k=0): Set the initial state vector $s = \{x_0^i | i = 1,2, \ldots N\}$ and weights $w_0^i = \frac{1}{N_s}, i = 1,2, \ldots N_s$ such that all particles are equally probable at the beginning of the algorithm.

2. Measurement Update: Update the weights $w_k^i = w_{k-1}^i . p(y_k | x_k^i)$ with normal probability density function and normalize the weight $w_k^i = \frac{w_k^i}{\sum w_k^i}$

3. Resample: Obtain a new set of particles and weights after the resampling method

4. Compute the new state estimate $x_{est_k} = \sum_{j=1}^{N_s} w_k^j . x_k^j$

5. Set $k = k + 1$ and iterate the algorithm to step 2

In past, various works have discussed the use of particle filters in visual tracking using RGB data and RGB with Depth data. Particles filter has been most popularly used by RGB based trackers. [115] uses adaptive fusion model to integrate multi-cue extracted for each evaluated particle to enhance tracker performance. [116] uses color-based likelihood (RGB data) in particle filter to track unmanned aerial vehicles. Many recent works[117] [39] have shown that particle filters have been used successfully to handle scale variations and partial occlusions using both RGB and depth data. Readers are referred to [118] for recent advances on particle filters and their challenges. Motivated by recent success with particle filter in improving tracking accuracy, we explore if we can use particle filter to better localize the target (instead of using sliding window approach) and benefit our occlusion aware re-detection tracker which uses image features for correlation tracking and depth information for identifying occlusions..

## 5.2. Particle Filter with RGB-D Kernel Correlation Filter Tracker

For our current methodology, we propose to augment our tracker such that RGB and depth data (in KCF tracker) can help us identify occlusions and particle filter framework can help better localize the area of the target location (when hidden) to re-detect the target when it comes out of occlusion. This can make the target more accurate during the re-detection stage. Hence, our tracker will not suffer from the large computational complexity of using a sliding window at each frame as mentioned in Section 4.1.4. Motivated by this idea, this chapter studies the use of particle filter framework with depth-based tracking and notes the observations.

This chapter explores the potential of using particle filter as an additional layer of tracking for finding the optimal target location by propagating the samples around the target. This is especially needed when the correlation between two subsequent frames is low (e.g. occlusion) and the model needs additional information (e.g. more target samples) to decide the future (when the target reappears).

This chapter is organized as follows: Section 5.3 discusses the tracking pipeline. In this section, we discuss the parameters which were taken into consideration as a part of the analysis, sample propagation, and target estimation. It also discusses how particle filter is used to localize the target along with the existing KCF tracker. It is followed by Section 5.4 where we show experimental observations on a selected dataset for occlusion scenarios which the authors collect. This section also shows the success cases and failure cases to better understand the use case of particle filter Finally, in Section 5.5. we discuss how particle filters can be used to improve the tracker and its existing shortcomings.

## 5.3.  Tracking Pipeline

### 5.3.1. Implementing Particle Filter

The first target is detected using the ground truth available to us. We then define the state (configurations, particle) of the target as follows:

$$s = \{x, y, \dot{x}, \dot{y}, H_x, H_y, \alpha\}$$

where $(x, y)$ is the top-left coordinates of the bounding box, $\dot{x}, \dot{y}$ are the velocity of the target, $(H_x, H_y)$ is the width and the height of the target bounding box and $\alpha$ is the scale parameter. With these configurations set, we can now define the prediction model as follows:

$$\begin{cases} x_{t+1} = x_t + \dot{x} + (0, \textstyle\sum_R) \\ y_{t+1} = y_t + \dot{y} + N(0, \textstyle\sum_R) \\ \dot{x}_{t+1} = \dot{x} + N(0, \textstyle\sum_R) \\ \dot{y}_{t+1} = \dot{y} + N(0, \textstyle\sum_R) \\ H_{x,t+1} = (1 + \alpha_t) . H_{x,t} + N(0, \textstyle\sum_R) \\ H_{y,t+1} = (1 + \alpha_t) . H_{y,t} + N(0, \textstyle\sum_R) \\ \alpha_{t+1} = \alpha_t + N(0, \textstyle\sum_R) \end{cases}$$

where $N$ is the normal distribution and $\sum_R$ represents the covariance of the process noise. The prediction model assumes that the target moves with constant velocity ($\dot{x}, \dot{y}$). This assumption is made since we don't expect our targets to increase their speed with time. Hence, for each time step, we can get a new location $x_{t+1}, y_{t+1}$ using this constant velocity. Velocity at the next time step is computing by adding a value from a normal distribution within the range of process noise. Parameter scale $\alpha$ simply changes the size of the bounding box. Process noise [119] represents the idea/feature that the state of the system changes over time, but we do not know the exact details of when/how those changes occur, and thus we need to model them as a random process. Hence, it is a concept to ensure that prediction made based on the model is having some errors since no prediction is perfect. This is done to accommodate nearby particles from the distribution for better representation of the samples.

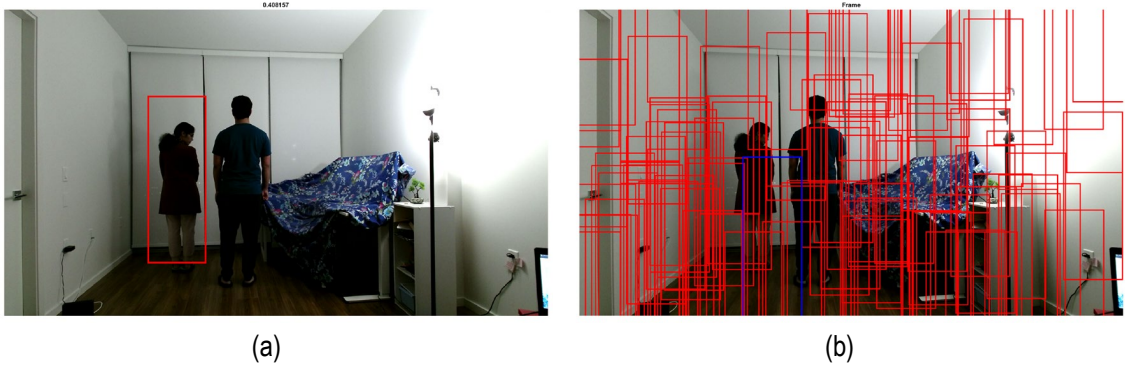

<center>(a)                          (b)</center>

**Figure 5.1.**    **Figure shows (a) the original image with the target, here, the girl, and (b) N random particle propagated at the initial stage**

Once we have defined the prediction model, we can now randomly initialize N samples in the image space as shown in Figure 5.1. with equal weights. Each of these particles is then propagated using the prediction model defined above i.e. each particle state is defined by $1x7$ vector $\{x, y, \dot{x}, \dot{y}, H_x, H_y, \alpha\}$. Now, we calculate the likelihood of each particle being encompassing the target and assign weights to them. This is done using a color histogram (Appendix D). For each sample region, we assign each pixel value into their corresponding color bins in different RGB channels.

We also weigh these pixels depending on the distance from the center i.e. farther the pixel, lower the weight, and vice-versa. The weighting function can be defined as follows:

$$k(r) = \begin{cases} 1 - r^2, & r < 1 \\ 0, & otherwise \end{cases}$$

where $r$ is the normalized distance between pixel $p$ and region center $c$. These weighted pixels (computed from $k(r)$) form our color distribution in each state. We can now calculate the similarity between two states using Bhattacharya Distance [120]. For two pixels $p, q$, Bhattacharya Distance can be defined as:

$$f(p, q) = \sqrt{p.q}$$

For two similar pixels, $d(p, q)$ will be very low and vice versa. Hence, we will define similarity measure as:

$$d = \sqrt{(1 - f(p, q))}$$

Hence, if $d$ is large, two pixels are very similar, and if $d$ is low, they are very dissimilar. Once we have the similarity metrics, we can calculate the *likelihood* of each particle being the target by probability density function (PDF) of the normal distribution:

$$w_i = \frac{1}{\sqrt{2\pi\sigma^2}} \cdot \exp(-\frac{d_i^2}{2\sigma^2})$$

where $i$ is the particle such that $i = 1, 2, \dots N$ and $w_i$ is the weight. Weights are nothing but the likelihood of each particle. After normalization, each particle will be assigned a weight $w_i$ according to its distance $d_i$.

After particle propagation and weight update, we can calculate the *posterior* (subsequent) state $s_E$ using a weighted mean of all targets:

$$s_E = \sum_{i=1}^{N} w_i \cdot s_i$$

 This new state is used as the predicted target location as shown in Figure 5.2 and is used by the KCF tracker to interpolate the model and make detection at the next frame.

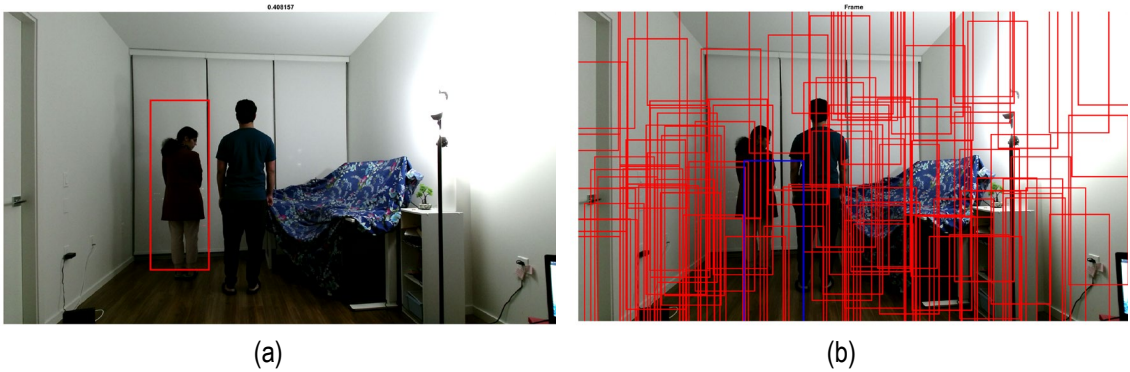

(a)                                                    (b)

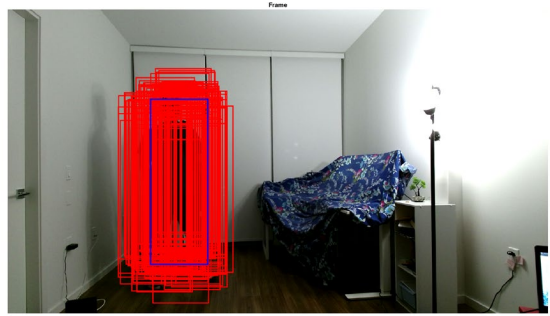

(c)

**Figure 5.2.** **The figure shows target localization by particle filter. (a) shows the original image where the target is the girl (N=1) b.) shows N samples propagated to locate the target (N=3) c.) shows the target localization after some time by particle filter (target estimate is shown with a blue box in the center) (N = 7) where N = frame number {1,2,…50} and defines the frame number.**

## 5.3.2. Training, Detection, and Re-detection of Target.

The tracking pipeline is based on the idea that given the knowledge of the depth information; we will use particle filter to better localize the target (search in image space where the target is expected to be present). Hence, the particle filter framework can help in target estimations during tracking when the target is visible, however, once occlusion is detected, we will use the estimate of the location provided by particle filter framework to find the best match between the target patch at frame $i$ and the original target patch. If the correlation is high, we expect that the particle filter is correctly tracking the target. However, if the correlation is low (e.g. coming out of occlusion scenarios), depending on a certain threshold we can either re-initialize the particles at the last known location to find the target or retain its previous position. Once the location is predicted by this framework, kernel correlation filter tracker continues to interpolate the model (using Equation 26) and make final predictions and detect the target in subsequent frames.

## Training Pipeline

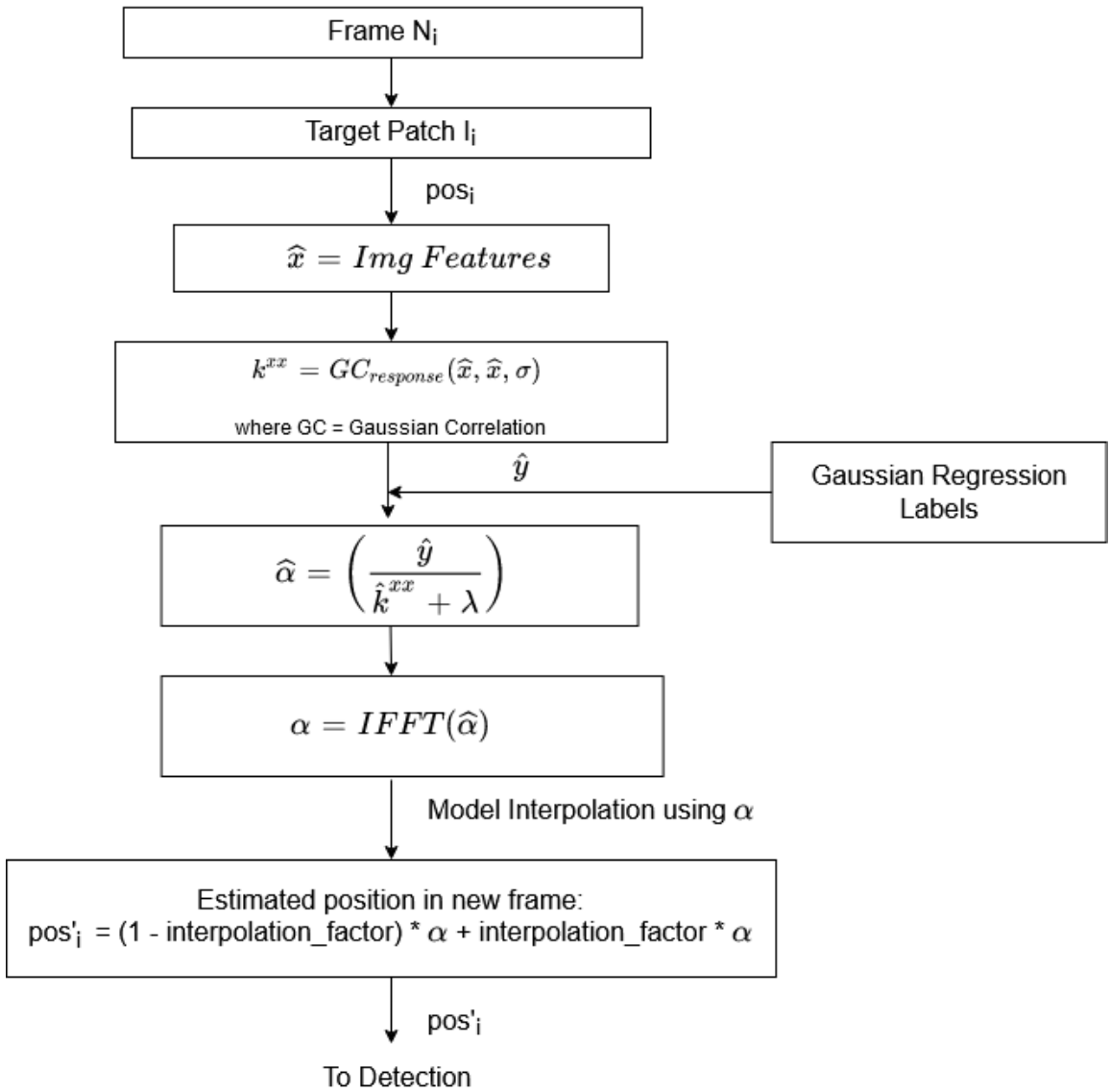

**(a) Training pipeline of the PF based tracker**

**Detection Pipeline**

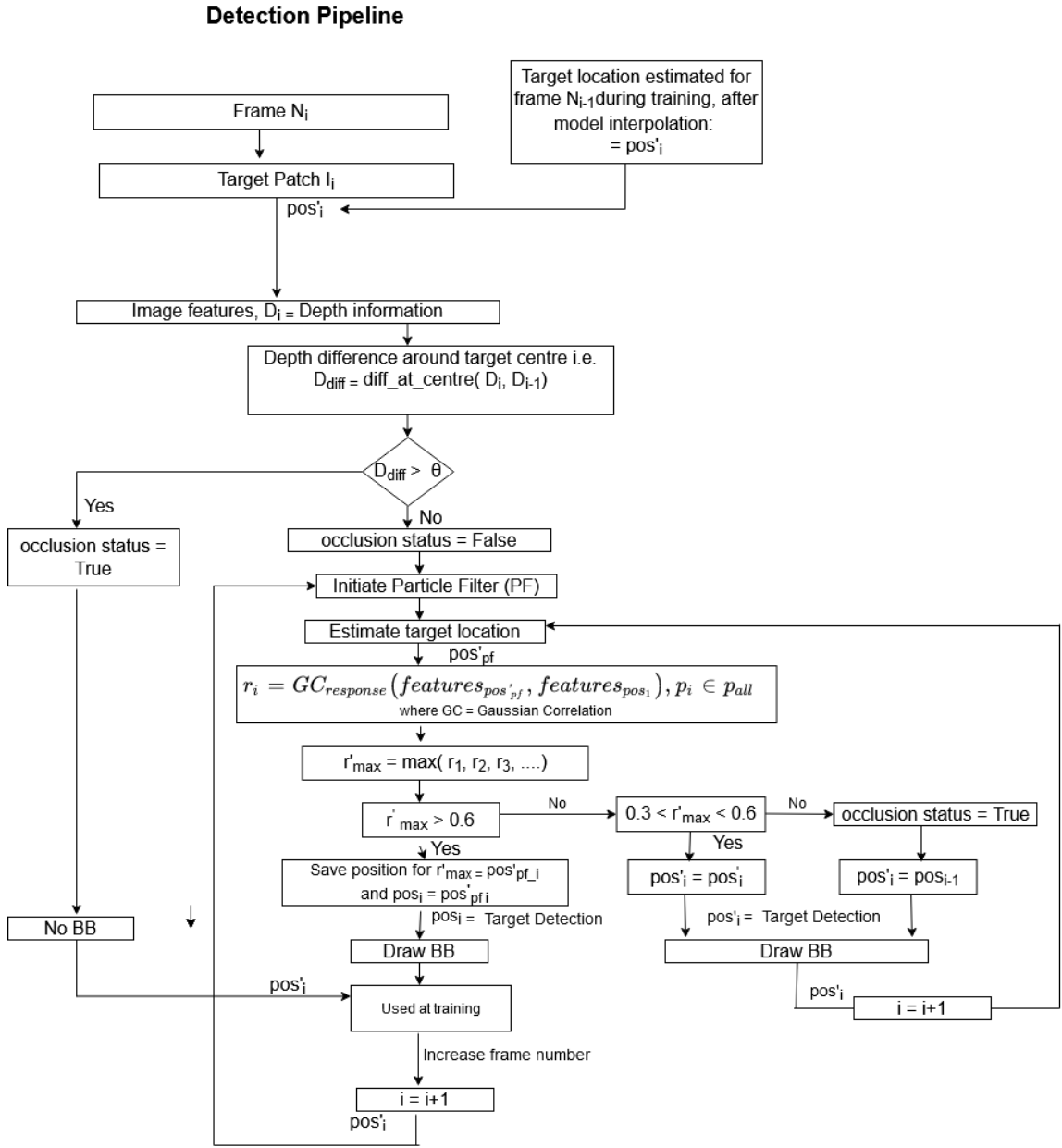

**(b) Detection pipeline of the PF based tracker**

**Figure 5.3.    Flowchart of the training and detection pipeline of the RGBD KCF tracker with particle filter framework**

The tracking pipeline for the proposed tracker is shown in Figure 5.3 and is also detailed in Algorithm 2 in the table below:

| kernel_correlation( $x$ , $z$, $\sigma$) | |
|---|---|
| 1 | Calculate $\hat{k}^{xz} = FFT\ (\exp(-\frac{1}{\sigma^2}\ (\ \|x\|^2 + \|z\|^2 - 2F^{-1}\ (\hat{x}^* \odot \hat{z}\ )))$ |
| 2 | Return: k = IFFT $(\hat{k}^{xz})$ |

(a)

| Training (x, y, $\lambda$) | |
|---|---|
| 1 | kernel_correlation(( $x$ , z, $\sigma$) |
| 2 | Compute $\hat{\alpha} = \left(\frac{\hat{y}}{\hat{k}^{xz}+\lambda}\right)$ |
| 3 | Return $\alpha = IFFT(\hat{\alpha})$ |

(b)

| Detection $(x, z, \lambda)$ | |
|---|---|
| 1 | $\hat{k}^{xz}$ = kernel_correlation(( $x$ , z, $\sigma$) |
| 2 | Calculate $\hat{f}(z) = \hat{k}^{xz} \odot \hat{\alpha}$ |
| 3 | Response = $f(z) = IFFT\ (\hat{f}(z))$ |
| 4 | Return max $(f(z))$, position of max $(f(z))$ |

(c)

| Algorithm 2: RGB-D kernel correlation tracker with particle filter (PF) framework | | |
|---|---|---|
| 1 | | Extract patch $patch_{rgb}$ from image $I_i$ |
| 2 | | Calculate $D_i$ = Depth of image $I_i$ |
| 3 | | Get initial target position $pos_i$ (i = 1) from ground truth |
| 4 | | *Training* ( $features_x$ , features$_x$, $\sigma$) where $features_x$ is the hog features of the (RGB) target at $pos_i$. For frame 1, $pos_i$ is provided by ground truth, for subsequent images, is it is output of detection pipeline image $I_i$. Interpolate the model with training output. Estimate target position $pos'_i$ for next frame. $i = i + 1$ |
| 5 | <u>while</u> <u>$i < N$</u> | |
| 6 | | In image $I_i$.from $pos'_i$ extract image patch $patch_{pos'_i}$ and calculate, $D_i$ |
| 7 | | Calculate $D_{diff} = diff\_at\_centre\ (D_i, D_{i-1})$ where $diff\_at\_centre$ computes difference of depths at target's centre area. |
| 8 | If | $D_{diff} < \theta$ for certain threshold $\theta$ ; set $occ_{status} = False$. (no occlusion) and go to Step 10 |
| 9 | else | go to Step 19 |
| 10 | | Initiate particle filter and propagate $M$ samples around $pos'_i$ |
| 11 | | Estimate the target location using particle filter. Call it $pos'_{pf}$. |
| 12 | | Detect new target location using Detection (features$_{pos'_{pf}}$, features$_{pos_1}$). This computation is done in Fourier domain. Note maximum correlation response $r'_{max_{i+1}}$. |

| Algorithm 2: RGB-D kernel correlation tracker with particle filter (PF) framework | | |
|---|---|---|
| 13 | If | $r'_{max} > 0.6$; set $pos_i = pos'_{pf}$ Draw bounding box at $pos_i$ and increase frame number i.e. $i = i + 1$. Go to Step 11. |
| 14 | else | go to Step 15. |
| 15 | If | $0.3 < r'_{max} < 0.6$, set $pos_i = pos'_i$ (patch has some feature similarity but most likely dislocated so retain this frame's estimated position) and go to Step 11 * |
| 16 | else | if $r'_{max_{i+1}} < 0.3$ (most likely coming out of occlusion); set $pos_i = pos_{i-1}$ go to Step 11 |
| 17 | | Draw bounding box at $pos_i$ |
| 18 | | Go to Step 6 |
| 19 | | If $Depth_{diff} > \theta$, set $occ_{status} = True$ , save $pos_i$ . Stop tracking i.e. don't draw bounding box. |
| 21 | | Go to next frame $i + 1$, and go to Step 10 |
| 22 | | End |
| * Note: Since the target has already been localized and has knowledge of the target previous location and velocity, it will better propagate the particles and not randomly initialize particles at random places in the whole image | | |

**Figure 5.4.** **Figure showing the PF based RGB-D tracking algorithm. (d) shows the full algorithm which uses modules shown in (a) – kernel correlation computation, (b) training using kernel correlation module, and (c) detection using kernel correlation module**

# 5.4. Experimental Observations

## 5.4.1. System Setup

At the beginning of the chapter, we hypothesized that we can use RGB based KCF tracker to track the target, depth to identify occlusions, and particle filter to better localize the target to improve predictions To test our hypothesis, we study the use of particle filter on RGBD data. One of the main focus of adding particle filter is to assist the RGB and Depth information in recovering the tracking target in cases of occlusion, especially when the target is re-appearing after being occluded.

Since we are proposing to study an extension to our proposed tracking framework which is expected to perform better in occlusion scenarios, we collect data that focus particularly on this challenge but in less cluttered scenarios. We ensure that our dataset accommodates different scenarios of occlusion e.g. moving target occluded by a moving object, stable (non-moving) target occluded by a moving object. We also ensure that our targets (being occluded) have various ranges of speed, slow speed, and fast motion (which can cause motion blur). The data collected assumes that the target is moving unidirectionally or bidirectionally in the horizontal plane. We do not consider scenarios where the target may tend to move towards and away from the sensor. They are useful in scenarios where the sensors are mounted in hallways (at a height at an angle) or mounted on stationary mobile robots which for majority part observe target moving from left to right or right to left. Hence, our data is a fair representation of various scenarios which are likely to occur in day to day lives, albeit in ideal environment (less clutter, constant speed).

**Data collection & Hardware Setup**: Similar to our previous experiments, the data is collected using Microsoft Kinect V2 in an indoor home environment. Due to COVID-19, we were restricted in the amount of space we perform the experiments and the number of subjects that we can consider. Figure 5.5 shows our setup used to collect the data.

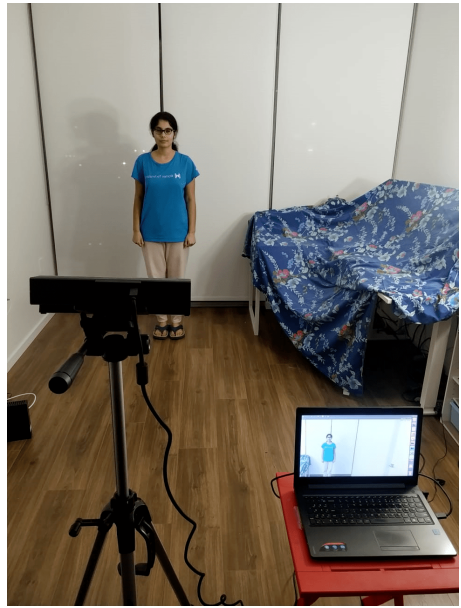

**Figure 5.5.**     **The figure shows the setup using Microsoft Kinect V2 which we used to collect the data. We removed objects that can cause background clutter and kept the environment as simple as possible.**

Figure 5.6 shows few samples of RGB images and their corresponding depth images from our dataset.

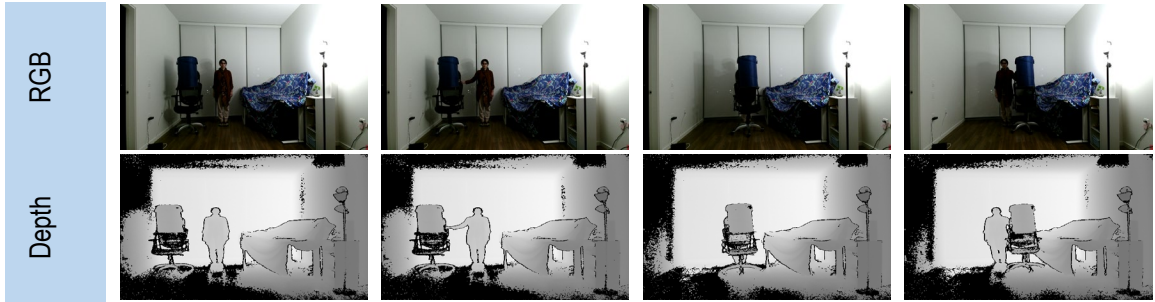

**Figure 5.6.** Sample RGB and Depth images from our dataset.

**Data Annotation**: All the data were manually annotated by the author to collect the ground truth. Each ground truth bounding box depicts the most tightly fitting box that can be drawn to have the target within the box.

## 5.4.2. Observations

The particle filter framework showed some positive results when the target is re-detected during the re-detection stage (target is coming out of the occlusion). In such situations, the target is usually partly hidden, and the features are not sufficient for the depth-based RGB tracker to track. Particle filter, using its prior knowledge of target velocity attempts to locate the target in the frame as can be seen in Figure 5.7. The scenes in our dataset shown in Figure 5.7 has single human target being occluded by other object/human in the scenes. The tracker aims to correctly re-identify the tracker when it is just coming out of occlusion, something the RGB-D KCF tracker was not able to achieve.

**Table 5.1.** Comparison of confusion matrix of occlusion dataset results between (a) KCF RGB tracker and (b) PF tracker with RGB-D. Note that. TP + FP = 100% and FN + TN = 100%

|  | TP (%) | FP (%) | FN (%) | TN (%) |
|---|---|---|---|---|
| KCF RGB Tracker | 39.39 | 60.61 | Tracker can't make a decision | Tracker can't make a decision |
| Our RGB-D-PF Tracker | 74.07 | 25.93 | 41.67 | 58.33 |

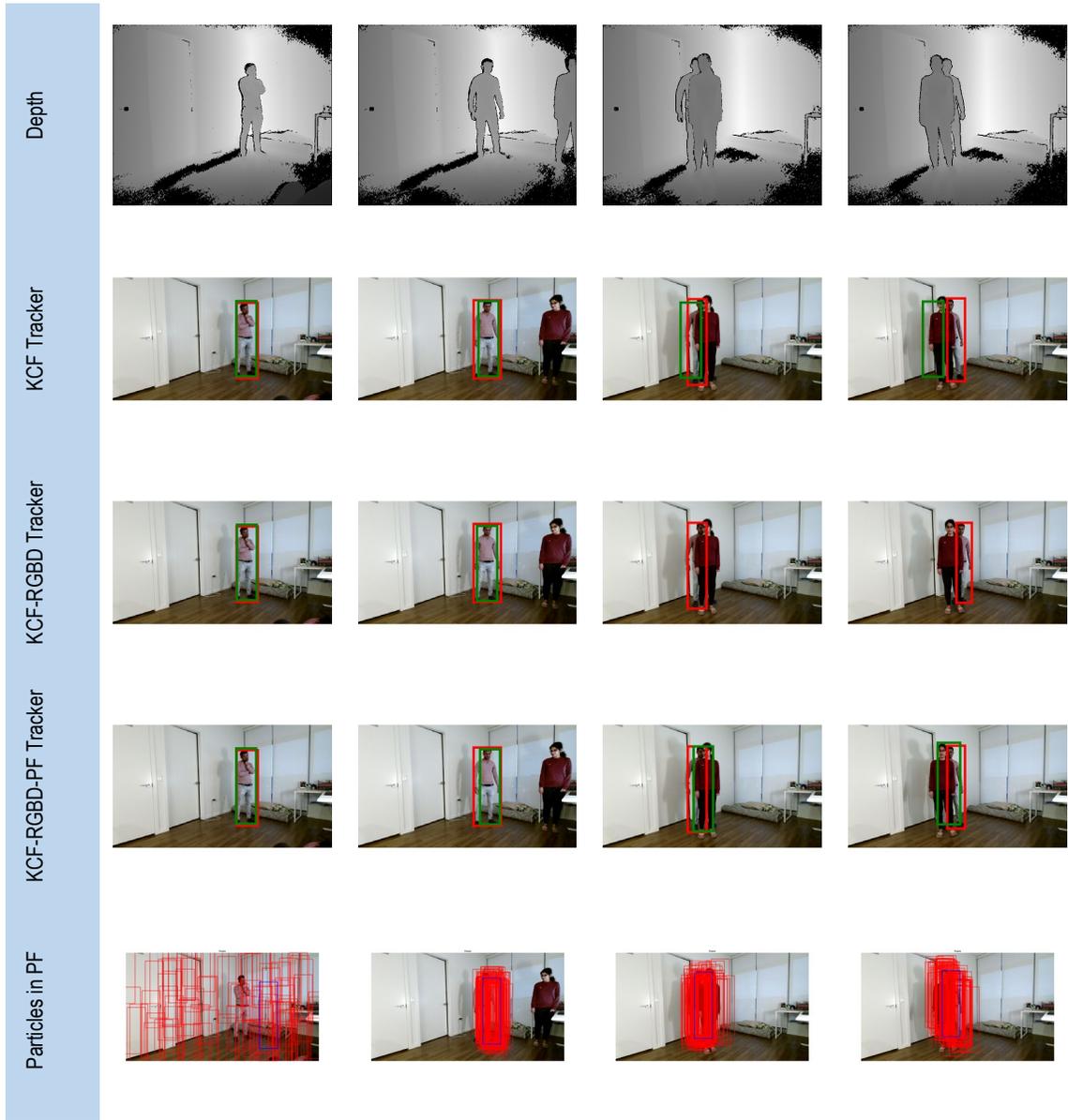

(a) Dataset 1

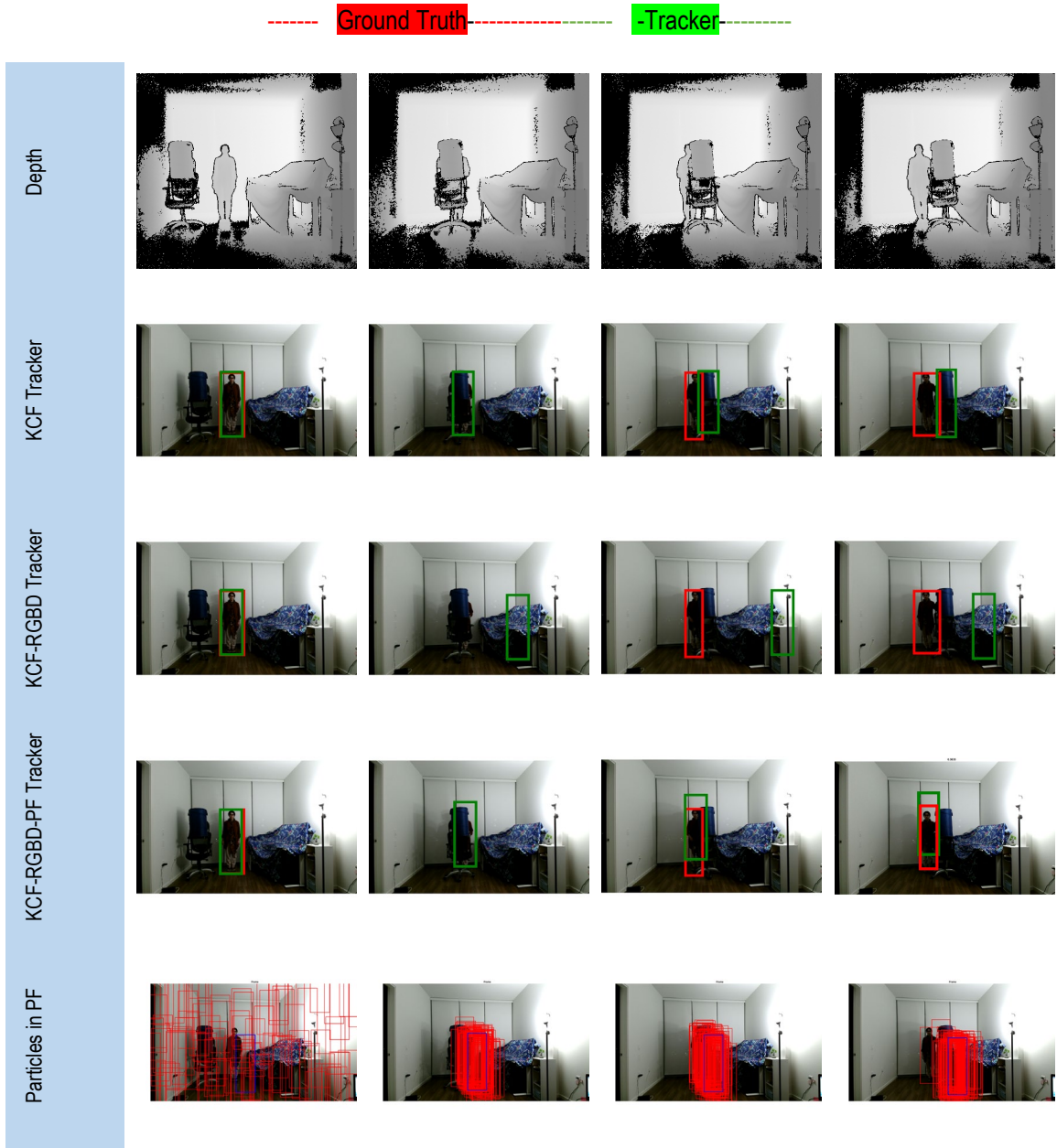

(b) Dataset-2 (This example shows that if particle filter don't give better patches, it continues to track using correlation with previously known location)

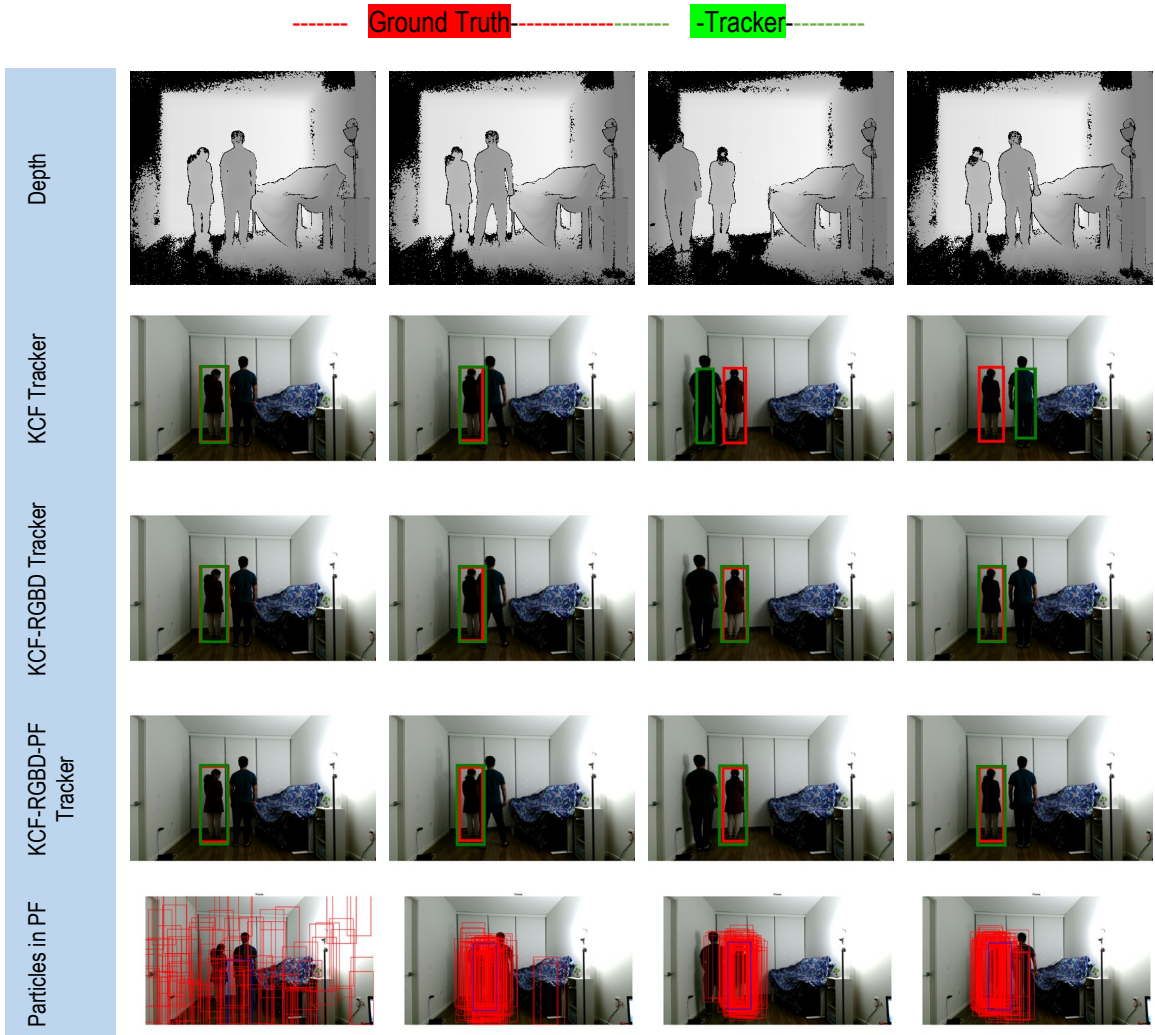

(c) Dataset-3

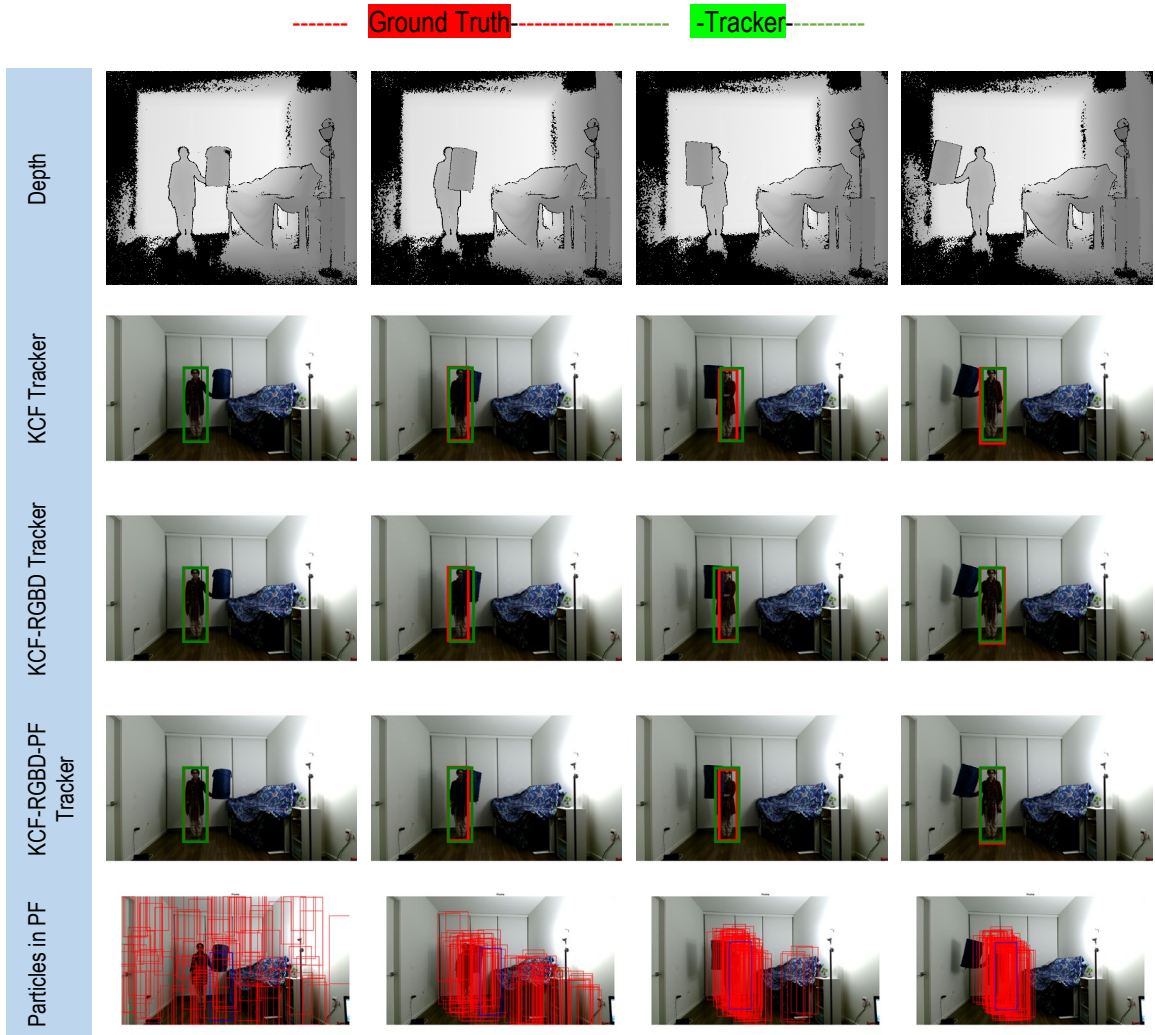

(d) Dataset-4

**Figure 5.7.** **The figure shows tracking performance on occlusion dataset in different challenging scenarios. This dataset does not include when the target is moving it at a very high speed causing motion blur.**

**Failure Case:** One of the most challenging scenarios for any tracker is its speed. If the target is moving too fast, not only its features vary (because of motion blur as shown in Figure 5.8) but also the depth of the target has a higher chance of varying quickly.

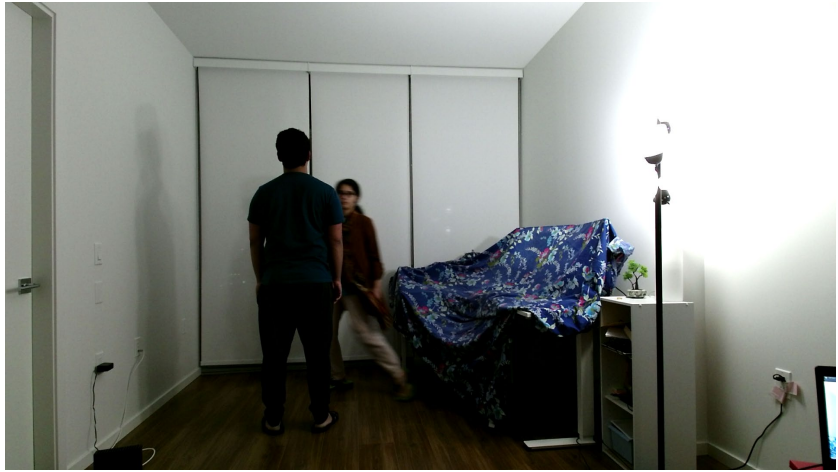

**Figure 5.8.** The figure shows the case of motion blur when the target is moving at a high speed.

We tested our algorithm on a fast-moving target and observed that True Positives (TP) and True Negatives (TN) dropped significantly. Figure 5.9. shows that the particle filter does make an attempt to locate the target and with the knowledge of the target velocity moves the particles from left to right (the direction of the target). However, after some time, with rapidly changing velocity and direction (target moving back from right to left), the particles are unable to localize the target (since the KCF is dependent on correlating features of two subsequent frames which here doesn't show many similarities). Hence, the tracker loses the target and gives a higher number of False Positives (60%) and False Negatives (71%) (yet lower than other trackers) as shown in Table 5.2

**Table 5.2.** The table shows True Positives, False Positives, False Negatives, and True Negatives (all in %) when the target is moving very fast.

|  | TP (%) | FP (%) | FN (%) | TN (%) |
|---|---|---|---|---|
| KCF RGB Tracker | 26.32 | 73.68 | Tracker can't make a decision | Tracker can't make a decision |
| Our RGBD Tracker | 10.53 | 89.47 | Tracker can't make a decision | Tracker can't make a decision |
| Our RGBD-PF Tracker | 40.00 | 60.00 | 71.43 | 28.57 |

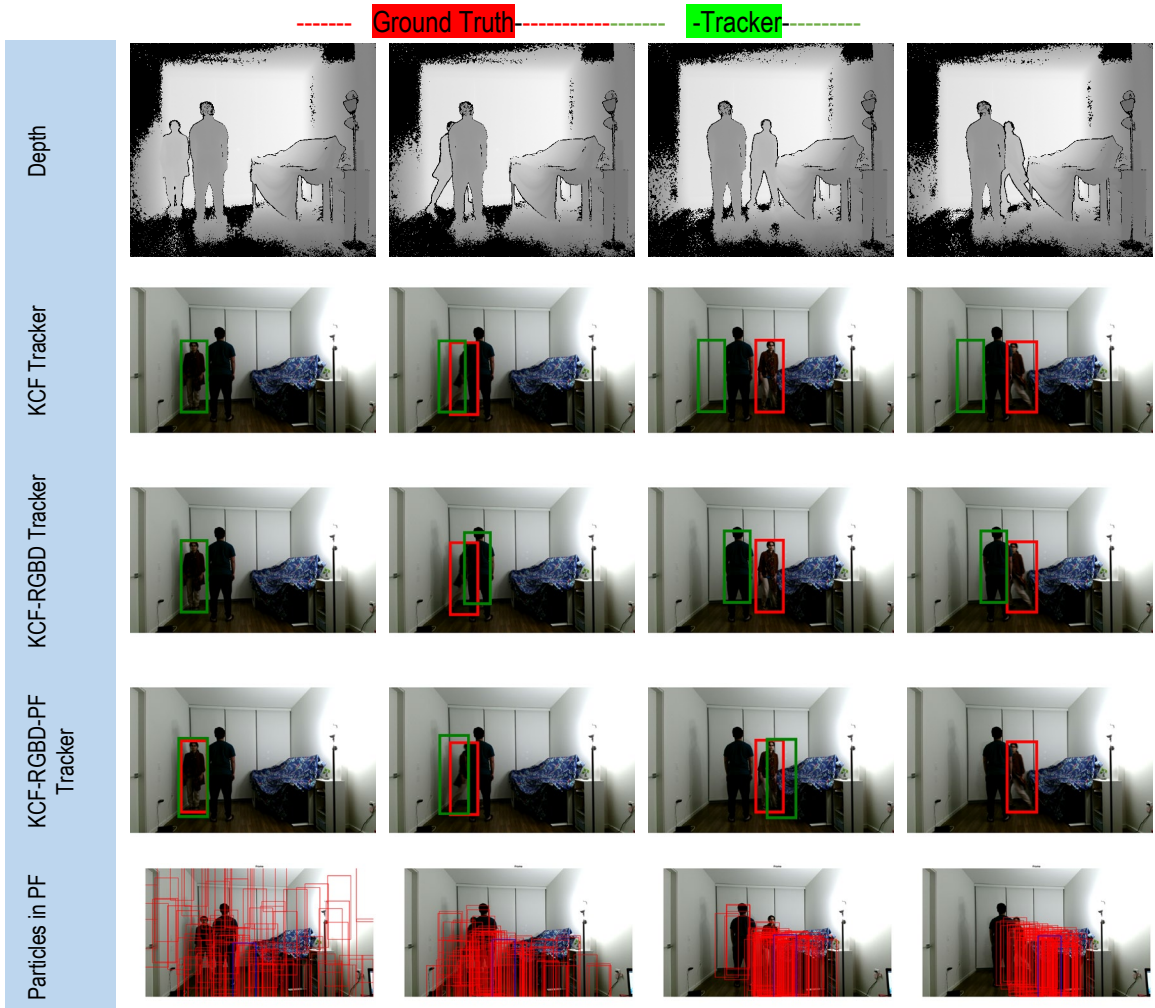

**Figure 5.9.** **Figure the failure case of particle filter based RGBD tracker. It shows performance on occlusion dataset in the challenging scenario of high speed causing motion blur. Frame number (N) is N=1, N=2, N=6, and N=9**

## 5.5. Discussion

In our study for using particle filter to augment the performance of RGB-D KCF tracker, we observe that it helps localize (finding the location) the target. However, our failure case shows that particle need some sample frames before the occlusion in order to better establish the location of the target. We are assuming the tracking happens in an indoor environment; hence, the nominal walking speed is considered. For example, if the tracker is initiated at frame N=1 and the target is hidden at N=3, (where N is the frame number), it was observed that there is not sufficient time for particle filter to distribute the particles around the target. If the target gets sufficient time (e.g. target is visible and is moving slowly before it is occluded) to be localized by the particle filter, it uses our algorithm (discussed in Section 5.1.2) to track the target. From our experimental observations, we see that the tracking performance of the tracker is improved, though marginally, as compared to our KCF RGB-D tracker.

# Chapter 6.

# Conclusions and Future

## 6.1. Conclusion

This thesis explores the working of kernel correlation filter tracking algorithm in great detail (theoretically, mathematically, and visually) and proposes a depth-based kernel correlation filter that is robust to occlusions and can re-detect the targets. The thesis first explains the working and implementation of kernel correlation filter in detail. The code for the demonstrations mentioned in Chapter 2 is publicly available at [https://github.com/copperwiring/KCF_tutorial_code](https://github.com/copperwiring/KCF_tutorial_code) for the user to understand the underlying intuitions, strengths, and weakness of the tracker. The detailed study and understanding of the KCF tracker uncovered opportunities to build a more robust RGB-D KCF tracker that is occlusion aware and hence can track a target in long term. However, due to the computational complexity of this tracker which searches exhaustively using a sliding window to locate the target, a study was done to identify other alternatives for target estimation in cases of occlusion. One such approach is a Bayesian approach like particle filter which we studied to observe its efficiency over RGB-D KCF tracker. Some of the key results of the analysis of kernel correlation tracker, RGB and depth based KCF tracking, tracking with particle filter framework, and experimental studies are:

- Detailed experimental observations done on standard benchmarks show that kernel correlation filter (KCF) tracker is a very efficient tracker which can learn the features on the fly, making it suitable for online learning (without requiring learning features on large dataset offline like in deep learning methods). However, it has significant weaknesses like the inability to adapt to scale variations, losing the target information in cases of occlusion, etc.

- The existing tracker benefits from the use of depth data as additional information to make the tracker more robust by addressing the issue of occlusion. The proposed RGB-D tracker is an occlusion aware RGB and depth based KCF tracker which tracks the target using kernel correlation and uses the depth information to identify instances of occlusion. Once occluded, the tracker stops tracking the target and attempts to re-locate the target using the sliding window approach. Once the target re-appears from the occluding object, the tracker searches (in specific search space) for the target location by identifying a target patch that best correlates to the original target features. We tested this approach on a benchmark like Princeton Dataset as well using data collected by the author (using Microsoft Kinect V2 sensors). Our results show that the proposed RGBD tracker shows promising results and is robust to occlusion. However, the sliding window approach tries to localize the target by exhaustive search around the target location (both in x and y direction around the target). This approach is not ideal since, given a current state of movement, the target is expected to move in only one dimension (e.g. in the x-direction). Also, our dataset assumes that the target is moving with minimal speed so searching far from the target's current location will also not yield good results.

• Searching on redundant spaces can be avoided by approaches like particle filter framework. This framework, given the tracker's current state and configurations (location and velocity), propagates particles/samples to estimate the target. Once localized, any future particles are propagated only in the possible places the framework estimates the target to be present. Hence, particle filter when used with depth data and image features used by kernel correlation tracker, not only better re-identifies targets (coming out of occlusion) but also make it more accurate since tracker searches the right places reducing the possibility of false positives. The study of our proposed approach was experimented on a dataset particularly focussing on occlusion scenarios (with human presence). The dataset included diverse scenarios including the speed of the moving target (fast, slow) to the state of target w.r.t time (stable or moving). The results show that though overall tracking accuracy is marginally better, it helps in the stages where the target needs to be re-detected right after occlusion.

## 6.2. Future Works:

The proposed structure of the depth-based and probabilistic tracking system is a comprehensive framework that includes color-based tracking and depth-based tracking. The potential scope of future work from this thesis is as follows:

**More intelligent allocation of particles**: In this thesis, the Bhattacharya method of sampling was proposed to generate and propagate samples more effectively. This version of generating particles uses color histogram and information acquired from motion space. For a future study, we can improve the sampling strategy to address the particle degeneracy problem. For example, as shown in [121], we can use an improved particle filter based on Pearson Correlation Coefficient (PPC). PPC when used in the prediction step, can help in determining whether the particles are close to the true states and solve the degeneracy problem. Another approach is the use of a quantum particle filter. For example in [122] [123], authors propose using Quantum Particle Filter (QPF) which enables the tracker to handle the abrupt changes in speed and direction, whereas, the hybridization with Mean-Shift enhances the computational efficiency by reducing the number of particles. In [124], the authors propose the use of QPF at the resampling stage in particle filters to expand the diversity of particle sets which also solves the degeneracy problem.

**Tracking using multiple sensors**: In this thesis, the proposed framework is implemented and experimented using a single Microsoft Kinect sensor that is fixed on top of a tripod. Future works could include a network of sensors which can be achieved by expanding the number and capacity of the sensors. Using a multiple sensor system, or stationary sensors mounted on a mobile platform, we can also expand the coverage and flexibility of tracking. Problems like partial or full occlusions can be addressed by such a network of sensors as it would provide different sides of the subject and thus more information. For example, if a view of a body part is obstructed from one camera/sensor, it might be possible that another camera/sensor can obtain a view of the body part hidden from the first sensor. Appropriately combining data obtained from multiple Kinect sensors in this way can provide more accurate tracking compared with a single sensor. Works like [125] have explored the use of five Kinect sensors to achieve the best performance in the workspace for skeleton tracking in case of self-occlusion. Hence, the use of multiple Kinect devices could account for problems caused by occlusions.

**Hardware accelerations:** Both particle filter and sliding window approaches have a challenge when it comes to high computational loads. Particle filter algorithms are inherently parallelizable, however, likelihood evaluations are still a big bottleneck. When N particles are used to approximate the true posterior probability distribution, N independent, identical likelihood evaluations should be performed at each time step. Due to the independence of each likelihood evaluation for a particle from one another, the likelihood computation can be parallelized. For example, works like [126] report the acceleration of their particle filter algorithm on a single GPU at the resampling stage. In [127], they propose a robust particle filter parallelized on a GPU that can track a known 3D object model over a sequence of RGB-D images. Hence, with the availability of massive computation power in the form of modern GPUs, the software implementation of the tracker can benefit by designing it on the GPU for fast and robust real-time people tracking.

**Multi-target tracking**: Currently the system focuses on single targets on scenarios of occlusion and out-of-view. However, we can extend this work to multi-target tracking for both indoor and outdoor environments. Similar to the Princeton dataset we used for single target tracking evacuation, we can use challenges like MOT [128] for multi-target tracking evaluations. In [129], the authors present a multi-target tracking algorithm based on optical flow histogram and Kalman filtering. In [130], authors re-define the well-known mean-shift clustering method using asynchronous events instead of conventional frames and demonstrate its potential in multi-target tracking applications.

**Deep-learning method for target re-identification**: This thesis focuses on depth-based online learning and a probabilistic framework for tracking and target re-identification. The work discussed in this thesis does not involve the use of neural networks to extract deep features for target identification. In future work, we intend to incorporate deep learning methods to solve this issue. Despite immense progress in the field of computer vision and deep learning, target re-identification remains to be completely solved. Works like Yu-Jhe *et. al* [131] propose a person re-identification strategy using a novel representation learning method that can generate a shape-based feature representation that is in-variant to clothing changes. Most recently, [132] provided a survey of various deep learning methods, both supervised and unsupervised, available for person re-identification. This paper shows that using neural networks can improve the performance of people re-identification.

# References

[1]     X. Mei and H. Ling, "Robust visual tracking using l1 minimization," *2009 IEEE 12th Int. Conf. Comput. Vis.*, no. Iccv, pp. 1436–1443, 2009.

[2]     G. D. Hager, M. Dewan, and C. V. Stewart, "Multiple kernel tracking with SSD," *Proc. 2004 IEEE Comput. Soc. Conf. Comput. Vis. Pattern Recognition, 2004. CVPR 2004.*, vol. 1, pp. 790–797, 2004.

[3]     V. Parameswaran, V. Ramesh, and I. Zoghlami, "Tunable kernels for tracking," *Proc. IEEE Comput. Soc. Conf. Comput. Vis. Pattern Recognit.*, vol. 2, pp. 2179–2186, 2006.

[4]     A. Elgammal, R. Duraiswami, and L. S. Davis, "Probabilistic tracking in joint feature-spatial spaces," pp. I-781-I–788, 2003.

[5]     P. Viola and W. M. W. III, "Alignment by Maximization of Mutual Information PAUL," *Int. J. Comput. Vis.*, vol. 24, no. 2, pp. 137–154, 1997.

[6]     H. Xiao, Z. Li, C. Yang, W. Yuan, and L. Wang, "RGB-D sensor-based visual target detection and tracking for an intelligent wheelchair robot in indoors environments," *Int. J. Control. Autom. Syst.*, vol. 13, no. 3, pp. 521–529, 2015.

[7]     S. Payandeh, *Visual Tracking in Conventional Minimally Invasive Surgery*. Chapman and Hall/CRC, 2016.

[8]     X. Dai and S. Payandeh, "Geometry-based object association and consistent labeling in multi-camera surveillance," *IEEE J. Emerg. Sel. Top. Circuits Syst.*, vol. 3, no. 2, pp. 175–184, 2013.

[9]     Z. Pan, S. Liu, A. K. Sangaiah, and K. Muhammad, "Visual attention feature (VAF) : A novel strategy for visual tracking based on cloud platform in intelligent surveillance systems," *J. Parallel Distrib. Comput.*, vol. 120, pp. 182–194, 2018.

[10]    M. Mueller, N. Smith, and B. Ghanem, "A Benchmark and Simulator for UAV Tracking-Supplementary Material," pp. 1–8.

[11]    D. Wang and H. Lu, "Visual tracking via probability continuous outlier model," *Proc. IEEE Comput. Soc. Conf. Comput. Vis. Pattern Recognit.*, pp. 3478–3485, 2014.

[12]    T. Zhang, B. Ghanem, S. Liu, and N. Ahuja, "Low-rank sparse learning for robust visual tracking," *Lect. Notes Comput. Sci. (including Subser. Lect. Notes Artif. Intell. Lect. Notes Bioinformatics)*, vol. 7577 LNCS, no. PART 6, pp. 470–484, 2012.

[13]     T. Zhang, A. Bibi, and B. Ghanem, "In Defense of Sparse Tracking: Circulant Sparse Tracker," *2016 IEEE Conf. Comput. Vis. Pattern Recognit.*, no. 3, pp. 3880–3888, 2016.

[14]     S. Hare *et al.*, "Struck: Structured Output Tracking with Kernels," *IEEE Trans. Pattern Anal. Mach. Intell.*, vol. 38, no. 10, pp. 2096–2109, 2016.

[15]     Z. Yang, Z. Dai, Y. Yang, J. Carbonell, R. Salakhutdinov, and Q. V. Le, "XLNet: Generalized autoregressive pretraining for language understanding," *arXiv*, no. NeurIPS, pp. 1–18, 2019.

[16]     M. Tan and Q. V. Le, "EfficientNet: Rethinking model scaling for convolutional neural networks," *arXiv*, 2019.

[17]     J. Pineau *et al.*, "Improving reproducibility in machine learning research (A report from the neurips 2019 reproducibility program)," *arXiv*, 2020.

[18]     D. S. Bolme, J. R. Beveridge, B. A. Draper, and Y. M. Lui, "Visual object tracking using adaptive correlation filters," *Proc. IEEE Comput. Soc. Conf. Comput. Vis. Pattern Recognit.*, pp. 2544–2550, 2010.

[19]     M. Danelljan, F. S. Khan, M. Felsberg, and J. Van De Weijer, "Adaptive Color Attributes for Real-Time Visual Tracking Martin."

[20]     P. Liang, E. Blasch, and H. Ling, "Encoding Color Information for Visual Tracking: Algorithms and Benchmark," *IEEE Trans. Image Process.*, vol. 24, no. 12, pp. 5630–5644, 2015.

[21]     J. F. Henriques, R. Caseiro, P. Martins, and J. Batista, "High-speed tracking with kernelized correlation filters," *IEEE Trans. Pattern Anal. Mach. Intell.*, vol. 37, no. 3, pp. 583–596, 2015.

[22]     Y. Li and J. Zhu, "A scale adaptive kernel correlation filter tracker with feature integration," *Lect. Notes Comput. Sci. (including Subser. Lect. Notes Artif. Intell. Lect. Notes Bioinformatics)*, vol. 8926, pp. 254–265, 2015.

[23]     W. Zuo, X. Wu, L. Lin, L. Zhang, and M. H. Yang, "Learning support correlation filters for visual tracking," *IEEE Trans. Pattern Anal. Mach. Intell.*, vol. 41, no. 5, pp. 1158–1172, 2019.

[24]     M. Danelljan, G. Hager, F. S. Khan, and M. Felsberg, "Learning spatially regularized correlation filters for visual tracking," *Proc. IEEE Int. Conf. Comput. Vis.*, vol. 2015 Inter, pp. 4310–4318, 2015.

[25]     M. Danelljan, A. Robinson, F. S. Khan, and M. Felsberg, "Beyond Correlation Filters: Learning ContinuousConvolution Operators for Visual Tracking," *Eccv*, no. 5, pp. 1–9, 2016.

[26]    Y. Wu, J. Lim, and M. H. Yang, "Object tracking benchmark," *IEEE Trans. Pattern Anal. Mach. Intell.*, vol. 37, no. 9, pp. 1834–1848, 2015.

[27]    M. Kristan *et al.*, "The Visual Object Tracking VOT2017 Challenge Results," *Proc. - 2017 IEEE Int. Conf. Comput. Vis. Work. ICCVW 2017*, vol. 2018-Janua, pp. 1949–1972, 2018.

[28]    H. Fan *et al.*, "Lasot: A high-quality benchmark for large-scale single object tracking," *Proc. IEEE Comput. Soc. Conf. Comput. Vis. Pattern Recognit.*, vol. 2019-June, pp. 5369–5378, 2019.

[29]    M. Yang and Y. Jia, "Temporal dynamic appearance modeling for online multi-person tracking," *Comput. Vis. Image Underst.*, vol. 153, pp. 16–28, 2016.

[30]    Y. Xiang, A. Alahi, and S. Savarese, "Learning to track: Online multi-object tracking by decision making," *Proc. IEEE Int. Conf. Comput. Vis.*, vol. 2015 Inter, pp. 4705–4713, 2015.

[31]    H. Pirsiavash, D. Ramanan, and C. C. Fowlkes, "Globally-optimal greedy algorithms for tracking a variable number of objects," *Proc. IEEE Comput. Soc. Conf. Comput. Vis. Pattern Recognit.*, pp. 1201–1208, 2011.

[32]    W. Choi, "Near-online multi-target tracking with aggregated local flow descriptor," *Proc. IEEE Int. Conf. Comput. Vis.*, vol. 2015 Inter, pp. 3029–3037, 2015.

[33]    S. Yadav and S. Payandeh, "Understanding Tracking Methodology of Kernelized Correlation Filter," in *2018 IEEE 9th Annual Information Technology, Electronics and Mobile Communication Conference, IEMCON 2018*, 2019, pp. 1330–1336.

[34]    S. Yadav and S. Payandeh, "Real-Time Experimental Study of Kernelized Correlation Filter Tracker using RGB Kinect Camera," in *2018 IEEE 9th Annual Information Technology, Electronics and Mobile Communication Conference, IEMCON 2018*, 2019, pp. 1324–1329.

[35]    A. W. M. Smeulders, D. M. Chu, R. Cucchiara, S. Calderara, A. Dehghan, and M. Shah, "Visual tracking: An experimental survey," *IEEE Trans. Pattern Anal. Mach. Intell.*, vol. 36, no. 7, pp. 1442–1468, 2014.

[36]    Z. Chen, Z. Hong, and D. Tao, "An Experimental Survey on Correlation Filter-based Tracking," pp. 1–13, 2015.

[37]    S. Paheding, A. Essa, E. Krieger, and V. Asari, "Tracking visual objects using pyramidal rotation invariant features," *Opt. Pattern Recognit. XXVII*, vol. 9845, no. February 2016, p. 98450B, 2016.

[38]    K. Okuma, A. Taleghani, N. De Freitas, J. J. Little, and D. G. Lowe, "A boosted particle filter: Multitarget detection and tracking," *Lect. Notes Comput. Sci. (including Subser. Lect. Notes Artif. Intell. Lect. Notes Bioinformatics)*, vol. 3021, pp. 28–39, 2004.

[39]    A. Almeida, J. Almeida, and R. Araújo, "Real-time Tracking of Multiple Moving Objects Using Particle Filters and Probabilistic Data Association," *Autom. J.*, vol. 46, no. 1–2, pp. 39–48, 2005.

[40]    X. Yang, C. Ma, J.-B. Huang, and M.-H. Yang, "Hierarchical Convolutional Features for Visual Tracking," *Proc. IEEE Int. Conf. Comput. Vis.*, pp. 3074–3082, 2015.

[41]    João F. Henriques, Caseiro Rui, Martins Pedro, and Batista Horge, "Kernelized Correlation Filters," vol. 37, no. 3, pp. 583–596, 2015.

[42]    A. S. Montero, J. Lang, and R. Laganière, "Scalable Kernel Correlation Filter with Sparse Feature Integration," *Proc. IEEE Int. Conf. Comput. Vis.*, vol. 2016-Febru, pp. 587–594, 2016.

[43]    M. Danelljan, G. Häger, F. S. Khan, and M. Felsberg, "Accurate scale estimation for robust visual tracking," in *BMVC 2014 - Proceedings of the British Machine Vision Conference 2014*, 2014.

[44]    W. Zuo, S. Member, X. Wu, L. Lin, and S. Member, "Learning Support Correlation Filtersfor Visual Tracking," *IEEE Trans. Pattern Anal. Mach. Intell.*, vol. 41, no. 5, pp. 1158–1172, 2019.

[45]    K. Zhang, L. Zhang, Q. Liu, D. Zhang, and M. H. Yang, "Fast visual tracking via dense spatio-temporal context learning," *Lect. Notes Comput. Sci. (including Subser. Lect. Notes Artif. Intell. Lect. Notes Bioinformatics)*, vol. 8693 LNCS, no. PART 5, pp. 127–141, 2014.

[46]    T. Liu, G. Wang, and Q. Yang, "Real-time part-based visual tracking via adaptive correlation filters," *Proc. IEEE Comput. Soc. Conf. Comput. Vis. Pattern Recognit.*, vol. 07-12-June, pp. 4902–4912, 2015.

[47]    T. Zhang, C. Xu, and M. H. Yang, "Multi-task correlation particle filter for robust object tracking," in *Proceedings - 30th IEEE Conference on Computer Vision and Pattern Recognition, CVPR 2017*, 2017.

[48]    L. Čehovin, M. Kristan, and A. Leonardis, "Robust visual tracking using an adaptive coupled-layer visual model," *IEEE Trans. Pattern Anal. Mach. Intell.*, vol. 35, no. 4, pp. 941–953, 2013.

[49]    A. Lukežič, L. Čehovin Zajc, and M. Kristan, "Deformable Parts Correlation Filters for Robust Visual Tracking," *IEEE Trans. Cybern.*, vol. 48, no. 6, pp. 1849–1861, 2018.

[50]     L. Bertinetto, J. Valmadre, J. F. Henriques, A. Vedaldi, and P. H. S. Torr, "Fully-convolutional siamese networks for object tracking," *Lect. Notes Comput. Sci. (including Subser. Lect. Notes Artif. Intell. Lect. Notes Bioinformatics)*, vol. 9914 LNCS, pp. 850–865, 2016.

[51]     G. Fern, G. Nebehay, R. Pflugfelder, A. Gupta, and A. Bibi, "The Visual Object Tracking VOT2015 challenge results."

[52]     A. Lukežič, J. Matas, and M. Kristan, "D3S - A discriminative single shot segmentation tracker," *Proc. IEEE Comput. Soc. Conf. Comput. Vis. Pattern Recognit.*, pp. 7131–7140, 2020.

[53]     Q. Wang, L. Zhang, L. Bertinetto, W. Hu, and P. H. S. Torr, "Fast online object tracking and segmentation: A unifying approach," *arXiv*, pp. 1328–1338, 2018.

[54]     S. Song and J. Xiao, "Tracking revisited using RGBD camera: Unified benchmark and baselines," *Proc. IEEE Int. Conf. Comput. Vis.*, pp. 233–240, 2013.

[55]     D. Chrapek, V. Beran, and P. Zemcik, "Depth-Based Filtration for Tracking Boost," in *Advanced Concepts for Intelligent Vision Systems*, 2015, pp. 217–228.

[56]     Z. Kalal, K. Mikolajczyk, and J. Matas, "Tracking-Learning-Detection.," *IEEE Trans. Pattern Anal. Mach. Intell.*, vol. 34, no. 7, pp. 1409–1422, 2011.

[57]     Y. Xie, Y. Lu, and S. Gu, "RGB-D Object Tracking with Occlusion Detection," *Proc. - 2019 15th Int. Conf. Comput. Intell. Secur. CIS 2019*, pp. 11–15, 2019.

[58]     M. Camplani *et al.*, "Real-time RGB-D Tracking with Depth Scaling Kernelised Correlation Filters and Occlusion Handling," pp. 145.1-145.11, 2015.

[59]     N. A. Hou, X.-G. Zhao, and Zeng-Guang, "Online RGB-D Tracking via Detection-Learning-Segmentation," *Pattern Recognit. (ICPR), 2016 23rd Int. Conf.*, pp. 1231–1236, 2016.

[60]     Y. Qian, A. Lukežič, M. Kristan, J.-K. Kämäräinen, and J. Matas, "DAL -- A Deep Depth-aware Long-term Tracker," 2019.

[61]     S. Liu, D. Liu, G. Srivastava, D. Połap, and M. Woźniak, "Overview and methods of correlation filter algorithms in object tracking," *Complex Intell. Syst.*, no. 0123456789, 2020.

[62]     D. Kahneman, *Attention and effort. Englewood Cliffs, NJ.* 1973.

[63]     U. Neisser, *Cognitive psychology: Classic edition*. Psychology Press, 2014.

[64]     R. A. Kinchla and J. M. Wolfe, "The order of visual processing: 'Top-down,' 'bottom-up,' or 'middle-out,'" *Percept. Psychophys.*, vol. 25, no. 3, pp. 225–231, 1979.

[65]    C. E. Cohen, "Person categories and social perception: Testing some boundaries of the processing effect of prior knowledge," *J. Pers. Soc. Psychol.*, vol. 40, no. 3, pp. 441–452, 1981.

[66]    M. Danelljan, G. Häger, F. S. Khan, and M. Felsberg, "Adaptive decontamination of the training set: A unified formulation for discriminative visual tracking," *Proc. IEEE Comput. Soc. Conf. Comput. Vis. Pattern Recognit.*, vol. 2016-Decem, pp. 1430–1438, 2016.

[67]    Qing Wang, Feng Chen, Jimei Yang, Wenli Xu, and Ming-Hsuan Yang, "Transferring Visual Prior for Online Object Tracking," *IEEE Trans. Image Process.*, vol. 21, no. 7, pp. 3296–3305, 2012.

[68]    D. Miramontes-Jaramillo, V. Kober, and V. H. Díaz-Ramírez, "Robust illumination-invariant tracking algorithm based on HOGs," *Appl. Digit. Image Process. XXXVIII*, vol. 9599, no. September 2015, p. 95991Q, 2015.

[69]    M. Rasoulidanesh, S. Yadav, S. Herath, Y. Vaghei, and S. Payandeh, "Deep attention models for human tracking using RGBD," *Sensors (Switzerland)*, vol. 19, no. 4, 2019.

[70]    A. Yao, J. Gall, and L. Van Gool, "A hough transform-based voting framework for action recognition," *Proc. IEEE Comput. Soc. Conf. Comput. Vis. Pattern Recognit.*, pp. 2061–2068, 2010.

[71]    W. H. Li, A. M. Zhang, and L. Kleeman, "Fast Global Reflectional Symmetry Detection for Robotic Grasping and Visual Tracking," *Australas. Conf. Robot. Autom.*, no. 2, 2005.

[72]    N. I. Glumov, E. I. Kolomiyetz, and V. V. Sergeyev, "Detection of objects on the image using a sliding window mode," *Opt. Laser Technol.*, vol. 27, no. 4, pp. 241–249, 1995.

[73]    P. Viola and M. Jones, "Rapid Object Detection Using a Boosted Cascade of Simple Features," *Conf. Comput. Vis. Pattern Recognit.*, vol. 1, pp. 511–518, 2001.

[74]    N. Dalal and W. Triggs, "Histograms of Oriented Gradients for Human Detection," *2005 IEEE Comput. Soc. Conf. Comput. Vis. Pattern Recognit. CVPR05*, vol. 1, no. 3, pp. 886–893, 2004.

[75]    D. Forsyth, "Object detection with discriminatively trained part-based models," *Computer (Long. Beach. Calif).*, vol. 47, no. 2, pp. 6–7, 2014.

[76]    Y. Wei and L. Tao, "Efficient histogram-based sliding window," *Proc. IEEE Comput. Soc. Conf. Comput. Vis. Pattern Recognit.*, pp. 3003–3010, 2010.

[77]    S. Song and J. Xiao, "Deep sliding shapes for amodal 3D object detection in RGB-D images," *Proc. IEEE Comput. Soc. Conf. Comput. Vis. Pattern Recognit.*, vol. 2016-Decem, pp. 808–816, 2016.

[78]    D. Frossard and R. Urtasun, "End-to-end Learning of Multi-sensor 3D Tracking by Detection," *Proc. - IEEE Int. Conf. Robot. Autom.*, pp. 635–642, 2018.

[79]    L. Bertinetto, J. Valmadre, S. Golodetz, O. Miksik, and P. H. S. Torr, "Staple: Complementary learners for real-time tracking," *Proc. IEEE Comput. Soc. Conf. Comput. Vis. Pattern Recognit.*, vol. 2016-Decem, pp. 1401–1409, 2016.

[80]    S. Salti, A. Lanza, and L. Di Stefano, "Synergistic change detection and tracking," *IEEE Trans. Circuits Syst. Video Technol.*, vol. 25, no. 4, pp. 609–622, 2015.

[81]    R. Zhang and J. Ding, "Object tracking and detecting based on adaptive background subtraction," *Procedia Eng.*, vol. 29, pp. 1351–1355, 2012.

[82]    B. Babenko, M. H. Yang, and S. Belongie, "Robust object tracking with online multiple instance learning," *IEEE Trans. Pattern Anal. Mach. Intell.*, vol. 33, no. 8, pp. 1619–1632, 2011.

[83]    P. J. Davis, *Circulant matrices*. American Mathematical Soc., 2013.

[84]    D. Kalman and J. E. White, "Polynomial equations and circulant matrices," *Am. Math. Mon.*, vol. 108, no. 9, pp. 821–840, 2001.

[85]    S. Aranda, S. Mart, and F. Bullo, "On Optimal Sensor Placement and Motion Coordination for Target Tracking," pp. 4544–4549, 2011.

[86]    W. Yin, S. Morgan, J. Yang, and Y. Zhang, "Practical compressive sensing with Toeplitz and circulant matrices," *Vis. Commun. Image Process. 2010*, vol. 7744, p. 77440K, 2010.

[87]    H. Buchner, J. Benesty, T. Gänsler, and W. Kellermann, "Robust extended multidelay filter and double-talk detector for acoustic echo cancellation," *IEEE Trans. Audio, Speech Lang. Process.*, vol. 14, no. 5, pp. 1633–1643, 2006.

[88]    F. Xu, T. N. Davidson, J. K. Zhang, S. S. Chan, and K. M. Wong, "Design of block transceivers with MMSE decision feedback detection," *ICASSP, IEEE Int. Conf. Acoust. Speech Signal Process. - Proc.*, vol. III, 2005.

[89]    R. M. Gray, "Toeplitz and Circulant Matrices: A Review," *Found. Trends® Commun. Inf. Theory*, vol. 2, no. 3, pp. 155–239, 2005.

[90]    J. F. Henriques, R. Caseiro, P. Martins, and J. Batista, "Exploiting the circulant structure of tracking-by-detection with kernels," *Lect. Notes Comput. Sci. (including Subser. Lect. Notes Artif. Intell. Lect. Notes Bioinformatics)*, vol. 7575 LNCS, no. PART 4, pp. 702–715, 2012.

[91]    M. Mueller, N. Smith, and B. Ghanem, "Context-aware correlation filter tracking," *Proc. - 30th IEEE Conf. Comput. Vis. Pattern Recognition, CVPR 2017*, vol. 2017-Janua, pp. 1387–1395, 2017.

[92]    M. Tang and J. Feng, "Multi-kernel correlation filter for visual tracking," *Proc. IEEE Int. Conf. Comput. Vis.*, vol. 2015 Inter, pp. 3038–3046, 2015.

[93]    C. Ding *et al.*, "CirCNN: Accelerating and Compressing Deep Neural Networks Using Block-CirculantWeight Matrices," 2017.

[94]    C. Cortes and V. Vapnik, "Support Vector Network," in *Machine Learning*, 1995, pp. 273–297.

[95]    H. K. Galoogahi, T. Sim, and S. Lucey, "Multi-Channel Correlation Filters," in *CVPR*, 2013.

[96]    A. Cotter, J. Keshet, and N. Srebro, "Explicit Approximations of the Gaussian Kernel," pp. 1–11, 2011.

[97]    L. Čehovin, A. Leonardis, and M. Kristan, "Visual Object Tracking Performance Measures Revisited," *IEEE Trans. Image Process.*, vol. 25, no. 3, pp. 1261–1274, 2016.

[98]    D. Wang, H. Lu, and M. H. Yang, "Online object tracking with sparse prototypes," *IEEE Trans. Image Process.*, vol. 22, no. 1, pp. 314–325, 2013.

[99]    F. Yang, H. Lu, and M. H. Yang, "Robust superpixel tracking," *IEEE Trans. Image Process.*, vol. 23, no. 4, pp. 1639–1651, 2014.

[100]   G. Xu, H. Zhu, L. Deng, L. Han, Y. Li, and H. Lu, "Dilated-aware discriminative correlation filter for visual tracking," *World Wide Web*, vol. 22, no. 2, pp. 791–805, 2019.

[101]   L. Qu, K. Liu, B. Yao, J. Tang, and W. Zhang, "Real-time visual tracking with ELM augmented adaptive correlation filter," *Pattern Recognit. Lett.*, vol. 127, pp. 138–145, 2019.

[102]   J. Choi, H. J. Chang, S. Yun, T. Fischer, Y. Demiris, and J. Y. Choi, "Attentional correlation filter network for adaptive visual tracking," in *Proceedings - 30th IEEE Conference on Computer Vision and Pattern Recognition, CVPR 2017*, 2017, vol. 2017-Janua, pp. 4828–4837.

[103]   D. Mishra and J. Matas, "The Visual Object Tracking VOT2017 Challenge Results The Visual Object Tracking VOT2017 challenge results," *Icvc*, no. November 2017, 2019.

[104]   A. Lukežič, T. Vojíř, L. Č. Zajc, J. Matas, and M. Kristan, "Discriminative correlation filter with channel and spatial reliability," *Proc. - 30th IEEE Conf. Comput. Vis. Pattern Recognition, CVPR 2017*, vol. 2017-Janua, pp. 4847–4856, 2017.

[105]   M. Danelljan, G. Bhat, S. Khan, M. Felsberg, F. S. Khan, and M. Felsberg, "ECO: Efficient Convolution Operators for Tracking," pp. 6638–6646, 2016.

[106]   L. Keselman *et al.*, "Intel RealSense Stereoscopic Depth Cameras," *Comput. Vis. Pattern Recognit.*, vol. 68, no. January, pp. 163–176, 2017.

[107]   C. Premebida, G. Monteiro, U. Nunes, and P. Peixoto, "A Lidar and vision-based approach for pedestrian and vehicle detection and tracking," *IEEE Conf. Intell. Transp. Syst. Proceedings, ITSC*, pp. 1044–1049, 2007.

[108]   S. Hannuna *et al.*, "DS-KCF: a real-time tracker for RGB-D data," *J. Real-Time Image Process.*, vol. 16, no. 5, pp. 1439–1458, 2019.

[109]   M. Everingham, L. Van Gool, C. K. I. Williams, J. Winn, and A. Zisserman, "The pascal visual object classes (VOC) challenge," *Int. J. Comput. Vis.*, vol. 88, no. 2, pp. 303–338, 2010.

[110]   J. Kwon and K. M. Lee, "Visual tracking decomposition," *Proc. IEEE Comput. Soc. Conf. Comput. Vis. Pattern Recognit.*, pp. 1269–1276, 2010.

[111]   K. Zhang, L. Zhang, and M. H. Yang, "Fast Compressive Tracking," *IEEE Trans. Pattern Anal. Mach. Intell.*, vol. 36, no. 10, pp. 2002–2015, 2014.

[112]   B. Babenko, S. Belongie, and M. H. Yang, "Visual tracking with online multiple instance learning," *2009 IEEE Comput. Soc. Conf. Comput. Vis. Pattern Recognit. Work. CVPR Work. 2009*, vol. 2009 IEEE, pp. 983–990, 2009.

[113]   H. Grabner, C. Leistner, and H. Bischof, "Semi-supervised On-Line Boosting for Robust Tracking," no. 813399, pp. 234–247, 2008.

[114]   A. Bibi, T. Zhang, and B. Ghanem, "3D part-based sparse tracker with automatic synchronization and registration," *Proc. IEEE Comput. Soc. Conf. Comput. Vis. Pattern Recognit.*, vol. 2016-Decem, pp. 1439–1448, 2016.

[115]   G. S. Walia, A. Kumar, A. Saxena, K. Sharma, and K. Singh, "Robust object tracking with crow search optimized multi-cue particle filter," *Pattern Anal. Appl.*, vol. 23, no. 3, pp. 1439–1455, 2020.

[116]   N. P. Santos, V. Lobo, and A. Bernardino, "Unmanned Aerial Vehicle Tracking Using a Particle Filter Based Approach," *2019 IEEE Int. Underw. Technol. Symp. UT 2019 - Proc.*, 2019.

[117]  K. Meshgi, S. ichi Maeda, S. Oba, H. Skibbe, Y. zhe Li, and S. Ishii, "An occlusion-aware particle filter tracker to handle complex and persistent occlusions," *Comput. Vis. Image Underst.*, vol. 150, pp. 81–94, 2016.

[118]  X. Wang, T. Li, S. Sun, and J. M. Corchado, "A survey of recent advances in particle filters and remaining challenges for multitarget tracking," *Sensors (Switzerland)*, vol. 17, no. 12, pp. 1–21, 2017.

[119]  "Explain process noise terminology in Kalman Filter - Stack Overflow." [Online]. Available: https://stackoverflow.com/questions/19537884/explain-process-noise-terminology-in-kalman-filter. [Accessed: 20-Jan-2021].

[120]  A. Bhattacharyya, "On a measure of divergence between two statistical populations defined by their probability distributions," *Bull. Calcutta Math. Soc.*, vol. 35, pp. 99–109, 1943.

[121]  H. Zhou, Z. Deng, Y. Xia, and M. Fu, "A new sampling method in particle filter based on Pearson correlation coefficient," *Neurocomputing*, vol. 216, pp. 208–215, 2016.

[122]  P. P. Dash and D. Patra, "An efficient hybrid framework for visual tracking using Exponential Quantum Particle Filter and Mean Shift optimization," *Multimed. Tools Appl.*, vol. 79, no. 29–30, pp. 21513–21537, 2020.

[123]  P. P. Dash, S. K. Mishra, and Di. Patra, "Animal tracking in wildlife footage with quantum particle filter (QPF)," *Proc. - 2019 Int. Conf. Inf. Technol. ICIT 2019*, pp. 515–520, 2019.

[124]  Z. Liu, J. Shang, and X. Hua, "Smart City Moving Target Tracking Algorithm Based on Quantum Genetic and Particle Filter," *Wirel. Commun. Mob. Comput.*, vol. 2020, 2020.

[125]  S. Moon, Y. Park, D. W. Ko, and I. H. Suh, "Multiple kinect sensor fusion for human skeleton tracking using Kalman filtering," *Int. J. Adv. Robot. Syst.*, vol. 13, no. 2, pp. 1–10, 2016.

[126]  L. Fan, B. Hongkui, X. Jiajun, Y. Chenlong, and L. Xuhui, "A dual channel perturbation particle filter algorithm based on GPU acceleration," *J. Syst. Eng. Electron.*, vol. 29, no. 4, pp. 854–863, 2018.

[127]  C. Choi and H. I. Christensen, "RGB-D object tracking: A particle filter approach on GPU," *IEEE Int. Conf. Intell. Robot. Syst.*, pp. 1084–1091, 2013.

[128]  P. Dendorfer *et al.*, "MOT20: A benchmark for multi object tracking in crowded scenes," *arXiv*, pp. 1–7, 2020.

[129] Z. Ge, F. Chang, and H. Liu, "Multi-target tracking based on Kalman filtering and optical flow histogram," *Proc. - 2017 Chinese Autom. Congr. CAC 2017*, vol. 2017-Janua, pp. 2540–2545, 2017.

[130] F. Barranco, C. Fermuller, and E. Ros, "Real-time clustering and multi-target tracking using event-based sensors," *arXiv*, pp. 5764–5769, 2018.

[131] Y. Li, "Learning Shape Representations for Person Re-Identification under Clothing Change."

[132] M. Ye, J. Shen, G. Lin, T. Xiang, L. Shao, and S. C. H. Hoi, "Deep Learning for Person Re-identification: A Survey and Outlook," 2020.

# Appendix A.

# Microsoft Kinect Senor V2

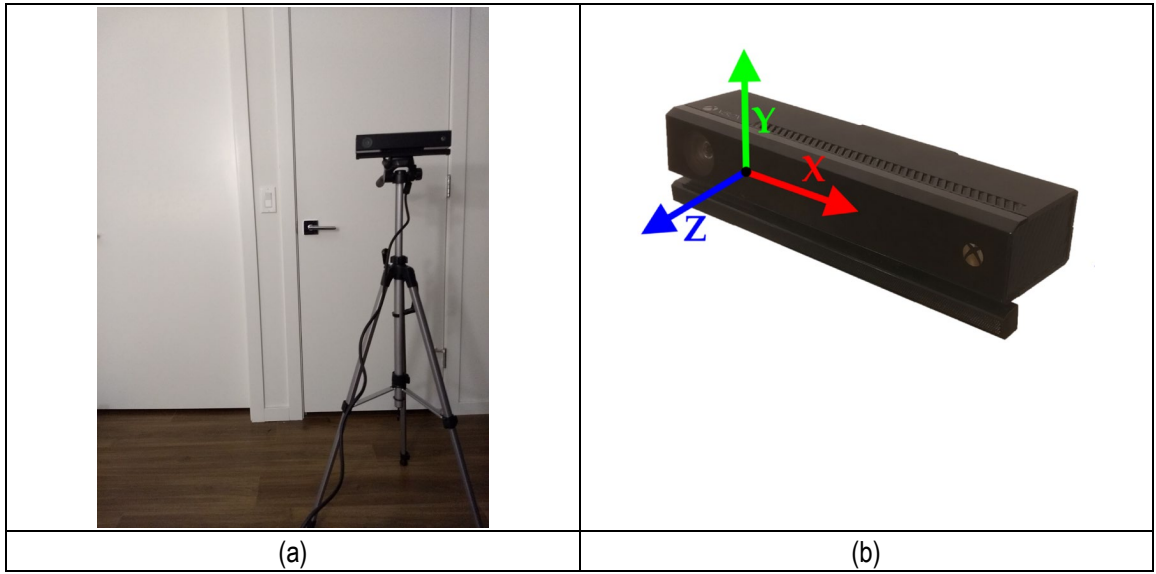

| (a) | (b) |
|---|---|

**Figure A.1.**  **(a) Kinect V2 sensor used for the experiments mentioned in the thesis (b) x,y,z coordinate for Kinect sensor**

Kinect is a line of motion sensing input device which has a VGA video camera and depth sensor, working together. It is developed and maintained by Microsoft. Kinect sensor accessory contains a combination of hardware and software. It has a flat black box that sits on a small platform that can be placed on a table, shelf, or on a tripod stand (shown in Figure A.1). The sensors working within it are:

**Color VGA video camera**. This video camera, commonly called an RGB camera, detects three color components: red, green and blue.

**Depth sensor**: An infrared projector and a monochrome CMOS (complementary metal-oxide-semiconductor) sensor work together to "see" the room in 3-D regardless of the lighting conditions

Kinect V2 was an improved version over its previous version Kinect V1. Kinect V2 runs at 30 FPS and has a depth resolution of 512 x 424 pixels and a color camera resolution of 1920 x 1080px. For accurate depth measurement, it is recommended to allow about 6 feet (1.8 meters) of play space between the user and the Kinect sensor.

# Appendix B.

# Ridge Regression (Dual)

We know that learning a linear classifier:

$$f(x) = \boldsymbol{w}^T \boldsymbol{x} + b \tag{B1}$$

can be formulated as learning an optimization problem over $\boldsymbol{w}$. This optimization problem is known as the primal problem. Instead, we can learn a linear classifier:

$$f(x) = \sum_{i}^{N} \alpha_i y_i \left( \boldsymbol{x}_i^T \boldsymbol{x} \right) + b \tag{B2}$$

by solving an optimization problem over $\alpha$. This is known as the dual problem as has its own advantageous.

This can be extended to ridge regression problems. Ridge regression has a closed-form solution for its optimization problem and can be given by:

$$\min_{\boldsymbol{w}} \sum_{i}^{n} (\boldsymbol{w}^T \boldsymbol{x_i} - y_i)^2 + \lambda \|\boldsymbol{w}\|^2 \tag{B3}$$

This optimization problem in $\boldsymbol{w}$ is a convex problem and can be solved by differentiating it w.r.t. $\boldsymbol{w}$ giving:

$$\boldsymbol{w} = (X^T X + \lambda I)^{-1} X^T \boldsymbol{y} \tag{B4}$$

where $X \in R^{nxm}$ is the data matrix. We can find its dual form solution by using Representer Theorem which states that for a regularized risk minimization problem:

$$\min_{w} \sum_{i}^{n} L\left(\boldsymbol{w}^T \boldsymbol{x_i}, y_i\right)^2 + \lambda \|\boldsymbol{w}\|^2 \tag{B5}$$

the solution can be given by:

$$\boldsymbol{w} = \sum_{i}^{n} \alpha_i \boldsymbol{x}_i = X^T \boldsymbol{\alpha}$$

(B6)

To find the dual form of solution given by Equation 31, we aim to find a relationship between Equation B4 and Equation B6. Hence, we use the Sherman-Morrison-Woodbury formula, which states:

$$(A^{-1} + B^T B)^{-1} B^T = AB^T (BAB^T + I)^{-1}$$

(B7)

Comparing the L.H.S. of Equation B7:

$$(A^{-1} + B^T B)^{-1} B^T$$

with (re-arranged) Equation B4:

$$(\lambda I + X^T X)^{-1} X^T$$

we get:

$$A = \frac{1}{\lambda} , B = X$$

(B8)

However, from Equation (B7), we know that:

$$(A^{-1} + B^T B)^{-1} B^T = AB^T (BAB^T + I)^{-1}$$

Therefore, if we use Equation (B8) to substitute $A = 1/\lambda$ , $B = X$ in the R.H.S. of Equation (B7), $AB^T (BAB^T + I)^{-1}$, we will get an equivalent of the closed-form solution of ridge regression given by Equation (B4) which is of form $(A^{-1} + B^T B)^{-1} B^T$ . The substitution can be shown as:

$$AB^T(BAB^T + I)^{-1}$$

Using $A = \frac{1}{\lambda}$, $B = X$ : $\qquad = \frac{1}{\lambda} X^T \left( X \frac{1}{\lambda} X^T + I \right)^{-1}$

$$= X^T (XX^T + \lambda I)^{-1} \tag{B9}$$

Hence, using the Sherman-Morrison-Woodbury formula and from Equation (B9), we now have:

$$\mathbf{w} = X^T (XX^T + \lambda I)^{-1} \mathbf{y}$$

Comparing this with Equation (B6) which dual form solution:

$$\boldsymbol{w} = \sum_i^n \alpha_i \boldsymbol{x}_i = X^T \boldsymbol{\alpha} \tag{B10}$$

we have the dual form solution of ridge regression as:

$$\boldsymbol{\alpha} = (XX^T + \lambda I)^{-1} \mathbf{y} \tag{B11}$$

or $\boldsymbol{\alpha} = (G + \lambda I)^{-1} \mathbf{y}$

where we have defined $G$ as Gram Matrix.

# Appendix C.

# Learned Function in Kernel Correlation

The learned function $f(z)$ in KCF can be obtained by mapping inputs by $\varphi$ (a mapping function) in B6 and apply into kernel function of the format:

$$k(\boldsymbol{u}, \boldsymbol{v}) = \varphi^T(\boldsymbol{u})\, \varphi(\boldsymbol{v}) \tag{C2}$$

to obtain

$$f(\boldsymbol{z'}) = \left( \sum_i^n \alpha_i \varphi(x_i') \right)^T \varphi(\boldsymbol{z'}) = \sum_i^n \alpha_i \kappa(x_i', z') \tag{C1}$$

# Appendix D

# Color Histogram

Color particle filters use a color histogram of a person as a reference to weigh the samples. It is computed using the distribution of color within the bounding box area of the subject/target. A color histogram can be created using RGB color space. A 1D histogram with $b$ bins (collection of pixels whose color falls in the range defined by the bin) is computed by calculating the number of pixels there exist for each color bin. Color space ranges from 0-255 and if no bin number is defined, it defaults to 255 bins as shown in

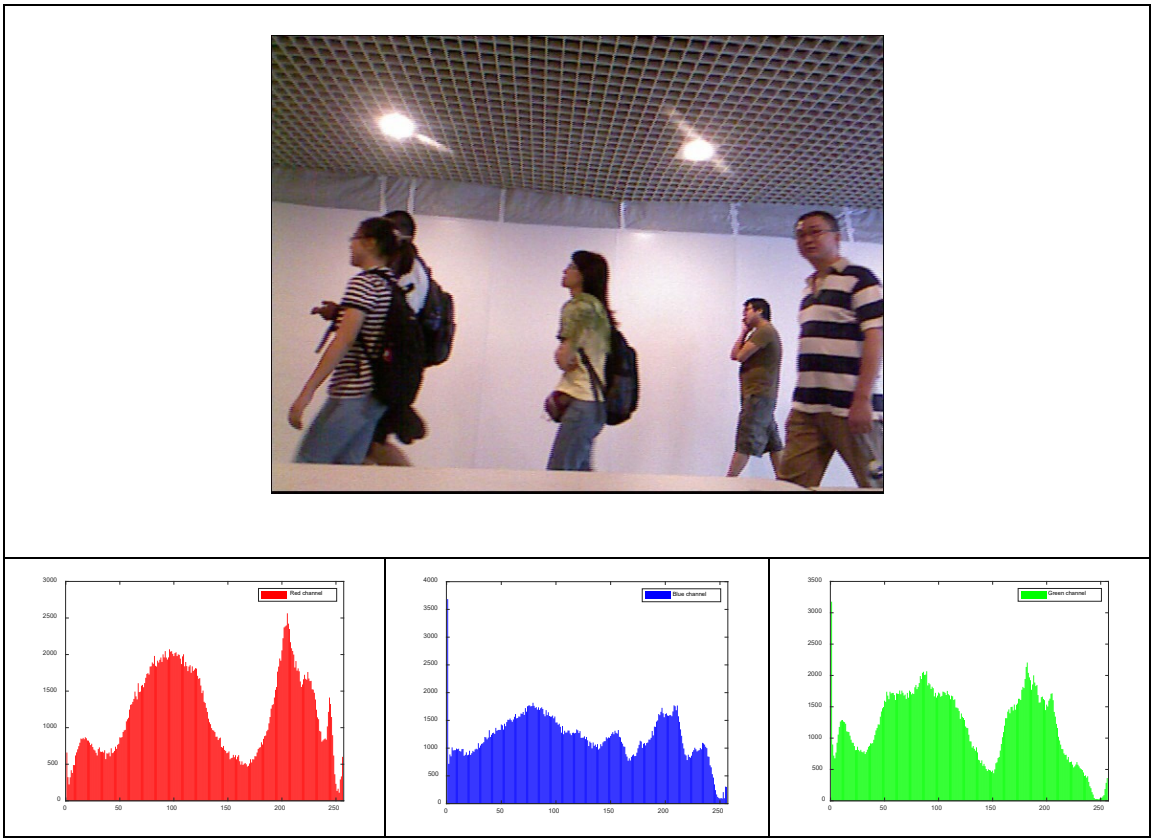

**Figure D.1.** **The figure shows the histogram of the RGB image where the y-axis is the intensity of the image and the x-axis is the bin (default to 255 bins)**